\documentclass{article}

\usepackage[accepted]{stylefiles/icml2021}

\usepackage[utf8]{inputenc} 
\usepackage[english]{babel} 		
\usepackage{xargs}                  
\usepackage[toc,page]{appendix}		

\usepackage{multibib} 
\newcites{appendix}{Appendix References}

\usepackage[colorinlistoftodos,prependcaption,textsize=tiny]{todonotes}
\newcommandx{\unsure}[2][1=]{\todo[linecolor=red,backgroundcolor=red!25,bordercolor=red,#1]{#2}}
\newcommandx{\change}[2][1=]{\todo[linecolor=blue,backgroundcolor=blue!25,bordercolor=blue,#1]{#2}}
\newcommandx{\info}[2][1=]{\todo[linecolor=orange,backgroundcolor=orange!25,bordercolor=orange,#1]{#2}}
\newcommandx{\improvement}[2][1=]{\todo[linecolor=violet,backgroundcolor=violet!25,bordercolor=violet,#1]{#2}}
\newcommandx{\thiswillnotshow}[2][1=]{\todo[disable,#1]{#2}}
\newcommandx{\draft}[2][1=]{\todo[inline,#1]{#2}}

\usepackage{graphicx} 		
\usepackage{caption}		%
\usepackage{subcaption}		
\usepackage{tikz}			

\usepackage{pgfplots}		
\pgfplotsset{compat=newest}
\PassOptionsToPackage{showframe=True}{graphicx}
\usepackage{graphbox}
\usepackage{wrapfig}
\usepackage{pifont} 
\usepackage{epstopdf}

\usepackage{float}

\usepackage{tabularx}
\usepackage{booktabs}       
\usepackage{multirow}

\newcolumntype{C}{>{\centering\arraybackslash}p{3cm}}
\newcolumntype{Y}{>{\centering\arraybackslash}X}

\usepackage{amsmath} 		
\usepackage{amssymb}
\usepackage{mathtools} 		
\usepackage{amsfonts}       
\usepackage{nicefrac}       

\usepackage[T1]{fontenc} 	
\usepackage{xspace} 		
\usepackage{microtype} 		
\usepackage{url} 			
\usepackage{enumitem}
\usepackage{siunitx}

\usepackage{hyperref}       
\definecolor{mydarkblue}{rgb}{0,0.08,0.45} 
\hypersetup{ %
	pdftitle={},
	pdfauthor={},
	pdfsubject={Proceedings of the International Conference on Machine Learning 2021},
	pdfkeywords={},
	pdfborder=0 0 0,
	pdfpagemode=UseNone,
	colorlinks=true,
	linkcolor=mydarkblue,
	citecolor=mydarkblue,
	filecolor=mydarkblue,
	urlcolor=mydarkblue,
	pdfview=FitH}

\usepackage{cleveref}		
\crefformat{footnote}{#2\footnotemark[#1]#3} %

\newcommand*{\eg}{e.g.\@\xspace}
\newcommand*{\ie}{i.e.\@\xspace}
\newcommand*{\cf}{cf.\@\xspace}

\makeatletter\newcommand*{\etc}{%
	\@ifnextchar{.}%
	{etc}%
	{etc.\@\xspace}%
}
\makeatother

\newcommand{\nopts}{$15$\xspace}
\newcommand{\noptstext}{fifteen\xspace}
\newcommand{\nruns}{\mbox{$53$,$760$}\xspace}
\newcommand{\nrunsapprox}{\mbox{$50$,$000$}\xspace}
\newcommand{\nrunsapproxbold}{\mbox{$\mathbf{50}$,$\mathbf{000}$}\xspace}

\newcommand{\nconfigs}{\mbox{$1$,$920$}\xspace}
\newcommand{\ntests}{\mbox{$128$}\xspace}
\newcommand{\nreruns}{\mbox{ten}\xspace}

\newcommand{\adabelief}{\textsc{\mbox{AdaBelief}}\xspace}
\newcommand{\amsbound}{\textsc{\mbox{AMSBound}}\xspace}
\newcommand{\adabound}{\textsc{\mbox{AdaBound}}\xspace}
\newcommand{\adagrad}{\textsc{\mbox{Adagrad}}\xspace}
\newcommand{\adam}{\textsc{\mbox{Adam}}\xspace}
\newcommand{\momentum}{\textsc{\mbox{Momentum}}\xspace}
\newcommand{\sgd}{\textsc{\mbox{SGD}}\xspace}
\newcommand{\rmsprop}{\textsc{\mbox{RMSProp}}\xspace}
\newcommand{\amsgrad}{\textsc{\mbox{AMSGrad}}\xspace}
\newcommand{\adadelta}{\textsc{\mbox{Adadelta}}\xspace}
\newcommand{\nag}{\textsc{\mbox{NAG}}\xspace}
\newcommand{\nadam}{\textsc{\mbox{Nadam}}\xspace}
\newcommand{\radam}{\textsc{\mbox{Radam}}\xspace}
\newcommand{\lamom}{\textsc{\mbox{LA(Mom.)}}\xspace}
\newcommand{\laradam}{\textsc{\mbox{LA(RAdam)}}\xspace}

\newcommand{\otheropt}{\textsc{\mbox{Other}}\xspace}
\newcommand{\lookaheadopt}{\textsc{\mbox{Lookahead}}\xspace}

\newcommand{\deepobs}{\textsc{DeepOBS}\xspace}
\newcommand{\taskset}{\textsc{TaskSet}\xspace}
\newcommand{\rnn}{\textsc{RNN}\xspace}
\newcommand{\allcnnc}{\textsc{All-CNN-C}\xspace}

\newcolumntype{H}{>{\setbox0=\hbox\bgroup}c<{\egroup}@{}}

\definecolor{TUred}{RGB}{141,45,57}
\definecolor{TUdark}{RGB}{55,65,74}
\definecolor{TUgold}{RGB}{174,159,109}
\definecolor{TUgray}{RGB}{175,179,183}
\definecolor{ERC_ora}{RGB}{233,93,15}

\definecolor{sns_blue}{HTML}{1f76b4}
\definecolor{sns_blue_shaded}{HTML}{d2e3f0}
\definecolor{sns_ora}{HTML}{ff7e0e}
\definecolor{sns_ora_shaded}{HTML}{ffcea3}

\definecolor{adabelief}{HTML}{63FFA4}
\definecolor{amsbound}{HTML}{FFFF00}
\definecolor{amsgrad}{HTML}{A63D08}
\definecolor{adabound}{HTML}{FF8C00}
\definecolor{adadelta}{HTML}{735859}
\definecolor{adagrad}{HTML}{F3CC9A}
\definecolor{adam}{HTML}{F90004}
\definecolor{lookaheadmomentum}{HTML}{008B8B}
\definecolor{lookaheadradam}{HTML}{35CC38}
\definecolor{momentum}{HTML}{00CBFF}
\definecolor{nag}{HTML}{991bf5}
\definecolor{nadam}{HTML}{8E8E05}
\definecolor{radam}{HTML}{FF00FF}
\definecolor{rmsprop}{HTML}{000000}
\definecolor{sgd}{HTML}{1E3CFF}
\definecolor{otheropt}{HTML}{B2B2B2}

\DeclareSIUnit{\sec}{sec}
\DeclareSIUnit{\hour}{h}

\newlength\figureheight
\newlength\figurewidth   
\DeclareRobustCommand{\colordot}[1]{%
	\tikz[baseline=(a.south)]{\node[circle, scale=0.75,color=white, fill=#1] (a) {};}
}

\icmltitlerunning{Descending through a Crowded Valley}

\begin{document}

\twocolumn[
\icmltitle{Descending through a Crowded Valley ---\\
	Benchmarking Deep Learning Optimizers}

\icmlsetsymbol{equal}{*}

\begin{icmlauthorlist}
\icmlauthor{Robin M. Schmidt}{equal,tue}
\icmlauthor{Frank Schneider}{equal,tue}
\icmlauthor{Philipp Hennig}{tue,mpi}
\end{icmlauthorlist}

\icmlaffiliation{tue}{Methods of Machine Learning, University of Tübingen, Tübingen, Germany}
\icmlaffiliation{mpi}{Max Planck Institute for Intelligent Systems, Tübingen, Germany}

\icmlcorrespondingauthor{Robin M. Schmidt}{robin.schmidt.97@web.de}
\icmlcorrespondingauthor{Frank Schneider}{f.schneider@uni-tuebingen.de}

\icmlkeywords{Machine Learning, Deep Learning, Optimization, Optimizers, Benchmark, Benchmarking, Comparison}

\vskip 0.3in
]

\printAffiliationsAndNotice{\icmlEqualContribution} 


\begin{abstract}
	Choosing the optimizer is considered to be among the most crucial design decisions in deep learning, and it is not an easy one.
	The growing literature now lists hundreds of optimization methods.
	In the absence of clear theoretical guidance and conclusive empirical evidence, the decision is often made based on anecdotes. 
	In this work, we aim to replace these anecdotes, if not with a conclusive ranking, then at least with evidence-backed heuristics.
	To do so, we perform an extensive, standardized benchmark of \noptstext particularly popular deep learning optimizers while giving a concise overview of the wide range of possible choices.
	Analyzing more than \nrunsapprox individual runs, we contribute the following three points:
	(i) Optimizer performance varies greatly across tasks.
	(ii) We observe that evaluating multiple optimizers with default parameters works approximately as well as tuning the hyperparameters of a single, fixed optimizer.
	(iii) While we cannot discern an optimization method clearly dominating across all tested tasks, we identify a significantly reduced subset of specific optimizers and parameter choices that generally lead to competitive results in our experiments:
	\adam remains a strong contender, with newer methods failing to significantly and consistently outperform it.
	Our open-sourced results\footnote{\label{footnote:url}\url{https://github.com/SirRob1997/Crowded-Valley---Results}} are available as challenging and well-tuned baselines for more meaningful evaluations of novel optimization methods without requiring any further computational efforts.
\end{abstract}


\section{Introduction} \label{sec:Introduction}

\begin{figure}[!t]
	\centering
	\includegraphics[width=\columnwidth]{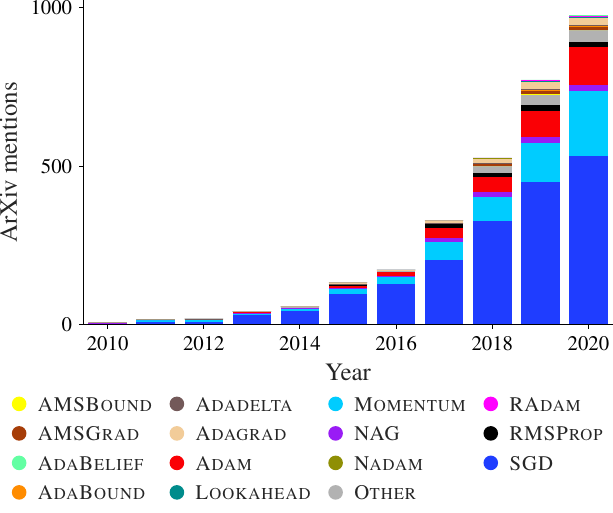}
	\caption{Number of times ArXiv titles and abstracts mention specific optimizer per year. All non-selected optimizers from \Cref{tab:Optimizers} in the appendix are grouped into \textit{Other}. This figure illustrates not only the expected increase in both methods and mentions, but also that our selection covers the most popular methods. In $2020$, the excluded methods accounted for $<4\,\%$ of the mentions (see \Cref{fig:arxiv_normalized}).}
	\label{fig:arxiv}
\end{figure}

Large-scale stochastic optimization drives a wide variety of machine learning tasks.
Because choosing the right optimization method and effectively tuning its hyperparameters heavily influences the training speed and final performance of the learned model, it is an important, every-day challenge to practitioners. It is probably the task that requires the most time and resources in many applications. Hence, stochastic optimization has been a focal point of research, engendering an ever-growing list of methods (\cf \Cref{fig:arxiv}), many of them targeted at deep learning. 
The hypothetical machine learning practitioner who is able to keep up with the literature now has the choice among hundreds of methods (see \Cref{tab:Optimizers} in the appendix), each with their own set of tunable hyperparameters, when deciding how to train a model.

There is limited theoretical analysis that clearly favors one of these choices over the others.
Some authors have offered empirical comparisons on comparably small sets of popular methods \citep[e.g.][]{Wilson2017, Choi2019, Sivaprasad2020}; but for most optimizers, the only empirical evaluation is offered by the original work introducing the method. 
Many practitioners and researchers, meanwhile, rely on personal and anecdotal experience, and informal discussion with colleagues or on social media.
The result is an often unclear, ever-changing ``state of the art'' occasionally driven by hype.
The key obstacle for an objective benchmark is the combinatorial cost of such an endeavor posed by comparing a large number of methods on a large number of problems, with the high resource and time cost of tuning each method's parameters and repeating each (stochastic) experiment repeatedly for fidelity.

We conduct a large-scale benchmark of optimizers to ground the ongoing debate about deep learning optimizers on empirical evidence, and to help understand how the choice of optimization methods and hyperparameters influences the training performance.
Specifically, we examine whether recently proposed methods show an improved performance compared to more established methods such as \sgd or \adam. Additionally, we assess whether there exist optimization methods with well-working default hyperparameters that are able to keep up with tuned optimizers.
To this end, we evaluate \noptstext optimization methods, selected for their perceived popularity, on a range of representative deep learning problems (see \Cref{fig:ParallelCoordinates}) drawing conclusions from tens of thousands of individual training runs.

Right up front, we want to state that it is impossible to include all optimizers (see \Cref{tab:Optimizers} in the appendix), and to satisfy any and all expectations readers may have on tuning, initialization, or the choice of problems---not least because everyone has different expectations in this regard.
In our \emph{personal opinion}, what is needed is an empirical comparison by a third party not involved in the original works. As the target audience of our work, we assume a careful practitioner who does not have access to near-limitless resources, nor to a broad range of personal experiences.
As such, the core contributions of our work are:

\paragraph{1. Assessing the progress in deep learning optimization.} 
A literature review provides a compact but extensive list of recent advances in stochastic optimization.
We identify more than a hundred optimization methods (see \Cref{tab:Optimizers} in the appendix) and more than $20$ families of hyperparameter schedules (see \Cref{tab:LRS} in the appendix) proposed for deep learning.
We conduct a large-scale optimizer benchmark, specifically focusing on problems arising in deep learning.
We evaluate \noptstext optimizers on eight deep learning problems using four different schedules, tuning over dozens of hyperparameter settings.
To our knowledge, this is the most comprehensive empirical evaluation of deep learning optimizers to date (see \Cref{sec:RelatedWork} on related work).

\paragraph{2. Insights from more than \nrunsapproxbold optimization runs.}
Our empirical experiments indicate that an optimizer's performance highly depends on the problem (see \Cref{fig:ParallelCoordinates}).
But some high-level trends emerge, too:
(1) Evaluating multiple optimizers with default hyperparameters works approximately as well as tuning the hyperparameters for a fixed optimizer.
(2) Using an additional untuned learning rate schedule helps on average, but its effect varies greatly depending on the optimizer and the problem.
(3) While there is no optimizer that clearly dominates across all tested workloads, some of the methods we tested exhibited highly variable performance. Others demonstrated decent performance consistently. We deliberately abstain from recommending a single one among them, because we could not find a clear winner with statistical confidence.

\paragraph{3. An open-source baseline for future optimizer benchmarks and meta-learning approaches.}
Our results are available in an open and easily accessible form (see Footnote \ref{footnote:url} on Page 1).
This data set contains \nruns unique runs, each consisting of thousands of individual data points, such as the mini-batch training losses of every iteration or epoch-wise performance measures, for example, the loss on the full validation set or test set accuracy.
These results can be used as competitive and well-tuned baselines for future benchmarks of new optimizers, drastically reducing the amount of computational budget required for a meaningful optimizer comparison.
This collection of training curves could also be used for meta-learning novel optimization methods, hyperparameter search strategies, or hyperparameter adaptation strategies.
To encourage researches to contribute to this collection, we made our baselines easily expandable. \cref{footnote:url}

The high-level result of our benchmark is, perhaps expectedly, \emph{not} a clear winner.
Instead, our comparison shows that, while some optimizers are frequently decent, they also generally perform similarly, often switching their positions in the ranking.
This result is reminiscent, albeit not formally a rigorous result of the No Free Lunch Theorem \citep{DolpertM97}.
A key insight of our comparison is that a practitioner with a new deep learning task can expect to do about \emph{equally well} by taking almost any method from our benchmark and \emph{tuning} it, as they would by investing the same computational resources into running a set of optimizers with their \emph{default} settings and picking the winner.

Possibly the most important takeaway from our comparison is that ``there are now enough optimizers''.
Methods research in stochastic optimization should focus on \emph{significant} (conceptual, functional, performance) improvements---such as methods specifically suited for certain problem types, inner-loop parameter tuning or structurally novel methods.
We make this claim not to discourage research but, quite on the contrary, to offer a motivation for more meaningful, non-incremental research.

\subsection{Related work} \label{sec:RelatedWork}

Following the rapid increase in publications on optimizers, \emph{benchmarking} these methods for the application in deep learning has only recently attracted significant interest.
\citet{Schneider2019} introduced a benchmarking framework called \deepobs, which includes a wide range of realistic deep learning problems together with standardized procedures for evaluating optimizers.
\citet{Metz2020} presented \taskset, another collection of optimization problems focusing on smaller but more numerous problems.
For the empirical analysis presented here, we use \deepobs as it provides optimization problems closer to real-world deep learning tasks.
In contrast to our evaluation of \emph{existing} methods, \taskset and its analysis focuses on meta-learning \emph{new} optimizers or hyperparameters.

Both \citet{Choi2019} and \citet{Sivaprasad2020} analyzed specific aspects of the benchmarking process.
\citet{Sivaprasad2020} used \deepobs to illustrate that the relative 
performance of an optimizer depends significantly on the used hyperparameter 
tuning budget.
The analysis by \citet{Choi2019} supports this point, stating that ``the hyperparameter search space may be the single most important factor explaining the rankings''.
They further stress a hierarchy among optimizers, demonstrating that, given sufficient hyperparameter tuning, more general optimizers can never be outperformed by special cases.
In their study, however, they manually defined a hyperparameter search space \textit{per optimizer and problem} basing it either on prior published results, prior experiences, or pre-tuning trials.

Here, we instead aim to identify well-performing general-purpose optimizers for deep learning, especially when there is no prior knowledge about well-working hyperparameter values for each specific problem. We further elaborate on the influence of our chosen hyperparameter search strategy in \Cref{sec:Limitations} discussing the limitations of our empirical study.

Our work is also related to empirical generalization studies of adaptive methods, such as that of \citet{Wilson2017} which sparked an extensive discussion whether adaptive methods (\eg \adam) tend to generalize worse than standard first-order methods (\ie \sgd). By focusing on and reporting the \emph{test set accuracy} we implicitly include the generalization capabilities of different optimizers in our benchmark results, an important characteristic of deep learning optimization.


\section{Benchmarking process} \label{sec:Benchmark}

Any benchmarking effort requires tricky decisions on the experimental setup that influence the results.
Evaluating on a specific task or picking a certain tuning budget may favor or disadvantage certain methods \citep{Sivaprasad2020}.
It is impossible to avoid these decisions or to cover all possible choices. 
Aiming for generality, we evaluate the performance on eight diverse real-world deep learning problems from different disciplines (\Cref{sec:Testproblems}).
From a collection of more than a hundred deep learning optimizers (\Cref{tab:Optimizers} in the appendix) we select \noptstext of the most popular choices (see \Cref{fig:arxiv}) for this benchmark (\Cref{sec:Optimizers}).
For each problem and optimizer we evaluate all possible combinations of four different tuning budgets (\Cref{sec:Tuning}) and four selected learning rate schedules (\Cref{sec:Schedules}), covering the following combinatorial space:
\begin{align}
	\begin{aligned} & \text{\textbf{Problem}} \\
	\phantom{
		\left.
		\begin{aligned}
		. \\ . \\ . \\ .
		\end{aligned} 
		\right.
	}
	& \left\{\begin{aligned} 
	& \text{P1} \\ & \text{P2} \\ & \dots \\ & \text{P8}
	\end{aligned}
	\right\}_{8}
	\phantom{
		\left.
		\begin{aligned}
		. \\ . \\ . \\ .
		\end{aligned} 
		\right.
	}
	\hspace*{-1em}
	\times
	\hspace*{-1em}
	\end{aligned}
	\begin{aligned} & \, \text{\textbf{Optimizer}} \\
	\phantom{
		\left.
		\begin{aligned}
		. \\ . \\ . \\ .
		\end{aligned} 
		\right.
	} 
	& \left\{\begin{aligned}
	& \text{\adam} \\ & \text{\nag} \\ & \dots \\ & \text{\sgd}
	\end{aligned}
	\right\}_{\text{\nopts}}
	\phantom{
		\left.
		\begin{aligned}
		. \\ . \\ . \\ .
		\end{aligned} 
		\right.
	}
	\hspace*{-1.2em}
	\times
	\hspace*{-.5em}
	\end{aligned}
	\begin{aligned} & \quad \text{\textbf{Tuning}} \\
	\phantom{
		\left.
			\begin{aligned}
			. \\ . \\ . \\ .
			\end{aligned} 
			\right.
	}
	& \left\{\begin{aligned}
	& \text{one-shot} \\ & \text{small} \\ & \text{medium} \\ & \text{large}
	\end{aligned}
	\right\}_{4}
	\phantom{
		\left.
		\begin{aligned}
		. \\ . \\ . \\ .
		\end{aligned} 
		\right.
	}
	\hspace*{-0.8em}
	\times
	\hspace*{-0.4em}
	\end{aligned}
	\begin{aligned} & \quad \text{\textbf{Schedule}} \\
	\phantom{
		\left.
		\begin{aligned}
		. \\ . \\ . \\ .
		\end{aligned} 
		\right.
	}
	& \left\{\begin{aligned}
	& \text{constant} \\ & \text{cosine} \\ & \text{cosine wr} \\ & \text{trapez.}
	\end{aligned}
	\right\}_{4}
	.
	\end{aligned}
	\notag
\end{align}

Combining those options results in \nconfigs configurations, where each of the \noptstext optimizers is evaluated in \ntests settings (\ie on \emph{eight} problems, with \emph{four} budgets and \emph{four} schedules).
Including hyperparameter search and estimating the confidence interval, our main benchmark consists of \nruns unique training curves.

\subsection{Problems} \label{sec:Testproblems}

We consider the eight optimization tasks summarized in \Cref{tab:TestProblems}, available as the ``small'' (P1--P4) and ``large'' (P5--P8) problem sets in \deepobs.
A detailed description of these problems, including architectures, training parameters, etc.~can be found in the work of \citet{Schneider2019}.\footnote{All experiments were performed using version \texttt{1.2.0-beta} of \deepobs and TensorFlow version \texttt{1.15} \citep{Abadi2015}.}
\begin{table*}
	\caption{Summary of problems used in our experiments. Exact model configurations can be found in \citet{Schneider2019}.}
	\label{tab:TestProblems}
	\centering
	\begin{tabularx}{0.99\textwidth}{lllllrrr}
		\toprule
		& \textbf{Data set}            & \textbf{Model}   & \textbf{Task}                     & \textbf{Metric}        & \textbf{Batch} & \textbf{Budget} & \textbf{Approx.} \\
		&                              &                  &                                   & 		               & \textbf{size}  & \textit{in epochs}             & \textbf{run time}\footnotemark   \\ \midrule
		\textbf{P1} & Artificial & Noisy quadratic  & \mbox{Minimization} & Loss     & 128            & 100       & $<$ 1 \si{\minute} \\
		\textbf{P2} & MNIST                        & VAE              & Generative               & Loss     & 64             & 50   &    10 \si{\minute} \\
		\textbf{P3} & Fashion-MNIST                       & Simple CNN: \textit{2c2d}       & \mbox{Classification}       & Accuracy & 128            & 100    &  20 \si{\minute} \\
		\textbf{P4} & CIFAR-10                     & Simple CNN: \textit{3c3d}       & \mbox{Classification}       & Accuracy & 128            & 100      & 35 \si{\minute} \\
		\rule{0pt}{3ex}\textbf{P5} & Fashion-MNIST                & VAE              & Generative              & Loss     & 64             & 100      &  20 \si{\minute} \\
		\textbf{P6} & CIFAR-100                    & \textit{All-CNN-C}        & \mbox{Classification}       & Accuracy & 256            & 350     &  4 \si{\hour} 00 \si{\minute} \\
		\textbf{P7} & SVHN                         & \textit{Wide ResNet 16-4} & \mbox{Classification}       & Accuracy & 128            & 160     &  3 \si{\hour} 30 \si{\minute}\\
		\textbf{P8} & War and Peace                & RNN              & Character Prediction                & Accuracy & 50             & 200     &  5 \si{\hour} 30 \si{\minute} \\ \bottomrule
	\end{tabularx}
\end{table*}

\deepobs provides several performance metrics, including the training and test loss, and the validation accuracy.
While these are all relevant, any comparative evaluation of optimizers requires picking only a few, if not just one particular performance metric.
For our analysis (\Cref{sec:Results}), we focus on the final test accuracy (or the final test loss, if accuracy is not defined for this problem).
This metric captures the optimizer's ability to generalize and is thus highly relevant for practical use.
Our publicly released results include all metrics for completeness.
An example of training loss performance is shown in \Cref{fig:appendix_pc_train_loss} in the appendix.
Accordingly, the tuning (\Cref{sec:Tuning}) is done with respect to the validation metric.
We discuss possible limitations resulting from these choices in \Cref{sec:Limitations}. 

\subsection{Optimizer} \label{sec:Optimizers}

In \Cref{tab:Optimizers} in the appendix we collect over a hundred optimization methods 
introduced for or used in deep learning. 
This list was collected by multiple researchers trying to keep up with the field over recent years.
It is thus necessarily incomplete, although it may well represent one of the most exhaustive of such collections. 
Even this incomplete list, though, contains too many entries for a benchmark with the degrees of freedom collected above. This is a serious problem for research: Even an author of a new optimizer, let alone a practitioner, cannot be expected to compare their work with every possible previous method.

We thus select a subset of \noptstext optimizers, which we consider to be currently the most popular choices in the community (see \Cref{tab:selected_optimizers} in the appendix).
These do not necessarily reflect the ``best'' methods, but are either 
commonly used by practitioners and researchers, or have recently generated attention.
Our selection is focused on first-order optimization methods, both due to their 
prevalence for non-convex optimization problems in deep learning as 
well as to simplify the comparison. Whether there is a significant difference 
between these optimizers or if they are inherently redundant is one of the 
questions this work investigates.

\footnotetext{All approximations are for \adam on a Tesla K80 GPU.}

Our list focuses on optimizers over optimization techniques, although the line between the two is admittedly blurry.
Techniques such as averaging weights \citep[\eg]{Izmailov2018} or 
ensemble methods \citep[\eg]{Garipov2018} have been shown to be simple but 
effective at improving the optimization performance.
Those methods, however, can be applied to all methods in our lists, similar to 
regularization techniques, learning rate schedules, or tuning method.
We have, therefore, decided to omit them from \Cref{tab:Optimizers}.

\subsection{Tuning} \label{sec:Tuning}

\paragraph{Budget}
Optimization methods for deep learning regularly expose hyperparameters to the user.
The user either relies on the default suggestion or sets them using experience from previous experiments, or using additional tuning runs to find the best-performing setting.
All optimizers in our benchmark have tunable hyperparameters, and we consider four different \emph{tuning budgets}.

The first budget consists of just a single run.
This \emph{one-shot} budget uses the default values proposed by the original authors, where available (\Cref{tab:selected_optimizers} in the appendix lists the default parameters).
If an optimizer performs well in this setting, this has great practical value, as it drastically reduces the computational resources required for successful training.

The \emph{small}, \emph{medium} and \emph{large} budgets consist of $25$, $50$, and $75$ tuning runs, where the parameters for each setting are sampled using random search.
Tuning runs for the small and medium budget were sampled using the distributions defined in \Cref{tab:selected_optimizers}.
The additional $25$ tuning runs of the large budget, however, were sampled using refined bounds:
For each combination of optimizer, problem, and learning rate schedule we use the same distribution as before, but restrict the search space, to contain all hyperparameter configurations of the top-performing $20\,\%$ tuning runs from the medium budget are included.

We use a single seed for tuning, but for all configurations repeat the best setting with \nreruns different seeds.
This allows us to report standard deviations in addition to means, assessing stability.
Our tuning process can sometimes pick ``lucky'' seeds, which do not perform well when averaging over multiple runs.
This is arguably a feature rather than a bug, since it reflects practical reality.
If an optimizer is so unstable that ten random seeds are required for tuning---which would render this benchmark practically infeasible---it would be impractical for the end-user as well.
Our scoring naturally prefers stable optimizers.
\Cref{sec:FailingSeeds,sec:retuning} provide further analysis of these cases and the general stability of our benchmark, showing amongst other things that failing seeds occur in less than $0.5\,\%$ of the tuning runs.

\paragraph{Tuning method}
We tune parameters by random search without early-stopping for the small, medium and large budget.
Random search is a popular choice due to its efficiency over grid search \citep{BergstraB12} and its ease of implementation and parallelization compared to Bayesian optimization (further discussed in \Cref{sec:Limitations}).
A minor complication of random search is that the sampling distribution affects the performance of the optimizer.
The sampling distribution acts as a prior over good parameter settings, and bad priors consequently ruin performance.
We followed the valid interval and intuition provided by the optimizers' authors for relevant hyperparameters.
The resulting sampling distributions can be found in \Cref{tab:selected_optimizers} in the appendix.
Even though a hyperparameter might have a similar name in different optimization methods (\eg~learning rate $\alpha$), its appropriate search space can differ.
However, without grounded heuristics guiding the practitioner on how the hyperparameters differ between optimizers, the most straightforward approach for any user is to use the same search space.
Therefore, in case there was no prior knowledge provided in the cited work we chose similar distributions for similar hyperparameters across different optimizers.

\paragraph{What should be considered a hyperparameter?}
There is a fuzzy boundary between (tunable) hyperparameters and (fixed) design parameters.
A recently contentious example is the $\varepsilon$ in adaptive methods like \adam.
It was originally introduced as a safeguard against division by zero, but has recently been re-interpreted as a problem-dependent hyperparameter (see \citet{Choi2019} for a discussion).
Under this view, one can actually consider several optimizers called \adam: From an easy-to-tune but potentially limited $\adam_{\alpha}$, only tuning the learning rate, to the tricky-to-tune but all-powerful $\adam_{\alpha, \beta_1, \beta_2, \varepsilon}$, which can approximate \sgd in its hyperparameter space. While both share the update rule, we consider them to be different optimizers. For each update rule, we selected one popular choice of tunable parameters, \eg $\adam_{\alpha, \beta_1, \beta_2}$ (see \Cref{tab:selected_optimizers}).

\subsection{Schedules} \label{sec:Schedules}

The literature on learning rate schedules is now nearly as extensive as that on optimizers (see \Cref{tab:LRS} in the appendix).
\textit{In theory}, schedules can be applied to all hyperparameters of an optimization method but to keep our configuration space feasible, we only apply schedules to the learning rate, by far the most popular practical choice \citep{Goodfellow2016, zhang2020dive}.
We choose four different learning rate schedules, trying to cover all major 
types of schedules (see \Cref{sec:schedules}):
\begin{itemize}[noitemsep, topsep=0pt] 
	\item A \emph{constant} learning rate;
	\item A \emph{cosine decay} \citep{Loshchilov2017} as an example of a smooth decay;
	\item A \emph{cosine with warm restarts} schedule \citep{Loshchilov2017} as 
	a cyclical schedule;
	\item A \emph{trapezoidal} schedule \citep{Xing2018} from the warm-up schedules introduced in \citet{goyal2017accurate}.
\end{itemize}


\section{Results} \label{sec:Results}

\begin{figure*}[htb]
	\centering
	\includegraphics[width=\textwidth]{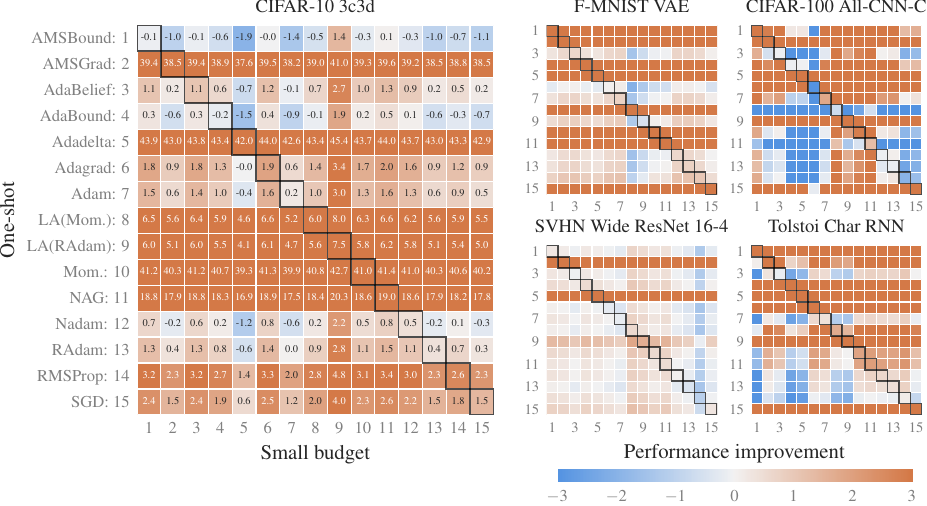}
	\caption{The test set performance improvement after switching from any untuned optimizer ($y$-axis, \textit{one-shot}) to any tuned optimizer ($x$-axis, \textit{small budget}) as an average over $10$ random seeds for the \emph{constant} schedule. For example, the bottom left cell of the largest matrix indicates that the tuned version of \amsbound (1) reaches a $2.4\,\%$ higher test accuracy than untuned \sgd (15). We discuss the unintuitive occurrence of negative diagonal entries in \Cref{sec:heatmaps_appendix}. The colormap is capped at $\pm 3$ to improve presentation, although larger values occur.}
	\label{fig:Heatmap}
\end{figure*}

\paragraph{How well do optimizers work out-of-the-box?}
By comparing each optimizer's one-shot results against the tuned versions of all \noptstext optimizers, we can construct a $\nopts \times \nopts$ matrix of performance gains.
\Cref{fig:Heatmap} illustrates this on five problems showing improvements by a positive sign and an orange cell.
Detailed plots for all problems are in \Cref{fig:Heatmap__o__s__0,fig:Heatmap__o__s__1} in the appendix.
For example, the bottom left cell of the largest matrix in \Cref{fig:Heatmap} shows that \amsbound \textit{(1)} tuned using a small budget performs $2.4\%$ better than \sgd \textit{(15)} with default parameters on this specific problem.

An \textbf{orange row} in \Cref{fig:Heatmap} indicates that an optimizer's default setting is performing badly, since it can be beaten by any well-tuned competitor.
We can observe badly-performing default settings for \momentum, \nag and \sgd, advocating the intuition that non-adaptive optimization methods require more tuning, but also for \amsgrad and \adadelta.
This is just a statement about the default parameters suggested by the authors or the popular frameworks; well-working default parameters might well exist for those methods.
Conversely, a \textbf{white \& blue row} signals a well-performing default setting, since even tuned optimizers do not significantly outperform it.
\adam, \nadam and \radam, as well as \amsbound, \adabound and \adabelief all have white or blue rows on several (but not all!) problems, supporting the rule of thumb that adaptive methods have well-working default parameters. 
Conversely, \textbf{orange} (or \textbf{blue}) \textbf{columns} highlight optimizers that, when tuned, perform better (or worse) than all untuned optimization methods.
We do not observe such columns consistently across tasks. This supports the conclusion that an optimizer's performance is heavily problem-dependent and that there is no single \textit{best} optimizer across workloads.

\Cref{fig:Heatmap__o__s__0,fig:Heatmap__o__s__1,fig:Heatmap__o__l__0__TUNING_SEED,fig:Heatmap__o__l__1__TUNING_SEED} in the appendix suggest an interesting alternative approach for machine learning practitioners: Instead of picking a single optimizer and tuning its hyperparameters extensively, trying out a few optimizers with default settings and picking the best one yields competitive results with less computational and tuning choice efforts.
However, this might not hold for more complicated, structurally different tasks such as GANs \citep{Goodfellow2014} or Transformer models \cite{Vaswani2017}.
The similarity of those two approaches might be due to the fact that optimizers have implicit learning rate schedules \citep{Agarwal2020} and trying out different optimizers is similar to trying out different (well-tested) schedules.

\begin{figure*}[htb]
	\centering
	\includegraphics[width=\textwidth]{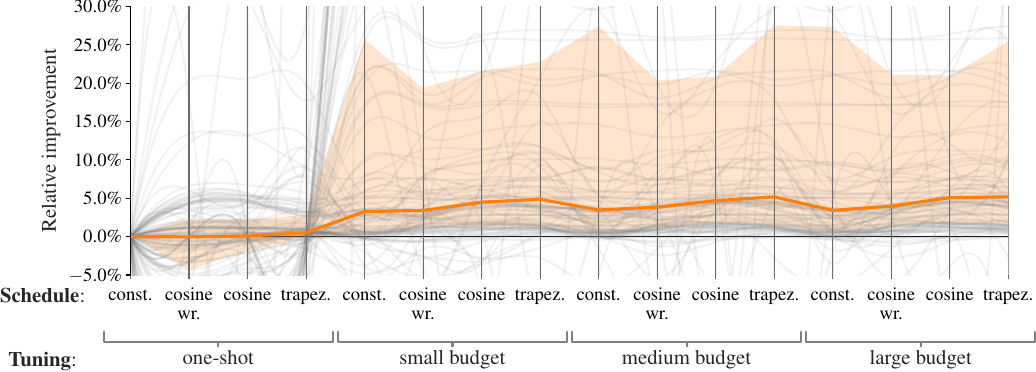}
	\caption{Lines in gray (\textcolor{TUgray}{\textbf{---}}, smoothed by cubic splines for visual guidance only) show the relative improvement for a certain tuning budget and schedule (compared to the \textit{one-shot} tuning without schedule) for all \noptstext optimizers on all eight problems. The median over all lines is plotted in orange (\textcolor{sns_ora}{\textbf{---}}) with the shaded area (\textcolor{sns_ora_shaded}{\ding{122}}) indicating the area between the 25th and 75th percentile. With an increased budget and a schedule, one can expect a performance increase \emph{on average} (orange lines), but not automatically for individual settings (\ie gray lines can be unaffected or even decrease).}
	\label{fig:TuningPlot}
\end{figure*}

\paragraph{How much do tuning and schedules help?}
We consider the final performance achieved by varying budgets and schedules to quantify the usefulness of tuning and applying parameter-free schedules (\Cref{fig:TuningPlot}).
While there is no clear trend for any individual setting (gray lines), in the median we observe that increasing the budget improves performance, albeit with diminishing returns.
For example, using the medium budget without any schedule leads to a median 
relative improvement of roughly $3.4\,\%$ compared to the 
default parameters (without schedule).

Applying an untuned schedule improves median performance as well.
For example, the large tuning budget coupled with a trapezoidal learning rate schedule leads to a median relative improvement of the performance of roughly $5.2\,\%$ compared to the default parameters.
However, while these trends hold in the median, their individual effect varies wildly among optimizers and problems, as is apparent from the noisy structure of the individual lines shown in \Cref{fig:TuningPlot}.

\begin{figure*}[htb]
	\centering
	\includegraphics[width=\textwidth]{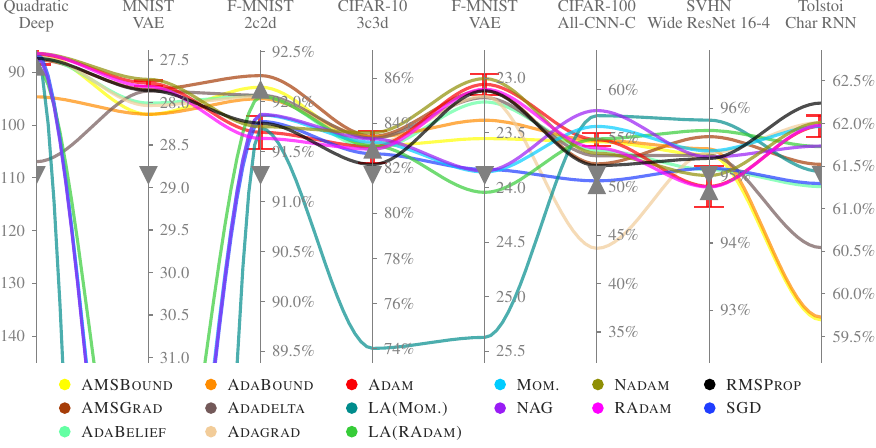}
	\caption{Mean test set performance over $10$ random seeds of all tested optimizers on all eight optimization problems using the \emph{large budget} for tuning and \emph{no learning rate schedule}. One standard deviation for the \emph{tuned} \adam optimizer is shown with a red error bar (\textcolor{adam}{\textbf{I}}; error bars for other methods omitted for legibility). The performance of \emph{untuned} \adam (\textcolor{TUgray}{\ding{116}}) and \adabound (\textcolor{TUgray}{\ding{115}}) are marked for reference. The upper bound of each axis represents the best performance achieved in the benchmark, while the lower bound is chosen in relation to the performance of \adam with default parameters. Tabular version available in the Appendix as \Cref{tab:app_tabular_large_none}.}
	\label{fig:ParallelCoordinates}
\end{figure*}

\paragraph{Which optimizers work well after tuning?}
\Cref{fig:ParallelCoordinates} compares the optimizers' performance across all eight problems. There is no single optimizer that dominates its competitors across all tasks. Nevertheless, some optimizers generally perform well, while others can vary greatly in their behavior, most notably performing poorly on VAEs. 
Further supporting the hypothesis of previous sections, we note that taking the best out of a small set of \emph{untuned} optimizers --- for example, \adam and \adabound~--- frequently results in competitive performance. Except for the two VAE problems, the best of those two untuned optimizers generally falls within the distribution of the well-tuned methods.
Combining these runs with a \emph{tuned} version of \adam (or a variant thereof) provides stable and slightly improved results across many problems in our benchmark.
To further increase the performance, our results suggest trying a different optimizer next, such as \rmsprop or \nag.
Across multiple budgets and schedules, both optimizers show a consistently good performance on the \rnn and \allcnnc model, respectively.

Nevertheless, achieving (or getting close to) the absolute best performance still requires testing numerous optimizers.
Which optimizer wins in the end is problem-dependent: optimizers that achieve top scores on one problem can perform poorly on other tasks.
We note in passing that the individual optimizer rankings changes when considering \eg a smaller budget or an additional learning rate schedule (see \Cref{fig:appendix_pc_sb,fig:appendix_pc_ltr,fig:appendix_pc_cos} in the appendix). However, the overall trends described here are consistent. 

The idea that optimizers perform consistently better or worse for specific model architectures or tasks has been regularly theorized and mentioned in the literature.
Indeed, our results support this hypothesis, with \nag often beating \adam on image classification tasks, and \rmsprop being consistently on top for the natural language modeling task (see \Cref{tab:app_tabular_large_none,tab:app_tabular_small_none,tab:app_tabular_medium_cosine,tab:app_tabular_large_ltr}).
Understanding whether and why certain optimizers favor specific problem types presents an interesting research avenue and might lead to more sophisticated optimizers that utilize the problem characteristics.


\section{Limitations} \label{sec:Limitations}

Any empirical benchmark has constraints and limitations.
Here we highlight some of ours' and characterize the context within which our results should be considered.

\paragraph{Generalization of the results}
By using the problems from \deepobs, which span models and data sets of varying complexity, size, and different domains, we aim for generalization.
Our results are, despite our best efforts, reflective of not just these setups, but also of the chosen training parameters, the software framework, and further unavoidable choices.
The design of our comparisons aims to be close to what an informed practitioner would encounter for a relatively novel problem in practice.
It goes without saying that even a carefully curated range of problems 
cannot cover all challenges of machine learning or even just deep learning.
In particular, our conclusions may not generalize to other workloads 
such as GANs, reinforcement learning, or applications where \eg memory usage is 
crucial.

Similarly, our benchmark does not cover more large-scale problems such as 
ImageNet \citep{Deng2009} or transformer models \citep{Vaswani2017}.
While there is oft-mentioned anecdotal evidence that the characteristics of deep learning problems change for larger models, it would simply be impossible to perform the kind of combinatorial exploration of choices covered in our benchmark, even with significant hardware resources.
The inclusion of larger models would require reducing the number of tested optimizers, schedules or tuning methods and would thus shift the focus of the benchmark.
Studying whether there are systematic differences between different types of deep learning problems presents an interesting avenue for further research.

We do not consider this study the definitive work on benchmarking deep learning 
optimizers, but rather an important and significant step in the right direction.
While our comparison includes many ``dimensions'' of deep learning 
optimization, \eg by considering different problems, tuning budgets, and 
learning rate schedules, there are certainly many more.
To keep the benchmark feasible, we chose to use the fixed $L_2$~regularization and 
batch size that \deepobs suggests for each problem.
We also did not include optimization techniques such as weight averaging or 
ensemble methods as they can be combined with all evaluated optimizers and hence would increase the computational cost further.
Future works could study how these techniques interact with different 
optimization methods.
However, to keep our benchmark feasible, we have selected what we believe to 
be the most important aspects affecting an optimizer comparison.
We hope, that our study lays the groundwork so that other works can build 
on it and analyze these questions.

\paragraph{Influence of the hyperparameter search strategy}
As noted by, \eg, \citet{Choi2019} and \citet{Sivaprasad2020}, the hyperparameter tuning method, its budget, and its search domain, can significantly affect performance. 
By reporting results from four different hyperparameter optimization budgets (including the tuning-free one-shot setting) we try to quantify the effect of tuning. 
We argue that our random search process presents a realistic setting for many but certainly not all deep learning practitioners.
One may criticize our approach as simplistic, but note that more elaborate schemes, in particular Bayesian optimization, would multiply the number of design decisions (kernels, search utilities, priors, etc.) and thus significantly complicate the analysis.

The individual hyperparameter sampling distributions significantly affect the relative rankings of the optimizers.
A poorly chosen search space can make successful tuning next to impossible.
In our benchmark, we use relatively broad initial search spaces, dozens of tuning runs and a refining of those search spaces for the large budget.
Note, though, that the problem of finding appropriate search spaces is inherited by practitioners.
It is arguably an implicit flaw of an optimization method that expects hyperparameter tuning not to come with well-identified search spaces for those parameters and this should thus be reflected in a benchmark.


\section{Conclusion} \label{sec:Conclusion}

Faced with an avalanche of research developing new stochastic optimization methods, 
practitioners are left with the near-impossible task of not just picking a method from this ever-growing list, but also to guess or tune hyperparameters for them, even to continuously tune them during training.
Despite efforts by the community, there is currently no method that clearly dominates the competition.

We have provided an extensive empirical benchmark of optimization methods for deep learning.
It reveals structure in the crowded field of training methods for deep learning:
First, although many methods perform competitively, a subset of methods tends to come up near the top across the spectrum of problems. Despite years of new research by many committed authors, \adam remains a viable (but also not a clearly superior) choice for many problems, with \nag or \rmsprop being interesting alternatives that were able to boost performance on individual problems.
Secondly, tuning helps about as much as trying other optimizers.
Our open and extendable data set allows many, more technical observations, for example, that the stability to re-runs is an often overlooked challenge. 

Perhaps the most important takeaway from our study is hidden in plain sight: the field is in danger of being drowned by noise.
Different optimizers exhibit a surprisingly similar performance distribution compared to a single method that is re-tuned or simply re-run with different random seeds.
It is thus questionable how much insight the development of new methods yields, at least if they are conceptually and functionally close to the existing population. We hope that benchmarks like ours can help the community to move beyond inventing yet another optimizer and to focus on key challenges, such as automatic, inner-loop tuning for truly robust and efficient optimization.
We are releasing our data to allow future authors to ensure that their method contributes to such ends.


\subsubsection*{Acknowledgments}
The authors gratefully acknowledge financial support by the European Research Council through ERC StG Action 757275 / PANAMA; the DFG Cluster of Excellence “Machine Learning - New Perspectives for Science”, EXC 2064/1, project number 390727645; the German Federal Ministry of Education and Research (BMBF) through the Tübingen AI Center (FKZ: 01IS18039A); and funds from the Ministry of Science, Research and Arts of the State of Baden-Württemberg.
Moreover, the authors thank the International Max Planck Research School for Intelligent Systems (IMPRS-IS) for supporting Frank Schneider.
We would like to thank Aaron Bahde for providing his analysis on the robustness to random seeds.
Further, we are grateful to Lukas Balles, Frederik Künstner, and Felix Dangel for, among other things, helping to create the list of optimizers and providing feedback to the manuscript. Lastly, we want to thank Agustinus Kristiadi, Jonathan Wenger, Marius Hobbhahn, and Lukas Tatzel for their additional feedback.


\bibliography{bibliography/references}

\begin{thebibliography}{160}
\providecommand{\natexlab}[1]{#1}
\providecommand{\url}[1]{\texttt{#1}}
\expandafter\ifx\csname urlstyle\endcsname\relax
  \providecommand{\doi}[1]{doi: #1}\else
  \providecommand{\doi}{doi: \begingroup \urlstyle{rm}\Url}\fi

\bibitem[Aitchison(2020)]{aitchison2018bayesian}
Aitchison, L.
\newblock {Bayesian filtering unifies adaptive and non-adaptive neural network
  optimization methods}.
\newblock In \emph{Advances in Neural Information Processing Systems 33,
  NeurIPS}, 2020.

\bibitem[Almeida et~al.(2021)Almeida, Winter, Tang, and Zaremba]{Almeida2021}
Almeida, D., Winter, C., Tang, J., and Zaremba, W.
\newblock {A Generalizable Approach to Learning Optimizers}. \emph{arXiv}\emph{
  preprint:} \href{http://arxiv.org/abs/2106.00958}{{\ttfamily
  \emph{2106.00958}}}, 2021.

\bibitem[Anil et~al.(2019)Anil, Gupta, Koren, and Singer]{Anil2019}
Anil, R., Gupta, V., Koren, T., and Singer, Y.
\newblock {Memory Efficient Adaptive Optimization}.
\newblock In \emph{Advances in Neural Information Processing Systems 32,
  NeurIPS}, 2019.

\bibitem[Anil et~al.(2020)Anil, Gupta, Koren, Regan, and Singer]{Anil2020}
Anil, R., Gupta, V., Koren, T., Regan, K., and Singer, Y.
\newblock {Second Order Optimization Made Practical}. \emph{arXiv}\emph{
  preprint:} \href{http://arxiv.org/abs/2002.09018}{{\ttfamily
  \emph{2002.09018}}}, 2020.

\bibitem[Ayadi \& Turinici(2020)Ayadi and Turinici]{ayadi2020}
Ayadi, I. and Turinici, G.
\newblock {Stochastic Runge-Kutta methods and adaptive SGD-G2 stochastic
  gradient descent}. \emph{arXiv}\emph{ preprint:}
  \href{http://arxiv.org/abs/2002.09304}{{\ttfamily \emph{2002.09304}}}, 2020.

\bibitem[Bae et~al.(2019)Bae, Ryu, and Shin]{bae2019does}
Bae, K., Ryu, H., and Shin, H.
\newblock {Does Adam optimizer keep close to the optimal point?}.
  \emph{arXiv}\emph{ preprint:}
  \href{http://arxiv.org/abs/1911.00289}{{\ttfamily \emph{1911.00289}}}, 2019.

\bibitem[Bahrami \& Zadeh(2021)Bahrami and Zadeh]{Bahrami2021}
Bahrami, D. and Zadeh, S.~P.
\newblock {Gravity Optimizer: a Kinematic Approach on Optimization in Deep
  Learning}. \emph{arXiv}\emph{ preprint:}
  \href{http://arxiv.org/abs/2101.09192}{{\ttfamily \emph{2101.09192}}}, 2021.

\bibitem[Bai \& Zhang(2019)Bai and Zhang]{bai2019bgadam}
Bai, J. and Zhang, J.
\newblock {BGADAM: Boosting based Genetic-Evolutionary ADAM for Convolutional
  Neural Network Optimization}. \emph{arXiv}\emph{ preprint:}
  \href{http://arxiv.org/abs/1908.08015}{{\ttfamily \emph{1908.08015}}}, 2019.

\bibitem[Balles \& Hennig(2018)Balles and Hennig]{Balles2018}
Balles, L. and Hennig, P.
\newblock {Dissecting Adam: The Sign, Magnitude and Variance of Stochastic
  Gradients}.
\newblock In \emph{35th International Conference on Machine Learning, ICML},
  2018.

\bibitem[Bello et~al.(2017)Bello, Zoph, Vasudevan, and Le]{Bello2017}
Bello, I., Zoph, B., Vasudevan, V., and Le, Q.~V.
\newblock {Neural Optimizer Search with Reinforcement Learning}.
\newblock In \emph{34th International Conference on Machine Learning, {ICML}},
  2017.

\bibitem[Bernstein et~al.(2018)Bernstein, Wang, Azizzadenesheli, and
  Anandkumar]{BernsteinWAA18}
Bernstein, J., Wang, Y., Azizzadenesheli, K., and Anandkumar, A.
\newblock {SIGNSGD: Compressed Optimisation for Non-Convex Problems}.
\newblock In \emph{35th International Conference on Machine Learning, {ICML}},
  2018.

\bibitem[Berrada et~al.(2020)Berrada, Zisserman, and
  Kumar]{berrada2019training}
Berrada, L., Zisserman, A., and Kumar, M.~P.
\newblock {Training Neural Networks for and by Interpolation}.
\newblock In \emph{37th International Conference on Machine Learning, {ICML}},
  2020.

\bibitem[Borysenko \& Byshkin(2020)Borysenko and
  Byshkin]{oleks2020coolmomentum}
Borysenko, O. and Byshkin, M.
\newblock {CoolMomentum: A Method for Stochastic Optimization by Langevin
  Dynamics with Simulated Annealing}. \emph{arXiv}\emph{ preprint:}
  \href{http://arxiv.org/abs/2005.14605}{{\ttfamily \emph{2005.14605}}}, 2020.

\bibitem[Botev et~al.(2017)Botev, Ritter, and Barber]{Botev2017}
Botev, A., Ritter, H., and Barber, D.
\newblock {Practical Gauss-Newton Optimisation for Deep Learning}.
\newblock In \emph{34th International Conference on Machine Learning, ICML},
  2017.

\bibitem[Bottou(2012)]{Bottou2012}
Bottou, L.
\newblock Stochastic gradient descent tricks.
\newblock In \emph{Neural networks: Tricks of the trade}. Springer, 2012.

\bibitem[Castera et~al.(2021)Castera, Bolte, Févotte, and
  Pauwels]{Castera2021}
Castera, C., Bolte, J., Févotte, C., and Pauwels, E.
\newblock {Second-order step-size tuning of SGD for non-convex optimization}.
  \emph{arXiv}\emph{ preprint:}
  \href{http://arxiv.org/abs/2103.03570}{{\ttfamily \emph{2103.03570}}}, 2021.

\bibitem[Chae et~al.(2021)Chae, Wilke, and Kafka]{Chae2021}
Chae, Y., Wilke, D.~N., and Kafka, D.
\newblock {GOALS: Gradient-Only Approximations for Line Searches Towards Robust
  and Consistent Training of Deep Neural Networks}. \emph{arXiv}\emph{
  preprint:} \href{http://arxiv.org/abs/2105.10915}{{\ttfamily
  \emph{2105.10915}}}, 2021.

\bibitem[Chakrabarti \& Chopra(2021)Chakrabarti and Chopra]{Chakrabarti2021}
Chakrabarti, K. and Chopra, N.
\newblock {Generalized AdaGrad (G-AdaGrad) and Adam: A State-Space
  Perspective}. \emph{arXiv}\emph{ preprint:}
  \href{http://arxiv.org/abs/2106.00092}{{\ttfamily \emph{2106.00092}}}, 2021.

\bibitem[Chen et~al.(2018)Chen, Choi, Brand, Agrawal, Zhang, and
  Gopalakrishnan]{ChenCBAZG18}
Chen, C., Choi, J., Brand, D., Agrawal, A., Zhang, W., and Gopalakrishnan, K.
\newblock {AdaComp: Adaptive Residual Gradient Compression for Data-Parallel
  Distributed Training}.
\newblock In \emph{32nd {AAAI} Conference on Artificial Intelligence, {AAAI}},
  2018.

\bibitem[Chen et~al.(2020{\natexlab{a}})Chen, Zhou, Tang, Yang, Cao, and
  Gu]{ChenZTYCG20}
Chen, J., Zhou, D., Tang, Y., Yang, Z., Cao, Y., and Gu, Q.
\newblock {Closing the Generalization Gap of Adaptive Gradient Methods in
  Training Deep Neural Networks}.
\newblock In \emph{29th International Joint Conference on Artificial
  Intelligence, {IJCAI}}, 2020{\natexlab{a}}.

\bibitem[Chen et~al.(2020{\natexlab{b}})Chen, Choi, Balles, Duvenaud, and
  Hennig]{Chen2020a}
Chen, R. T.~Q., Choi, D., Balles, L., Duvenaud, D., and Hennig, P.
\newblock {Self-Tuning Stochastic Optimization with Curvature-Aware Gradient
  Filtering}. \emph{arXiv}\emph{ preprint:}
  \href{http://arxiv.org/abs/2011.04803}{{\ttfamily \emph{2011.04803}}},
  2020{\natexlab{b}}.

\bibitem[Chen et~al.(2021)Chen, Guo, Sun, and Yin]{Chen2021}
Chen, T., Guo, Z., Sun, Y., and Yin, W.
\newblock {CADA: Communication-Adaptive Distributed Adam}.
\newblock In \emph{24th International Conference on Artificial Intelligence and
  Statistics, {AISTATS}}, 2021.

\bibitem[Chen et~al.(2019{\natexlab{a}})Chen, Liu, Sun, and
  Hong]{chen2018convergence}
Chen, X., Liu, S., Sun, R., and Hong, M.
\newblock {On the Convergence of A Class of Adam-Type Algorithms for Non-Convex
  Optimization}.
\newblock In \emph{7th International Conference on Learning Representations,
  {ICLR}}, 2019{\natexlab{a}}.

\bibitem[Chen et~al.(2019{\natexlab{b}})Chen, Jing, Zhao, Liu, Li, Qiao, Xue,
  Fu, and Yang]{chen2019adaptive}
Chen, Y., Jing, H., Zhao, W., Liu, Z., Li, O., Qiao, L., Xue, W., Fu, H., and
  Yang, G.
\newblock {An Adaptive Remote Stochastic Gradient Method for Training Neural
  Networks}. \emph{arXiv}\emph{ preprint:}
  \href{http://arxiv.org/abs/1905.01422}{{\ttfamily \emph{1905.01422}}},
  2019{\natexlab{b}}.

\bibitem[Chen et~al.(2019{\natexlab{c}})Chen, Jing, Zhao, Liu, Qiao, Xue, Fu,
  and Yang]{chen2019}
Chen, Y., Jing, H., Zhao, W., Liu, Z., Qiao, L., Xue, W., Fu, H., and Yang, G.
\newblock {NAMSG: An Efficient Method For Training Neural Networks}.
  \emph{arXiv}\emph{ preprint:}
  \href{http://arxiv.org/abs/1905.01422}{{\ttfamily \emph{1905.01422}}},
  2019{\natexlab{c}}.

\bibitem[Chen \& Zhou(2020)Chen and Zhou]{chen2020momentum}
Chen, Z. and Zhou, Y.
\newblock {Momentum with Variance Reduction for Nonconvex Composition
  Optimization}. \emph{arXiv}\emph{ preprint:}
  \href{http://arxiv.org/abs/2005.07755}{{\ttfamily \emph{2005.07755}}}, 2020.

\bibitem[Choi et~al.(2019)Choi, Shallue, Nado, Lee, Maddison, and
  Dahl]{Choi2019}
Choi, D., Shallue, C.~J., Nado, Z., Lee, J., Maddison, C.~J., and Dahl, G.~E.
\newblock {On Empirical Comparisons of Optimizers for Deep Learning}.
  \emph{arXiv}\emph{ preprint:}
  \href{http://arxiv.org/abs/1910.05446}{{\ttfamily \emph{1910.05446}}}, 2019.

\bibitem[Daley \& Amato(2020)Daley and Amato]{Daley2020}
Daley, B. and Amato, C.
\newblock {Expectigrad: Fast Stochastic Optimization with Robust Convergence
  Properties}. \emph{arXiv}\emph{ preprint:}
  \href{http://arxiv.org/abs/2010.01356}{{\ttfamily \emph{2010.01356}}}, 2020.

\bibitem[de~Roos et~al.(2021)de~Roos, Jidling, Wills, Schön, and
  Hennig]{Roos2021}
de~Roos, F., Jidling, C., Wills, A., Schön, T., and Hennig, P.
\newblock {A Probabilistically Motivated Learning Rate Adaptation for
  Stochastic Optimization}. \emph{arXiv}\emph{ preprint:}
  \href{http://arxiv.org/abs/2102.10880}{{\ttfamily \emph{2102.10880}}}, 2021.

\bibitem[Defazio \& Jelassi(2021)Defazio and Jelassi]{Defazio2021}
Defazio, A. and Jelassi, S.
\newblock {Adaptivity without Compromise: A Momentumized, Adaptive, Dual
  Averaged Gradient Method for Stochastic Optimization}. \emph{arXiv}\emph{
  preprint:} \href{http://arxiv.org/abs/2101.11075}{{\ttfamily
  \emph{2101.11075}}}, 2021.

\bibitem[Dellaferrera et~al.(2021)Dellaferrera, Wozniak, Indiveri, Pantazi, and
  Eleftheriou]{Dellaferrera2021}
Dellaferrera, G., Wozniak, S., Indiveri, G., Pantazi, A., and Eleftheriou, E.
\newblock {Learning in Deep Neural Networks Using a Biologically Inspired
  Optimizer}. \emph{arXiv}\emph{ preprint:}
  \href{http://arxiv.org/abs/2104.11604}{{\ttfamily \emph{2104.11604}}}, 2021.

\bibitem[Devarakonda et~al.(2017)Devarakonda, Naumov, and
  Garland]{devarakonda2017adabatch}
Devarakonda, A., Naumov, M., and Garland, M.
\newblock {AdaBatch: Adaptive Batch Sizes for Training Deep Neural Networks}.
  \emph{arXiv}\emph{ preprint:}
  \href{http://arxiv.org/abs/1712.02029}{{\ttfamily \emph{1712.02029}}}, 2017.

\bibitem[Ding et~al.(2019)Ding, Ren, Luo, and Sun]{ding2019adaptive}
Ding, J., Ren, X., Luo, R., and Sun, X.
\newblock {An Adaptive and Momental Bound Method for Stochastic Learning}.
  \emph{arXiv}\emph{ preprint:}
  \href{http://arxiv.org/abs/1910.12249}{{\ttfamily \emph{1910.12249}}}, 2019.

\bibitem[Dozat(2016)]{Dozat2016IncorporatingNM}
Dozat, T.
\newblock {Incorporating Nesterov Momentum into Adam}.
\newblock In \emph{4th International Conference on Learning Representations,
  {ICLR}}, 2016.

\bibitem[Dubey et~al.(2020)Dubey, Chakraborty, Roy, Mukherjee, Singh, and
  Chaudhuri]{Dubey2020}
Dubey, S.~R., Chakraborty, S., Roy, S.~K., Mukherjee, S., Singh, S.~K., and
  Chaudhuri, B.~B.
\newblock {diffGrad: An Optimization Method for Convolutional Neural Networks}.
\newblock \emph{{IEEE} Transactions on Neural Networks and Learning Systems},
  2020.

\bibitem[Duchi et~al.(2011)Duchi, Hazan, and Singer]{Duchi2011}
Duchi, J., Hazan, E., and Singer, Y.
\newblock {Adaptive Subgradient Methods for Online Learning and Stochastic
  Optimization}.
\newblock \emph{Journal of Machine Learning Research, JMLR}, 12, 2011.

\bibitem[Eliyahu(2020)]{ADASRepo}
Eliyahu, Y.
\newblock {ADAS Optimzier}.
\newblock \url{https://github.com/YanaiEliyahu/AdasOptimizer}, 2020.

\bibitem[Fetterman et~al.(2019)Fetterman, Kim, and Albrecht]{fetterman2019}
Fetterman, A.~J., Kim, C.~H., and Albrecht, J.
\newblock {SoftAdam: Unifying SGD and Adam for better stochastic gradient
  descent}.
\newblock \url{https://openreview.net/forum?id=Skgfr1rYDH}, 2019.

\bibitem[Foret et~al.(2021)Foret, Kleiner, Mobahi, and Neyshabur]{Foret2021}
Foret, P., Kleiner, A., Mobahi, H., and Neyshabur, B.
\newblock {Sharpness-aware Minimization for Efficiently Improving
  Generalization}.
\newblock In \emph{9th International Conference on Learning Representations,
  {ICLR}}, 2021.
\newblock URL \url{https://openreview.net/forum?id=6Tm1mposlrM}.

\bibitem[Gao et~al.(2020)Gao, Liu, Huang, Wang, Wang, Wang, Xu, and
  Yu]{Gao2020}
Gao, K.-X., Liu, X.-L., Huang, Z.-H., Wang, M., Wang, S., Wang, Z., Xu, D., and
  Yu, F.
\newblock {Eigenvalue-corrected Natural Gradient Based on a New Approximation}.
  \emph{arXiv}\emph{ preprint:}
  \href{http://arxiv.org/abs/2011.13609}{{\ttfamily \emph{2011.13609}}}, 2020.

\bibitem[George et~al.(2018)George, Laurent, Bouthillier, Ballas, and
  Vincent]{George2018}
George, T., Laurent, C., Bouthillier, X., Ballas, N., and Vincent, P.
\newblock {Fast Approximate Natural Gradient Descent in a Kronecker Factored
  Eigenbasis}.
\newblock In \emph{Advances in Neural Information Processing Systems 31,
  NeurIPS}, 2018.

\bibitem[Ginsburg et~al.(2019)Ginsburg, Castonguay, Hrinchuk, Kuchaiev,
  Lavrukhin, Leary, Li, Nguyen, and Cohen]{Ginsburg2019}
Ginsburg, B., Castonguay, P., Hrinchuk, O., Kuchaiev, O., Lavrukhin, V., Leary,
  R., Li, J., Nguyen, H., and Cohen, J.~M.
\newblock {Stochastic Gradient Methods with Layer-wise Adaptive Moments for
  Training of Deep Networks}. \emph{arXiv}\emph{ preprint:}
  \href{http://arxiv.org/abs/1905.11286}{{\ttfamily \emph{1905.11286}}}, 2019.

\bibitem[Goldfarb et~al.(2020)Goldfarb, Ren, and
  Bahamou]{goldfarb2020practical}
Goldfarb, D., Ren, Y., and Bahamou, A.
\newblock {Practical Quasi-Newton Methods for Training Deep Neural Networks}.
\newblock In \emph{Advances in Neural Information Processing Systems 33,
  NeurIPS}, 2020.

\bibitem[Goodfellow et~al.(2016)Goodfellow, Bengio, and
  Courville]{Goodfellow2016}
Goodfellow, I., Bengio, Y., and Courville, A.
\newblock \emph{{Deep Learning}}.
\newblock MIT Press, 2016.

\bibitem[Goyal et~al.(2017)Goyal, Dollár, Girshick, Noordhuis, Wesolowski,
  Kyrola, Tulloch, Jia, and He]{goyal2017accurate}
Goyal, P., Dollár, P., Girshick, R., Noordhuis, P., Wesolowski, L., Kyrola,
  A., Tulloch, A., Jia, Y., and He, K.
\newblock {Accurate, Large Minibatch SGD: Training ImageNet in 1 Hour}.
  \emph{arXiv}\emph{ preprint:}
  \href{http://arxiv.org/abs/1706.02677}{{\ttfamily \emph{1706.02677}}}, 2017.

\bibitem[Grankin(2020)]{RangerLars}
Grankin, M.
\newblock {RangerLars}.
\newblock \url{https://github.com/mgrankin/over9000}, 2020.

\bibitem[Granziol et~al.(2020)Granziol, Wan, and Roberts]{Granziol2020}
Granziol, D., Wan, X., and Roberts, S.
\newblock {Gadam: Combining Adaptivity with Iterate Averaging Gives Greater
  Generalisation}. \emph{arXiv}\emph{ preprint:}
  \href{http://arxiv.org/abs/2003.01247}{{\ttfamily \emph{2003.01247}}}, 2020.

\bibitem[Gupta et~al.(2018)Gupta, Koren, and Singer]{0001KS18}
Gupta, V., Koren, T., and Singer, Y.
\newblock {Shampoo: Preconditioned Stochastic Tensor Optimization}.
\newblock In \emph{35th International Conference on Machine Learning, {ICML}},
  2018.

\bibitem[Hayashi et~al.(2018)Hayashi, Koushik, and Neubig]{Hayashi2018}
Hayashi, H., Koushik, J., and Neubig, G.
\newblock {Eve: A Gradient Based Optimization Method with Locally and Globally
  Adaptive Learning Rates}. \emph{arXiv}\emph{ preprint:}
  \href{http://arxiv.org/abs/1611.01505}{{\ttfamily \emph{1611.01505}}}, 2018.

\bibitem[He et~al.(2016)He, Zhang, Ren, and Sun]{He2016}
He, K., Zhang, X., Ren, S., and Sun, J.
\newblock {Deep Residual Learning for Image Recognition}.
\newblock In \emph{IEEE Computer Society Conference on Computer Vision and
  Pattern Recognition}, 2016.

\bibitem[Henriques et~al.(2019)Henriques, Ehrhardt, Albanie, and
  Vedaldi]{Henriques2018}
Henriques, J.~F., Ehrhardt, S., Albanie, S., and Vedaldi, A.
\newblock {Small Steps and Giant Leaps: Minimal Newton Solvers for Deep
  Learning}.
\newblock In \emph{{IEEE/CVF} International Conference on Computer Vision,
  {ICCV}}, 2019.

\bibitem[Heo et~al.(2021)Heo, Chun, Oh, Han, Yun, Uh, and Ha]{Heo2021}
Heo, B., Chun, S., Oh, S.~J., Han, D., Yun, S., Uh, Y., and Ha, J.-W.
\newblock {AdamP: Slowing Down the Weight Norm Increase in Momentum-based
  Optimizers}.
\newblock In \emph{7th International Conference on Learning Representations,
  ICLR}, 2021.

\bibitem[Hosseini \& Plataniotis(2020)Hosseini and Plataniotis]{Hosseini2020}
Hosseini, M.~S. and Plataniotis, K.~N.
\newblock {AdaS: Adaptive Scheduling of Stochastic Gradients}.
  \emph{arXiv}\emph{ preprint:}
  \href{http://arxiv.org/abs/2006.06587}{{\ttfamily \emph{2006.06587}}}, 2020.

\bibitem[Howard \& Ruder(2018)Howard and Ruder]{Howard2018}
Howard, J. and Ruder, S.
\newblock {Universal Language Model Fine-tuning for Text Classification}.
\newblock In \emph{56th Annual Meeting of the Association for Computational
  Linguistics}, 2018.

\bibitem[Hu et~al.(2019)Hu, Lin, and Tang]{hu2019secondorder}
Hu, Y., Lin, L., and Tang, S.
\newblock {Second-order Information in First-order Optimization Methods}.
  \emph{arXiv}\emph{ preprint:}
  \href{http://arxiv.org/abs/1912.09926}{{\ttfamily \emph{1912.09926}}}, 2019.

\bibitem[Hu et~al.(2020)Hu, Zhang, Chen, and He]{hu2020biased}
Hu, Y., Zhang, S., Chen, X., and He, N.
\newblock {Biased Stochastic First-Order Methods for Conditional Stochastic
  Optimization and Applications in Meta Learning}.
\newblock In \emph{Advances in Neural Information Processing Systems 33,
  NeurIPS}, 2020.

\bibitem[Huang et~al.(2019)Huang, Wang, and Dong]{HuangWD19}
Huang, H., Wang, C., and Dong, B.
\newblock {Nostalgic Adam: Weighting More of the Past Gradients When Designing
  the Adaptive Learning Rate}.
\newblock In \emph{28th International Joint Conference on Artificial
  Intelligence, {IJCAI}}, 2019.

\bibitem[Huang et~al.(2020)Huang, Zhou, Xu, Wang, and Li]{huang2020adaptive}
Huang, X., Zhou, H., Xu, R., Wang, Z., and Li, L.
\newblock {Adaptive Gradient Methods Can Be Provably Faster than SGD after
  Finite Epochs}. \emph{arXiv}\emph{ preprint:}
  \href{http://arxiv.org/abs/2006.07037}{{\ttfamily \emph{2006.07037}}}, 2020.

\bibitem[Ida et~al.(2017)Ida, Fujiwara, and Iwamura]{IdaFI17}
Ida, Y., Fujiwara, Y., and Iwamura, S.
\newblock {Adaptive Learning Rate via Covariance Matrix Based Preconditioning
  for Deep Neural Networks}.
\newblock In \emph{26th International Joint Conference on Artificial
  Intelligence, {IJCAI}}, 2017.

\bibitem[Ilboudo et~al.(2020)Ilboudo, Kobayashi, and Sugimoto]{ilboudo2020}
Ilboudo, W. E.~L., Kobayashi, T., and Sugimoto, K.
\newblock {TAdam: A Robust Stochastic Gradient Optimizer}. \emph{arXiv}\emph{
  preprint:} \href{http://arxiv.org/abs/2003.00179}{{\ttfamily
  \emph{2003.00179}}}, 2020.

\bibitem[Jiang et~al.(2019)Jiang, Balu, Tan, Lee, Hegde, and
  Sarkar]{jiang2019higherorder}
Jiang, Z., Balu, A., Tan, S.~Y., Lee, Y.~M., Hegde, C., and Sarkar, S.
\newblock {On Higher-order Moments in Adam}. \emph{arXiv}\emph{ preprint:}
  \href{http://arxiv.org/abs/1910.06878}{{\ttfamily \emph{1910.06878}}}, 2019.

\bibitem[Jin et~al.(2021)Jin, Zhou, Zhao, Zhu, Guo, Canini, and
  Krishnamurthy]{Jin2021}
Jin, Y., Zhou, T., Zhao, L., Zhu, Y., Guo, C., Canini, M., and Krishnamurthy,
  A.
\newblock {AutoLRS: Automatic Learning-Rate Schedule by Bayesian Optimization
  on the Fly}.
\newblock In \emph{9th International Conference on Learning Representations,
  {ICLR}}, 2021.
\newblock URL \url{https://openreview.net/forum?id=SlrqM9_lyju}.

\bibitem[Johnson et~al.(2020)Johnson, Agrawal, Gu, and
  Guestrin]{johnson2020adascale}
Johnson, T.~B., Agrawal, P., Gu, H., and Guestrin, C.
\newblock {AdaScale SGD: A User-Friendly Algorithm for Distributed Training}.
\newblock In \emph{37th International Conference on Machine Learning, {ICML}},
  2020.

\bibitem[Kafka \& Wilke(2019)Kafka and Wilke]{kafka2019gradientonly}
Kafka, D. and Wilke, D.
\newblock {Gradient-only line searches: An Alternative to Probabilistic Line
  Searches}. \emph{arXiv}\emph{ preprint:}
  \href{http://arxiv.org/abs/1903.09383}{{\ttfamily \emph{1903.09383}}}, 2019.

\bibitem[Kelterborn et~al.(2020)Kelterborn, Mazur, and
  Petrenko]{kelterborn2020gravilon}
Kelterborn, C., Mazur, M., and Petrenko, B.~V.
\newblock {Gravilon: Applications of a New Gradient Descent Method to Machine
  Learning}. \emph{arXiv}\emph{ preprint:}
  \href{http://arxiv.org/abs/2008.11370}{{\ttfamily \emph{2008.11370}}}, 2020.

\bibitem[Keskar \& Socher(2017)Keskar and Socher]{Keskar2017}
Keskar, N.~S. and Socher, R.
\newblock {Improving Generalization Performance by Switching from Adam to SGD}.
  \emph{arXiv}\emph{ preprint:}
  \href{http://arxiv.org/abs/1712.07628}{{\ttfamily \emph{1712.07628}}}, 2017.

\bibitem[Khan et~al.(2018)Khan, Nielsen, Tangkaratt, Lin, Gal, and
  Srivastava]{KhanNTLGS18}
Khan, M.~E., Nielsen, D., Tangkaratt, V., Lin, W., Gal, Y., and Srivastava, A.
\newblock {Fast and Scalable Bayesian Deep Learning by Weight-Perturbation in
  Adam}.
\newblock In \emph{35th International Conference on Machine Learning, {ICML}},
  2018.

\bibitem[Kingma \& Ba(2015)Kingma and Ba]{Kingma2015}
Kingma, D.~P. and Ba, J.
\newblock {Adam: A Method for Stochastic Optimization}.
\newblock In \emph{3rd International Conference on Learning Representations,
  {ICLR}}, 2015.

\bibitem[Kwon et~al.(2021)Kwon, Kim, Park, and Choi]{Kwon2021}
Kwon, J., Kim, J., Park, H., and Choi, I.~K.
\newblock {ASAM: Adaptive Sharpness-Aware Minimization for Scale-Invariant
  Learning of Deep Neural Networks}. \emph{arXiv}\emph{ preprint:}
  \href{http://arxiv.org/abs/2102.11600}{{\ttfamily \emph{2102.11600}}}, 2021.

\bibitem[Landro et~al.(2020)Landro, Gallo, and Grassa]{Landro2020}
Landro, N., Gallo, I., and Grassa, R.~L.
\newblock {Mixing ADAM and SGD: a Combined Optimization Method}.
  \emph{arXiv}\emph{ preprint:}
  \href{http://arxiv.org/abs/2011.08042}{{\ttfamily \emph{2011.08042}}}, 2020.

\bibitem[Levy et~al.(2018)Levy, Yurtsever, and Cevher]{accelegrad}
Levy, K.~Y., Yurtsever, A., and Cevher, V.
\newblock {Online Adaptive Methods, Universality and Acceleration}.
\newblock In \emph{Advances in Neural Information Processing Systems 31,
  NeurIPS}, 2018.

\bibitem[Li et~al.(2020{\natexlab{a}})Li, Zhang, Wang, and Luo]{li2020adax}
Li, W., Zhang, Z., Wang, X., and Luo, P.
\newblock {AdaX: Adaptive Gradient Descent with Exponential Long Term Memory}.
  \emph{arXiv}\emph{ preprint:}
  \href{http://arxiv.org/abs/2004.09740}{{\ttfamily \emph{2004.09740}}},
  2020{\natexlab{a}}.

\bibitem[Li et~al.(2020{\natexlab{b}})Li, Bao, Zhang, and
  Richtárik]{li2020page}
Li, Z., Bao, H., Zhang, X., and Richtárik, P.
\newblock {PAGE: A Simple and Optimal Probabilistic Gradient Estimator for
  Nonconvex Optimization}. \emph{arXiv}\emph{ preprint:}
  \href{http://arxiv.org/abs/2008.10898}{{\ttfamily \emph{2008.10898}}},
  2020{\natexlab{b}}.

\bibitem[Liu et~al.(2021{\natexlab{a}})Liu, Chen, and Theodorou]{Liu2021}
Liu, G.-H., Chen, T., and Theodorou, E.~A.
\newblock Dynamic game theoretic neural optimizer. \emph{arXiv}\emph{
  preprint:} \href{http://arxiv.org/abs/2105.03788}{{\ttfamily
  \emph{2105.03788}}}, 2021{\natexlab{a}}.

\bibitem[Liu \& Tian(2020)Liu and Tian]{Liu2020c}
Liu, H. and Tian, X.
\newblock {AEGD: Adaptive Gradient Decent with Energy}. \emph{arXiv}\emph{
  preprint:} \href{http://arxiv.org/abs/2010.05109}{{\ttfamily
  \emph{2010.05109}}}, 2020.

\bibitem[Liu \& Luo(2020)Liu and Luo]{liu2020new}
Liu, L. and Luo, X.
\newblock {A New Accelerated Stochastic Gradient Method with Momentum}.
  \emph{arXiv}\emph{ preprint:}
  \href{http://arxiv.org/abs/2006.00423}{{\ttfamily \emph{2006.00423}}}, 2020.

\bibitem[Liu et~al.(2020{\natexlab{a}})Liu, Jiang, He, Chen, Liu, Gao, and
  Han]{Liu2019}
Liu, L., Jiang, H., He, P., Chen, W., Liu, X., Gao, J., and Han, J.
\newblock {On the Variance of the Adaptive Learning Rate and Beyond}.
\newblock In \emph{8th International Conference on Learning Representations,
  {ICLR}}, 2020{\natexlab{a}}.

\bibitem[Liu et~al.(2020{\natexlab{b}})Liu, Zhang, Orabona, and Yang]{Liu2020a}
Liu, M., Zhang, W., Orabona, F., and Yang, T.
\newblock {Adam$^+$: A Stochastic Method with Adaptive Variance Reduction}.
  \emph{arXiv}\emph{ preprint:}
  \href{http://arxiv.org/abs/2011.11985}{{\ttfamily \emph{2011.11985}}},
  2020{\natexlab{b}}.

\bibitem[Liu et~al.(2020{\natexlab{c}})Liu, Wu, and Mozafari]{Liu2020b}
Liu, R., Wu, T., and Mozafari, B.
\newblock {Adam with Bandit Sampling for Deep Learning}.
\newblock In \emph{Advances in Neural Information Processing Systems 33,
  NeurIPS}, 2020{\natexlab{c}}.

\bibitem[Liu et~al.(2021{\natexlab{b}})Liu, Bernstein, Meister, and
  Yue]{Liu2021a}
Liu, Y., Bernstein, J., Meister, M., and Yue, Y.
\newblock {Learning by Turning: Neural Architecture Aware Optimisation},
  \href{http://arxiv.org/abs/arXiv:2102.07227}{{\ttfamily
  \emph{arXiv:2102.07227}}}, 2021{\natexlab{b}}.

\bibitem[Loshchilov \& Hutter(2017)Loshchilov and Hutter]{Loshchilov2017}
Loshchilov, I. and Hutter, F.
\newblock {SGDR: Stochastic Gradient Descent with Warm Restarts}.
\newblock In \emph{5th International Conference on Learning Representations,
  ICLR}, 2017.

\bibitem[Loshchilov \& Hutter(2019)Loshchilov and Hutter]{Loshchilov2019a}
Loshchilov, I. and Hutter, F.
\newblock {Decoupled weight decay regularization}.
\newblock In \emph{7th International Conference on Learning Representations,
  ICLR}, 2019.

\bibitem[Luo et~al.(2019)Luo, Xiong, Liu, and Sun]{Luo2019}
Luo, L., Xiong, Y., Liu, Y., and Sun, X.
\newblock {Adaptive Gradient Methods with Dynamic Bound of Learning Rate}.
\newblock In \emph{7th International Conference on Learning Representations,
  ICLR}, 2019.

\bibitem[Ma \& Yarats(2019)Ma and Yarats]{MaY19}
Ma, J. and Yarats, D.
\newblock {Quasi-hyperbolic momentum and Adam for deep learning}.
\newblock In \emph{7th International Conference on Learning Representations,
  {ICLR}}, 2019.

\bibitem[Mahsereci \& Hennig(2017)Mahsereci and Hennig]{Mahsereci2017}
Mahsereci, M. and Hennig, P.
\newblock {Probabilistic Line Searches for Stochastic Optimization}.
\newblock In \emph{Journal of Machine Learning Research, JMLR}, volume~18,
  2017.

\bibitem[Malkiel \& Wolf(2020)Malkiel and Wolf]{malkiel2020mtadam}
Malkiel, I. and Wolf, L.
\newblock {MTAdam: Automatic Balancing of Multiple Training Loss Terms}.
  \emph{arXiv}\emph{ preprint:}
  \href{http://arxiv.org/abs/2006.14683}{{\ttfamily \emph{2006.14683}}}, 2020.

\bibitem[Martens \& Grosse(2015)Martens and Grosse]{Martens2015}
Martens, J. and Grosse, R.
\newblock {Optimizing Neural Networks with Kronecker-Factored Approximate
  Curvature}.
\newblock In \emph{32nd International Conference on Machine Learning, ICML},
  2015.

\bibitem[Mukkamala \& Hein(2017)Mukkamala and Hein]{MukkamalaH17}
Mukkamala, M.~C. and Hein, M.
\newblock {Variants of RMSProp and Adagrad with Logarithmic Regret Bounds}.
\newblock In \emph{34th International Conference on Machine Learning, {ICML}},
  2017.

\bibitem[Mutschler \& Zell(2020)Mutschler and Zell]{mutschler2019parabolic}
Mutschler, M. and Zell, A.
\newblock {Parabolic Approximation Line Search for DNNs}.
\newblock In \emph{Advances in Neural Information Processing Systems 33,
  NeurIPS}, 2020.

\bibitem[Nazari et~al.(2019)Nazari, Tarzanagh, and
  Michailidis]{nazari2019dadam}
Nazari, P., Tarzanagh, D.~A., and Michailidis, G.
\newblock {DADAM: A Consensus-based Distributed Adaptive Gradient Method for
  Online Optimization}. \emph{arXiv}\emph{ preprint:}
  \href{http://arxiv.org/abs/1901.09109}{{\ttfamily \emph{1901.09109}}}, 2019.

\bibitem[Nesterov(1983)]{Nesterov1983}
Nesterov, Y.
\newblock {A method for solving the convex programming problem with convergence
  rate $O(1/k^2)$}.
\newblock \emph{Soviet Mathematics Doklady}, 27, 1983.

\bibitem[Orabona \& P{\'{a}}l(2015)Orabona and P{\'{a}}l]{Orabona2015}
Orabona, F. and P{\'{a}}l, D.
\newblock {Scale-Free Algorithms for Online Linear Optimization}.
\newblock In \emph{Algorithmic Learning Theory - 26th International Conference,
  {ALT}}, 2015.

\bibitem[Orvieto et~al.(2019)Orvieto, K{\"{o}}hler, and Lucchi]{OrvietoKL19}
Orvieto, A., K{\"{o}}hler, J., and Lucchi, A.
\newblock {The Role of Memory in Stochastic Optimization}.
\newblock In \emph{35th Conference on Uncertainty in Artificial Intelligence,
  {UAI}}, 2019.

\bibitem[Polyak(1964)]{Polyak1964}
Polyak, B.~T.
\newblock {Some methods of speeding up the convergence of iteration methods}.
\newblock \emph{USSR Computational Mathematics and Mathematical Physics},
  4\penalty0 (5), 1964.

\bibitem[Preechakul \& Kijsirikul(2019)Preechakul and
  Kijsirikul]{Preechakul2019}
Preechakul, K. and Kijsirikul, B.
\newblock {CProp: Adaptive Learning Rate Scaling from Past Gradient
  Conformity}. \emph{arXiv}\emph{ preprint:}
  \href{http://arxiv.org/abs/1912.11493}{{\ttfamily \emph{1912.11493}}}, 2019.

\bibitem[Purkayastha \& Purkayastha(2020)Purkayastha and
  Purkayastha]{Purkayastha2020}
Purkayastha, S. and Purkayastha, S.
\newblock {A Variant of Gradient Descent Algorithm Based on Gradient
  Averaging}. \emph{arXiv}\emph{ preprint:}
  \href{http://arxiv.org/abs/2012.02387}{{\ttfamily \emph{2012.02387}}}, 2020.

\bibitem[Ramezani-Kebrya et~al.(2021)Ramezani-Kebrya, Khisti, and
  Liang]{RamezaniKebrya2021}
Ramezani-Kebrya, A., Khisti, A., and Liang, B.
\newblock {On the Generalization of Stochastic Gradient Descent with Momentum}.
  \emph{arXiv}\emph{ preprint:}
  \href{http://arxiv.org/abs/2102.13653}{{\ttfamily \emph{2102.13653}}}, 2021.

\bibitem[Reddi et~al.(2018)Reddi, Kale, and Kumar]{Reddi2018}
Reddi, S.~J., Kale, S., and Kumar, S.
\newblock {On the Convergence of Adam and Beyond}.
\newblock In \emph{6th International Conference on Learning Representations,
  ICLR}, 2018.

\bibitem[Ren \& Goldfarb(2021)Ren and Goldfarb]{Ren2021}
Ren, Y. and Goldfarb, D.
\newblock {Kronecker-factored Quasi-Newton Methods for Convolutional Neural
  Networks}. \emph{arXiv}\emph{ preprint:}
  \href{http://arxiv.org/abs/2102.06737}{{\ttfamily \emph{2102.06737}}}, 2021.

\bibitem[Robbins \& Monro(1951)Robbins and Monro]{Robbins1951}
Robbins, H. and Monro, S.
\newblock {A Stochastic Approximation Method}.
\newblock \emph{The Annals of Mathematical Statistics}, 22\penalty0 (3), 1951.

\bibitem[Rol{\'{i}}nek \& Martius(2018)Rol{\'{i}}nek and Martius]{Rolinek2018}
Rol{\'{i}}nek, M. and Martius, G.
\newblock {L4: Practical loss-based stepsize adaptation for deep learning}.
\newblock In \emph{Advances in Neural Information Processing Systems 31,
  NeurIPS}, 2018.

\bibitem[Roy et~al.(2021)Roy, Paoletti, Haut, Dubey, Kar, Plaza, and
  Chaudhuri]{Roy2021}
Roy, S.~K., Paoletti, M.~E., Haut, J.~M., Dubey, S.~R., Kar, P., Plaza, A., and
  Chaudhuri, B.~B.
\newblock {AngularGrad: A New Optimization Technique for Angular Convergence of
  Convolutional Neural Networks}. \emph{arXiv}\emph{ preprint:}
  \href{http://arxiv.org/abs/2105.10190}{{\ttfamily \emph{2105.10190}}}, 2021.

\bibitem[Salas et~al.(2018)Salas, Kessler, Zohren, and
  Roberts]{salas2018practical}
Salas, A., Kessler, S., Zohren, S., and Roberts, S.
\newblock {Practical Bayesian Learning of Neural Networks via Adaptive
  Subgradient Methods}. \emph{arXiv}\emph{ preprint:}
  \href{http://arxiv.org/abs/1811.03679}{{\ttfamily \emph{1811.03679}}}, 2018.

\bibitem[Savarese et~al.(2019)Savarese, McAllester, Babu, and
  Maire]{savarese2019domainindependent}
Savarese, P., McAllester, D., Babu, S., and Maire, M.
\newblock {Domain-independent Dominance of Adaptive Methods}.
  \emph{arXiv}\emph{ preprint:}
  \href{http://arxiv.org/abs/1912.01823}{{\ttfamily \emph{1912.01823}}}, 2019.

\bibitem[Schaul \& LeCun(2013)Schaul and LeCun]{Schaull3}
Schaul, T. and LeCun, Y.
\newblock {Adaptive learning rates and parallelization for stochastic, sparse,
  non-smooth gradients}.
\newblock In \emph{1st International Conference on Learning Representations,
  {ICLR}}, 2013.

\bibitem[Schaul et~al.(2013)Schaul, Zhang, and LeCun]{SchaulZL13}
Schaul, T., Zhang, S., and LeCun, Y.
\newblock {No more pesky learning rates}.
\newblock In \emph{30th International Conference on Machine Learning, {ICML}},
  2013.

\bibitem[Shang et~al.(2020)Shang, Zhou, Liu, Cheng, Tsang, Zhang, Tao, and
  Jiao]{ShangZLCTZTJ20}
Shang, F., Zhou, K., Liu, H., Cheng, J., Tsang, I.~W., Zhang, L., Tao, D., and
  Jiao, L.
\newblock {VR-SGD: A Simple Stochastic Variance Reduction Method for Machine
  Learning}.
\newblock \emph{{IEEE} Trans. Knowl. Data Eng.}, 32\penalty0 (1), 2020.

\bibitem[Shazeer \& Stern(2018)Shazeer and Stern]{Shazeer2018}
Shazeer, N. and Stern, M.
\newblock {Adafactor: Adaptive Learning Rates with Sublinear Memory Cost}.
\newblock In \emph{35th International Conference on Machine Learning, ICML},
  2018.

\bibitem[Smith(2017)]{Smith2017}
Smith, L.~N.
\newblock {Cyclical Learning Rates for Training Neural Networks}.
\newblock In \emph{{IEEE} Winter Conference on Applications of Computer Vision,
  {WACV}}, 2017.

\bibitem[Smith \& Topin(2017)Smith and Topin]{Smith2017super}
Smith, L.~N. and Topin, N.
\newblock {Super-Convergence: Very Fast Training of Neural Networks Using Large
  Learning Rates}. \emph{arXiv}\emph{ preprint:}
  \href{http://arxiv.org/abs/1708.07120}{{\ttfamily \emph{1708.07120}}}, 2017.

\bibitem[Sun et~al.(2019)Sun, Gu, and Sun]{Sun2019}
Sun, H., Gu, L., and Sun, B.
\newblock {Adathm: Adaptive Gradient Method Based on Estimates of Third-Order
  Moments}.
\newblock In \emph{4th {IEEE} International Conference on Data Science in
  Cyberspace, {DSC}}, 2019.

\bibitem[Sung et~al.(2020)Sung, Choi, Park, Choi, and Shin]{Sung2020}
Sung, W., Choi, I., Park, J., Choi, S., and Shin, S.
\newblock {S-SGD: Symmetrical Stochastic Gradient Descent with Weight Noise
  Injection for Reaching Flat Minima}. \emph{arXiv}\emph{ preprint:}
  \href{http://arxiv.org/abs/2009.02479}{{\ttfamily \emph{2009.02479}}}, 2020.

\bibitem[Tan et~al.(2016)Tan, Ma, Dai, and Qian]{TanMDQ16}
Tan, C., Ma, S., Dai, Y., and Qian, Y.
\newblock {Barzilai-Borwein Step Size for Stochastic Gradient Descent}.
\newblock In \emph{Advances in Neural Information Processing Systems 29, NIPS},
  2016.

\bibitem[Tao et~al.(2019)Tao, Xia, and Li]{tao2019}
Tao, Z., Xia, Q., and Li, Q.
\newblock {A new perspective in understanding of Adam-Type algorithms and
  beyond}.
\newblock \url{https://openreview.net/forum?id=SyxM51BYPB}, 2019.

\bibitem[Teixeira et~al.(2019)Teixeira, Tamersoy, Singh, and
  Kapoor]{teixeira2019adaloss}
Teixeira, B., Tamersoy, B., Singh, V., and Kapoor, A.
\newblock {Adaloss: Adaptive Loss Function for Landmark Localization}.
  \emph{arXiv}\emph{ preprint:}
  \href{http://arxiv.org/abs/1908.01070}{{\ttfamily \emph{1908.01070}}}, 2019.

\bibitem[Tieleman \& Hinton(2012)Tieleman and Hinton]{Tieleman2012}
Tieleman, T. and Hinton, G.
\newblock {Lecture 6.5---RMSProp: Divide the gradient by a running average of
  its recent magnitude}, 2012.

\bibitem[Tong et~al.(2019)Tong, Liang, and Bi]{tong2019calibrating}
Tong, Q., Liang, G., and Bi, J.
\newblock {Calibrating the Adaptive Learning Rate to Improve Convergence of
  ADAM}. \emph{arXiv}\emph{ preprint:}
  \href{http://arxiv.org/abs/1908.00700}{{\ttfamily \emph{1908.00700}}}, 2019.

\bibitem[Tran \& Cutkosky(2021)Tran and Cutkosky]{Tran2021}
Tran, H. and Cutkosky, A.
\newblock {Correcting Momentum with Second-order Information}.
  \emph{arXiv}\emph{ preprint:}
  \href{http://arxiv.org/abs/2103.03265}{{\ttfamily \emph{2103.03265}}}, 2021.

\bibitem[Tran \& Phong(2019)Tran and Phong]{Tran_2019}
Tran, P.~T. and Phong, L.~T.
\newblock {On the Convergence Proof of AMSGrad and a New Version}.
\newblock \emph{IEEE Access}, 7, 2019.

\bibitem[Tran et~al.(2020)Tran, Nguyen, and Tran-Dinh]{Tran2020}
Tran, T.~H., Nguyen, L.~M., and Tran-Dinh, Q.
\newblock {Shuffling Gradient-Based Methods with Momentum}. \emph{arXiv}\emph{
  preprint:} \href{http://arxiv.org/abs/2011.11884}{{\ttfamily
  \emph{2011.11884}}}, 2020.

\bibitem[Tutunov et~al.(2020)Tutunov, Li, Cowen-Rivers, Wang, and
  Bou-Ammar]{tutunov2020compositional}
Tutunov, R., Li, M., Cowen-Rivers, A.~I., Wang, J., and Bou-Ammar, H.
\newblock {Compositional ADAM: An Adaptive Compositional Solver}.
  \emph{arXiv}\emph{ preprint:}
  \href{http://arxiv.org/abs/2002.03755}{{\ttfamily \emph{2002.03755}}}, 2020.

\bibitem[Vaswani et~al.(2017)Vaswani, Shazeer, Parmar, Uszkoreit, Jones, Gomez,
  Kaiser, and Polosukhin]{Vaswani2017}
Vaswani, A., Shazeer, N., Parmar, N., Uszkoreit, J., Jones, L., Gomez, A.~N.,
  Kaiser, {\L}., and Polosukhin, I.
\newblock {Attention Is All You Need}.
\newblock In \emph{Advances in Neural Information Processing Systems 30, NIPS},
  2017.

\bibitem[Vaswani et~al.(2019)Vaswani, Mishkin, Laradji, Schmidt, Gidel, and
  Lacoste{-}Julien]{Vaswani2019a}
Vaswani, S., Mishkin, A., Laradji, I.~H., Schmidt, M., Gidel, G., and
  Lacoste{-}Julien, S.
\newblock {Painless Stochastic Gradient: Interpolation, Line-Search, and
  Convergence Rates}.
\newblock In \emph{Advances in Neural Information Processing Systems 32,
  NeurIPS}, 2019.

\bibitem[Vogels et~al.(2019)Vogels, Karimireddy, and Jaggi]{VogelsKJ19}
Vogels, T., Karimireddy, S.~P., and Jaggi, M.
\newblock {PowerSGD: Practical Low-Rank Gradient Compression for Distributed
  Optimization}.
\newblock In \emph{Advances in Neural Information Processing Systems 32,
  NeurIPS}, 2019.

\bibitem[Wang \& Ye(2020)Wang and Ye]{Wang2020}
Wang, B. and Ye, Q.
\newblock {Stochastic Gradient Descent with Nonlinear Conjugate Gradient-Style
  Adaptive Momentum}. \emph{arXiv}\emph{ preprint:}
  \href{http://arxiv.org/abs/2012.02188}{{\ttfamily \emph{2012.02188}}}, 2020.

\bibitem[Wang et~al.(2020{\natexlab{a}})Wang, Nguyen, Bertozzi, Baraniuk, and
  Osher]{wang2020scheduled}
Wang, B., Nguyen, T.~M., Bertozzi, A.~L., Baraniuk, R.~G., and Osher, S.~J.
\newblock {Scheduled Restart Momentum for Accelerated Stochastic Gradient
  Descent}. \emph{arXiv}\emph{ preprint:}
  \href{http://arxiv.org/abs/2002.10583}{{\ttfamily \emph{2002.10583}}},
  2020{\natexlab{a}}.

\bibitem[Wang et~al.(2019{\natexlab{a}})Wang, Liu, Tang, Shang, Liu, Sun, and
  Jiao]{WangLTSLSJ19}
Wang, D., Liu, Y., Tang, W., Shang, F., Liu, H., Sun, Q., and Jiao, L.
\newblock {signADAM++: Learning Confidences for Deep Neural Networks}.
\newblock In \emph{International Conference on Data Mining Workshops, {ICDM}},
  2019{\natexlab{a}}.

\bibitem[Wang et~al.(2020{\natexlab{b}})Wang, Lu, Cheng, Tu, and
  Zhang]{Wang2019}
Wang, G., Lu, S., Cheng, Q., Tu, W., and Zhang, L.
\newblock {SAdam: A Variant of Adam for Strongly Convex Functions}.
\newblock In \emph{8th International Conference on Learning Representations,
  {ICLR}}, 2020{\natexlab{b}}.

\bibitem[Wang \& Wiens(2020)Wang and Wiens]{wang2020adasgd}
Wang, J. and Wiens, J.
\newblock {AdaSGD: Bridging the gap between SGD and Adam}. \emph{arXiv}\emph{
  preprint:} \href{http://arxiv.org/abs/2006.16541}{{\ttfamily
  \emph{2006.16541}}}, 2020.

\bibitem[Wang et~al.(2019{\natexlab{b}})Wang, Sun, and Xu]{wang2018hyperadam}
Wang, S., Sun, J., and Xu, Z.
\newblock {HyperAdam: A Learnable Task-Adaptive Adam for Network Training}.
\newblock In \emph{33rd {AAAI} Conference on Artificial Intelligence, {AAAI}},
  2019{\natexlab{b}}.

\bibitem[Wright(2020{\natexlab{a}})]{DeepMemory}
Wright, L.
\newblock {Deep Memory}.
\newblock
  \url{https://github.com/lessw2020/Best-Deep-Learning-Optimizers/tree/master/DeepMemory},
  2020{\natexlab{a}}.

\bibitem[Wright(2020{\natexlab{b}})]{Ranger}
Wright, L.
\newblock {Ranger}.
\newblock \url{https://github.com/lessw2020/Ranger-Deep-Learning-Optimizer},
  2020{\natexlab{b}}.

\bibitem[Wu et~al.(2018)Wu, Ward, and Bottou]{wu18}
Wu, X., Ward, R., and Bottou, L.
\newblock {WNGrad: Learn the Learning Rate in Gradient Descent}.
  \emph{arXiv}\emph{ preprint:}
  \href{http://arxiv.org/abs/1803.02865}{{\ttfamily \emph{1803.02865}}}, 2018.

\bibitem[Xie et~al.(2019)Xie, Koyejo, Gupta, and Lin]{xie2019local}
Xie, C., Koyejo, O., Gupta, I., and Lin, H.
\newblock {Local AdaAlter: Communication-Efficient Stochastic Gradient Descent
  with Adaptive Learning Rates}. \emph{arXiv}\emph{ preprint:}
  \href{http://arxiv.org/abs/1911.09030}{{\ttfamily \emph{1911.09030}}}, 2019.

\bibitem[Xie et~al.(2020)Xie, Wang, Zhang, Sato, and Sugiyama]{Xie2020}
Xie, Z., Wang, X., Zhang, H., Sato, I., and Sugiyama, M.
\newblock {Adai: Separating the Effects of Adaptive Learning Rate and Momentum
  Inertia}. \emph{arXiv}\emph{ preprint:}
  \href{http://arxiv.org/abs/2006.15815}{{\ttfamily \emph{2006.15815}}}, 2020.

\bibitem[Xing et~al.(2018)Xing, Arpit, Tsirigotis, and Bengio]{Xing2018}
Xing, C., Arpit, D., Tsirigotis, C., and Bengio, Y.
\newblock {A Walk with SGD}. \emph{arXiv}\emph{ preprint:}
  \href{http://arxiv.org/abs/1802.08770}{{\ttfamily \emph{1802.08770}}}, 2018.

\bibitem[Xu(2020)]{xu2020momentumbased}
Xu, Y.
\newblock {Momentum-based variance-reduced proximal stochastic gradient method
  for composite nonconvex stochastic optimization}. \emph{arXiv}\emph{
  preprint:} \href{http://arxiv.org/abs/2006.00425}{{\ttfamily
  \emph{2006.00425}}}, 2020.

\bibitem[Yang et~al.(2020)Yang, Xu, Li, Wen, and Chen]{yang2020structured}
Yang, M., Xu, D., Li, Y., Wen, Z., and Chen, M.
\newblock {Structured Stochastic Quasi-Newton Methods for Large-Scale
  Optimization Problems}. \emph{arXiv}\emph{ preprint:}
  \href{http://arxiv.org/abs/2006.09606}{{\ttfamily \emph{2006.09606}}}, 2020.

\bibitem[Yao et~al.(2020)Yao, Gholami, Shen, Keutzer, and
  Mahoney]{yao2020adahessian}
Yao, Z., Gholami, A., Shen, S., Keutzer, K., and Mahoney, M.~W.
\newblock {ADAHESSIAN: An Adaptive Second Order Optimizer for Machine
  Learning}. \emph{arXiv}\emph{ preprint:}
  \href{http://arxiv.org/abs/2006.00719}{{\ttfamily \emph{2006.00719}}}, 2020.

\bibitem[You et~al.(2017)You, Gitman, and Ginsburg]{you2017large}
You, Y., Gitman, I., and Ginsburg, B.
\newblock {Large Batch Training of Convolutional Networks}. \emph{arXiv}\emph{
  preprint:} \href{http://arxiv.org/abs/1708.03888}{{\ttfamily
  \emph{1708.03888}}}, 2017.

\bibitem[You et~al.(2020)You, Li, Reddi, Hseu, Kumar, Bhojanapalli, Song,
  Demmel, Keutzer, and Hsieh]{you2019large}
You, Y., Li, J., Reddi, S., Hseu, J., Kumar, S., Bhojanapalli, S., Song, X.,
  Demmel, J., Keutzer, K., and Hsieh, C.-J.
\newblock {Large Batch Optimization for Deep Learning: Training BERT in 76
  minutes}.
\newblock In \emph{8th International Conference on Learning Representations,
  {ICLR}}, 2020.

\bibitem[Yuan \& Gao(2020)Yuan and Gao]{Yuan2020}
Yuan, W. and Gao, K.-X.
\newblock {EAdam Optimizer: How $\epsilon$ Impact Adam}. \emph{arXiv}\emph{
  preprint:} \href{http://arxiv.org/abs/2011.02150}{{\ttfamily
  \emph{2011.02150}}}, 2020.

\bibitem[Yue et~al.(2020)Yue, Nouiehed, and Kontar]{Yue2020}
Yue, X., Nouiehed, M., and Kontar, R.~A.
\newblock {SALR: Sharpness-aware Learning Rates for Improved Generalization}.
  \emph{arXiv}\emph{ preprint:}
  \href{http://arxiv.org/abs/2011.05348}{{\ttfamily \emph{2011.05348}}}, 2020.

\bibitem[Yun et~al.(2019)Yun, Lozano, and Yang]{yun2019stochastic}
Yun, J., Lozano, A.~C., and Yang, E.
\newblock {Stochastic Gradient Methods with Block Diagonal Matrix Adaptation}.
  \emph{arXiv}\emph{ preprint:}
  \href{http://arxiv.org/abs/1905.10757}{{\ttfamily \emph{1905.10757}}}, 2019.

\bibitem[Zaheer et~al.(2018)Zaheer, Reddi, Sachan, Kale, and
  Kumar]{ZaheerRSKK18}
Zaheer, M., Reddi, S.~J., Sachan, D.~S., Kale, S., and Kumar, S.
\newblock {Adaptive Methods for Nonconvex Optimization}.
\newblock In \emph{Advances in Neural Information Processing Systems 31,
  NeurIPS}, 2018.

\bibitem[Zeiler(2012)]{Zeiler2012}
Zeiler, M.~D.
\newblock {ADADELTA: An Adaptive Learning Rate Method}. \emph{arXiv}\emph{
  preprint:} \href{http://arxiv.org/abs/1212.5701}{{\ttfamily
  \emph{1212.5701}}}, 2012.

\bibitem[Zhang et~al.(2018)Zhang, Sun, Duvenaud, and Grosse]{Zhang2018}
Zhang, G., Sun, S., Duvenaud, D., and Grosse, R.
\newblock {Noisy Natural Gradient as Variational Inference}.
\newblock In \emph{35th International Conference on Machine Learning, ICML},
  2018.

\bibitem[Zhang \& Gouza(2018)Zhang and Gouza]{zhang2018gadam}
Zhang, J. and Gouza, F.~B.
\newblock {GADAM: Genetic-Evolutionary ADAM for Deep Neural Network
  Optimization}. \emph{arXiv}\emph{ preprint:}
  \href{http://arxiv.org/abs/1805.07500}{{\ttfamily \emph{1805.07500}}}, 2018.

\bibitem[Zhang \& Mitliagkas(2019)Zhang and Mitliagkas]{ZhangMR17}
Zhang, J. and Mitliagkas, I.
\newblock {YellowFin and the Art of Momentum Tuning}.
\newblock In \emph{Machine Learning and Systems, MLSys}, 2019.

\bibitem[Zhang et~al.(2020)Zhang, Karimireddy, Veit, Kim, Reddi, Kumar, and
  Sra]{zhang2020adaptive}
Zhang, J., Karimireddy, S.~P., Veit, A., Kim, S., Reddi, S.~J., Kumar, S., and
  Sra, S.
\newblock Why are adaptive methods good for attention models?
\newblock In \emph{Advances in Neural Information Processing Systems 33,
  NeurIPS}, 2020.

\bibitem[Zhang et~al.(2019)Zhang, Lucas, Hinton, and Ba]{Zhang2019a}
Zhang, M.~R., Lucas, J., Hinton, G., and Ba, J.
\newblock {Lookahead Optimizer: k steps forward, 1 step back}.
\newblock \emph{Advances in Neural Information Processing Systems 32, NeurIPS},
  2019.

\bibitem[Zhang et~al.(2017{\natexlab{a}})Zhang, Ma, Li, and
  Wu]{zhang2017normalized}
Zhang, Z., Ma, L., Li, Z., and Wu, C.
\newblock {Normalized Direction-preserving Adam}. \emph{arXiv}\emph{ preprint:}
  \href{http://arxiv.org/abs/1709.04546}{{\ttfamily \emph{1709.04546}}},
  2017{\natexlab{a}}.

\bibitem[Zhang et~al.(2017{\natexlab{b}})Zhang, Wu, and Wang]{Zhang2017}
Zhang, Z., Wu, Y., and Wang, G.
\newblock {BPGrad: Towards Global Optimality in Deep Learning via Branch and
  Pruning}. \emph{arXiv}\emph{ preprint:}
  \href{http://arxiv.org/abs/1711.06959}{{\ttfamily \emph{1711.06959}}},
  2017{\natexlab{b}}.

\bibitem[Zhao et~al.(2020)Zhao, Xie, and Li]{zhao2020stochastic}
Zhao, S.-Y., Xie, Y.-P., and Li, W.-J.
\newblock {Stochastic Normalized Gradient Descent with Momentum for Large Batch
  Training}. \emph{arXiv}\emph{ preprint:}
  \href{http://arxiv.org/abs/2007.13985}{{\ttfamily \emph{2007.13985}}}, 2020.

\bibitem[Zhou et~al.(2020)Zhou, Zheng, and Gao]{Zhou2020}
Zhou, B., Zheng, X., and Gao, J.
\newblock {On the Trend-corrected Variant of Adaptive Stochastic Optimization
  Methods}.
\newblock In \emph{International Joint Conference on Neural Networks, {IJCNN}},
  2020.

\bibitem[Zhou et~al.(2021{\natexlab{a}})Zhou, Huang, Cheng, Wang, and
  Liu]{Zhou2021}
Zhou, Y., Huang, K., Cheng, C., Wang, X., and Liu, X.
\newblock {FastAdaBelief: Improving Convergence Rate for Belief-based Adaptive
  Optimizer by Strong Convexity}. \emph{arXiv}\emph{ preprint:}
  \href{http://arxiv.org/abs/2104.13790}{{\ttfamily \emph{2104.13790}}},
  2021{\natexlab{a}}.

\bibitem[Zhou et~al.(2021{\natexlab{b}})Zhou, Li, and Banerjee]{Zhou2021a}
Zhou, Y., Li, X., and Banerjee, A.
\newblock {Noisy Truncated SGD: Optimization and Generalization}.
  \emph{arXiv}\emph{ preprint:}
  \href{http://arxiv.org/abs/2103.00075}{{\ttfamily \emph{2103.00075}}},
  2021{\natexlab{b}}.

\bibitem[Zhou et~al.(2019)Zhou, Zhang, Lu, Wang, Zhang, and
  Yu]{zhou2018adashift}
Zhou, Z., Zhang, Q., Lu, G., Wang, H., Zhang, W., and Yu, Y.
\newblock {AdaShift: Decorrelation and Convergence of Adaptive Learning Rate
  Methods}.
\newblock In \emph{7th International Conference on Learning Representations,
  {ICLR}}, 2019.

\bibitem[Zhuang et~al.(2020)Zhuang, Tang, Ding, Tatikonda, Dvornek,
  Papademetris, and Duncan]{Zhuang2020}
Zhuang, J., Tang, T., Ding, Y., Tatikonda, S., Dvornek, N., Papademetris, X.,
  and Duncan, J.~S.
\newblock {AdaBelief Optimizer: Adapting Stepsizes by the Belief in Observed
  Gradients}.
\newblock In \emph{Advances in Neural Information Processing Systems 33,
  NeurIPS}, 2020.

\bibitem[Ziyin et~al.(2020)Ziyin, Wang, and Ueda]{ziyin2020laprop}
Ziyin, L., Wang, Z.~T., and Ueda, M.
\newblock {LaProp: a Better Way to Combine Momentum with Adaptive Gradient}.
  \emph{arXiv}\emph{ preprint:}
  \href{http://arxiv.org/abs/2002.04839}{{\ttfamily \emph{2002.04839}}}, 2020.

\end{thebibliography}


\begin{thebibliography}{19}
\providecommand{\natexlab}[1]{#1}
\providecommand{\url}[1]{\texttt{#1}}
\expandafter\ifx\csname urlstyle\endcsname\relax
  \providecommand{\doi}[1]{doi: #1}\else
  \providecommand{\doi}{doi: \begingroup \urlstyle{rm}\Url}\fi

\bibitem[Abadi et~al.(2015)Abadi, Agarwal, Barham, Brevdo, Chen, Citro,
  Corrado, Davis, Dean, Devin, Ghemawat, Goodfellow, Harp, Irving, Isard, Jia,
  Jozefowicz, Kaiser, Kudlur, Levenberg, Man\'{e}, Monga, Moore, Murray, Olah,
  Schuster, Shlens, Steiner, Sutskever, Talwar, Tucker, Vanhoucke, Vasudevan,
  Vi\'{e}gas, Vinyals, Warden, Wattenberg, Wicke, Yu, and Zheng]{Abadi2015}
Abadi, M., Agarwal, A., Barham, P., Brevdo, E., Chen, Z., Citro, C., Corrado,
  G.~S., Davis, A., Dean, J., Devin, M., Ghemawat, S., Goodfellow, I., Harp,
  A., Irving, G., Isard, M., Jia, Y., Jozefowicz, R., Kaiser, L., Kudlur, M.,
  Levenberg, J., Man\'{e}, D., Monga, R., Moore, S., Murray, D., Olah, C.,
  Schuster, M., Shlens, J., Steiner, B., Sutskever, I., Talwar, K., Tucker, P.,
  Vanhoucke, V., Vasudevan, V., Vi\'{e}gas, F., Vinyals, O., Warden, P.,
  Wattenberg, M., Wicke, M., Yu, Y., and Zheng, X.
\newblock {TensorFlow}: {Large-Scale Machine Learning on Heterogeneous
  Systems}, 2015.
\newblock URL \url{http://tensorflow.org/}.

\bibitem[Agarwal et~al.(2020)Agarwal, Anil, Hazan, Koren, and
  Zhang]{Agarwal2020}
Agarwal, N., Anil, R., Hazan, E., Koren, T., and Zhang, C.
\newblock {Disentangling Adaptive Gradient Methods from Learning Rates}.
  \emph{arXiv}\emph{ preprint:}
  \href{http://arxiv.org/abs/2002.11803}{{\ttfamily \emph{2002.11803}}}, 2020.

\bibitem[Bergstra \& Bengio(2012)Bergstra and Bengio]{BergstraB12}
Bergstra, J. and Bengio, Y.
\newblock {Random Search for Hyper-Parameter Optimization}.
\newblock \emph{Journal of Machine Learning Research, JMLR}, 13, 2012.

\bibitem[Choi et~al.(2019)Choi, Shallue, Nado, Lee, Maddison, and
  Dahl]{Choi2019}
Choi, D., Shallue, C.~J., Nado, Z., Lee, J., Maddison, C.~J., and Dahl, G.~E.
\newblock {On Empirical Comparisons of Optimizers for Deep Learning}.
  \emph{arXiv}\emph{ preprint:}
  \href{http://arxiv.org/abs/1910.05446}{{\ttfamily \emph{1910.05446}}}, 2019.

\bibitem[Deng et~al.(2009)Deng, Dong, Socher, Li, Li, and Fei-Fei]{Deng2009}
Deng, J., Dong, W., Socher, R., Li, L.-J., Li, K., and Fei-Fei, L.
\newblock {ImageNet: A Large-Scale Hierarchical Image Database}.
\newblock In \emph{Proceedings of the IEEE Computer Society Conference on
  Computer Vision and Pattern Recognition}. IEEE, 2009.

\bibitem[Garipov et~al.(2018)Garipov, Izmailov, Podoprikhin, Vetrov, and
  Wilson]{Garipov2018}
Garipov, T., Izmailov, P., Podoprikhin, D., Vetrov, D.~P., and Wilson, A.~G.
\newblock {Loss Surfaces, Mode Connectivity, and Fast Ensembling of DNNs}.
\newblock In \emph{Advances in Neural Information Processing Systems 31,
  NeurIPS}, 2018.

\bibitem[Goodfellow et~al.(2016)Goodfellow, Bengio, and
  Courville]{Goodfellow2016}
Goodfellow, I., Bengio, Y., and Courville, A.
\newblock \emph{{Deep Learning}}.
\newblock MIT Press, 2016.

\bibitem[Goodfellow et~al.(2014)Goodfellow, Pouget-Abadie, Mirza, Xu,
  Warde-Farley, Ozair, Courville, and Bengio]{Goodfellow2014}
Goodfellow, I.~J., Pouget-Abadie, J., Mirza, M., Xu, B., Warde-Farley, D.,
  Ozair, S., Courville, A., and Bengio, Y.
\newblock {Generative Adversarial Nets}.
\newblock In Ghahramani, Z., Welling, M., Cortes, C., Lawrence, N.~D., and
  Weinberger, K.~Q. (eds.), \emph{Advances in Neural Information Processing
  Systems 27, NIPS}, 2014.

\bibitem[Goyal et~al.(2017)Goyal, Dollár, Girshick, Noordhuis, Wesolowski,
  Kyrola, Tulloch, Jia, and He]{goyal2017accurate}
Goyal, P., Dollár, P., Girshick, R., Noordhuis, P., Wesolowski, L., Kyrola,
  A., Tulloch, A., Jia, Y., and He, K.
\newblock {Accurate, Large Minibatch SGD: Training ImageNet in 1 Hour}.
  \emph{arXiv}\emph{ preprint:}
  \href{http://arxiv.org/abs/1706.02677}{{\ttfamily \emph{1706.02677}}}, 2017.

\bibitem[Izmailov et~al.(2018)Izmailov, Podoprikhin, Garipov, Vetrov, and
  G]{Izmailov2018}
Izmailov, P., Podoprikhin, D., Garipov, T., Vetrov, D., and G, W.~A.
\newblock {Averaging weights leads to wider optima and better generalization}.
\newblock In \emph{Uncertainty in Artificial Intelligence - Proceedings of the
  34th Conference, UAI 2018}, 2018.

\bibitem[Loshchilov \& Hutter(2017)Loshchilov and Hutter]{Loshchilov2017}
Loshchilov, I. and Hutter, F.
\newblock {SGDR: Stochastic Gradient Descent with Warm Restarts}.
\newblock In \emph{5th International Conference on Learning Representations,
  ICLR}, 2017.

\bibitem[Metz et~al.(2020)Metz, Maheswaranathan, Sun, Freeman, Poole, and
  Sohl-Dickstein]{Metz2020}
Metz, L., Maheswaranathan, N., Sun, R., Freeman, C.~D., Poole, B., and
  Sohl-Dickstein, J.
\newblock {Using a thousand optimization tasks to learn hyperparameter search
  strategies}. \emph{arXiv}\emph{ preprint:}
  \href{http://arxiv.org/abs/2002.11887}{{\ttfamily \emph{2002.11887}}}, 2020.

\bibitem[Schneider et~al.(2019)Schneider, Balles, and Hennig]{Schneider2019}
Schneider, F., Balles, L., and Hennig, P.
\newblock {DeepOBS: A Deep Learning Optimizer Benchmark Suite}.
\newblock In \emph{7th International Conference on Learning Representations,
  ICLR}, 2019.

\bibitem[Sivaprasad et~al.(2020)Sivaprasad, Mai, Vogels, Jaggi, and
  Fleuret]{Sivaprasad2020}
Sivaprasad, P.~T., Mai, F., Vogels, T., Jaggi, M., and Fleuret, F.
\newblock {Optimizer Benchmarking Needs to Account for Hyperparameter Tuning}.
\newblock In \emph{37th International Conference on Machine Learning, ICML},
  2020.

\bibitem[Vaswani et~al.(2017)Vaswani, Shazeer, Parmar, Uszkoreit, Jones, Gomez,
  Kaiser, and Polosukhin]{Vaswani2017}
Vaswani, A., Shazeer, N., Parmar, N., Uszkoreit, J., Jones, L., Gomez, A.~N.,
  Kaiser, {\L}., and Polosukhin, I.
\newblock {Attention Is All You Need}.
\newblock In \emph{Advances in Neural Information Processing Systems 30, NIPS},
  2017.

\bibitem[Wilson et~al.(2017)Wilson, Roelofs, Stern, Srebro, and
  Recht]{Wilson2017}
Wilson, A.~C., Roelofs, R., Stern, M., Srebro, N., and Recht, B.
\newblock {The Marginal Value of Adaptive Gradient Methods in Machine
  Learning}.
\newblock In \emph{Advances in Neural Information Processing Systems 30, NIPS},
  2017.

\bibitem[Wolpert \& Macready(1997)Wolpert and Macready]{DolpertM97}
Wolpert, D.~H. and Macready, W.~G.
\newblock No free lunch theorems for optimization.
\newblock \emph{{IEEE} Trans. Evol. Comput.}, 1\penalty0 (1):\penalty0 67--82,
  1997.

\bibitem[Xing et~al.(2018)Xing, Arpit, Tsirigotis, and Bengio]{Xing2018}
Xing, C., Arpit, D., Tsirigotis, C., and Bengio, Y.
\newblock {A Walk with SGD}. \emph{arXiv}\emph{ preprint:}
  \href{http://arxiv.org/abs/1802.08770}{{\ttfamily \emph{1802.08770}}}, 2018.

\bibitem[Zhang et~al.(2020)Zhang, Lipton, Li, and Smola]{zhang2020dive}
Zhang, A., Lipton, Z.~C., Li, M., and Smola, A.~J.
\newblock \emph{Dive into Deep Learning}.
\newblock 2020.
\newblock \url{https://d2l.ai}.

\end{thebibliography}
\bibliographystyle{bibliography/icml2021}


\onecolumn
\clearpage
\begin{appendices}
	\appendix

\renewcommand{\arraystretch}{1.1} 

\section{List of optimizers and schedules considered}
\begin{table}[H]
	\vspace{-0.5em}%
	\caption{
		List of optimizers considered for our benchmark. This is only a subset of all existing methods for deep learning.
}
	\label{tab:Optimizers}
	\vspace{-0.5em}%
	\centering
	\begin{tiny}
        \begin{tabularx}{.8\textwidth}{XlrXlrX}
                \toprule
                & \textbf{Name} & \textbf{Ref.} &  & \textbf{Name} & \textbf{Ref.} &  \\
                \midrule
                & AcceleGrad & \citepappendix{accelegrad} &  & HyperAdam & \citepappendix{wang2018hyperadam} &  \\
                & ACClip & \citepappendix{zhang2020adaptive} &  & K-BFGS/K-BFGS(L) & \citepappendix{goldfarb2020practical} &  \\
                & AdaAlter & \citepappendix{xie2019local} &  & KF-QN-CNN & \citepappendix{Ren2021} &  \\
                & AdaBatch & \citepappendix{devarakonda2017adabatch} &  & KFAC & \citepappendix{Martens2015} &  \\
                & AdaBayes/AdaBayes-SS & \citepappendix{aitchison2018bayesian} &  & KFLR/KFRA & \citepappendix{Botev2017} &  \\
                & AdaBelief & \citepappendix{Zhuang2020} &  & L4Adam/L4Momentum & \citepappendix{Rolinek2018} &  \\
                & AdaBlock & \citepappendix{yun2019stochastic} &  & LAMB & \citepappendix{you2019large} &  \\
                & AdaBound & \citepappendix{Luo2019} &  & LaProp & \citepappendix{ziyin2020laprop} &  \\
                & AdaComp & \citepappendix{ChenCBAZG18} &  & LARS & \citepappendix{you2017large} &  \\
                & Adadelta & \citepappendix{Zeiler2012} &  & LHOPT & \citepappendix{Almeida2021} &  \\
                & Adafactor & \citepappendix{Shazeer2018} &  & LookAhead & \citepappendix{Zhang2019a} &  \\
                & AdaFix & \citepappendix{bae2019does} &  & M-SVAG & \citepappendix{Balles2018} &  \\
                & AdaFom & \citepappendix{chen2018convergence} &  & MADGRAD & \citepappendix{Defazio2021} &  \\
                & AdaFTRL & \citepappendix{Orabona2015} &  & MAS & \citepappendix{Landro2020} &  \\
                & Adagrad & \citepappendix{Duchi2011} &  & MEKA & \citepappendix{Chen2020a} &  \\
                & ADAHESSIAN & \citepappendix{yao2020adahessian} &  & MTAdam & \citepappendix{malkiel2020mtadam} &  \\
                & Adai & \citepappendix{Xie2020} &  & MVRC-1/MVRC-2 & \citepappendix{chen2020momentum} &  \\
                & AdaLoss & \citepappendix{teixeira2019adaloss} &  & Nadam & \citepappendix{Dozat2016IncorporatingNM} &  \\
                & Adam & \citepappendix{Kingma2015} &  & NAMSB/NAMSG & \citepappendix{chen2019adaptive} &  \\
                & Adam$^+$ & \citepappendix{Liu2020a} &  & ND-Adam & \citepappendix{zhang2017normalized} &  \\
                & AdamAL & \citepappendix{tao2019} &  & Nero & \citepappendix{Liu2021a} &  \\
                & AdaMax & \citepappendix{Kingma2015} &  & Nesterov & \citepappendix{Nesterov1983} &  \\
                & AdamBS & \citepappendix{Liu2020b} &  & Noisy Adam/Noisy K-FAC & \citepappendix{Zhang2018} &  \\
                & AdamNC & \citepappendix{Reddi2018} &  & NosAdam & \citepappendix{HuangWD19} &  \\
                & AdaMod & \citepappendix{ding2019adaptive} &  & Novograd & \citepappendix{Ginsburg2019} &  \\
                & AdamP/SGDP & \citepappendix{Heo2021} &  & NT-SGD & \citepappendix{Zhou2021a} &  \\
                & AdamT & \citepappendix{Zhou2020} &  & Padam & \citepappendix{ChenZTYCG20} &  \\
                & AdamW & \citepappendix{Loshchilov2019a} &  & PAGE & \citepappendix{li2020page} &  \\
                & AdamX & \citepappendix{Tran_2019} &  & PAL & \citepappendix{mutschler2019parabolic} &  \\
                & ADAS & \citepappendix{ADASRepo} &  & PolyAdam & \citepappendix{OrvietoKL19} &  \\
                & AdaS & \citepappendix{Hosseini2020} &  & Polyak & \citepappendix{Polyak1964} &  \\
                & AdaScale & \citepappendix{johnson2020adascale} &  & PowerSGD/PowerSGDM & \citepappendix{VogelsKJ19} &  \\
                & AdaSGD & \citepappendix{wang2020adasgd} &  & Probabilistic Polyak & \citepappendix{Roos2021} &  \\
                & AdaShift & \citepappendix{zhou2018adashift} &  & ProbLS & \citepappendix{Mahsereci2017} &  \\
                & AdaSqrt & \citepappendix{hu2019secondorder} &  & PStorm & \citepappendix{xu2020momentumbased} &  \\
                & Adathm & \citepappendix{Sun2019} &  & QHAdam/QHM & \citepappendix{MaY19} &  \\
                & AdaX/AdaX-W & \citepappendix{li2020adax} &  & RAdam & \citepappendix{Liu2019} &  \\
                & AEGD & \citepappendix{Liu2020c} &  & Ranger & \citepappendix{Ranger} &  \\
                & ALI-G & \citepappendix{berrada2019training} &  & RangerLars & \citepappendix{RangerLars} &  \\
                & AMSBound & \citepappendix{Luo2019} &  & RMSProp & \citepappendix{Tieleman2012} &  \\
                & AMSGrad & \citepappendix{Reddi2018} &  & RMSterov & \citepappendix{Choi2019} &  \\
                & AngularGrad & \citepappendix{Roy2021} &  & S-SGD & \citepappendix{Sung2020} &  \\
                & ArmijoLS & \citepappendix{Vaswani2019a} &  & SAdam & \citepappendix{Wang2019} &  \\
                & ARSG & \citepappendix{chen2019adaptive} &  & Sadam/SAMSGrad & \citepappendix{tong2019calibrating} &  \\
                & ASAM & \citepappendix{Kwon2021} &  & SALR & \citepappendix{Yue2020} &  \\
                & AutoLRS & \citepappendix{Jin2021} &  & SAM & \citepappendix{Foret2021} &  \\
                & AvaGrad & \citepappendix{savarese2019domainindependent} &  & SC-Adagrad/SC-RMSProp & \citepappendix{MukkamalaH17} &  \\
                & BAdam & \citepappendix{salas2018practical} &  & SDProp & \citepappendix{IdaFI17} &  \\
                & BGAdam & \citepappendix{bai2019bgadam} &  & SGD & \citepappendix{Robbins1951} &  \\
                & BPGrad & \citepappendix{Zhang2017} &  & SGD-BB & \citepappendix{TanMDQ16} &  \\
                & BRMSProp & \citepappendix{aitchison2018bayesian} &  & SGD-G2 & \citepappendix{ayadi2020} &  \\
                & BSGD & \citepappendix{hu2020biased} &  & SGDEM & \citepappendix{RamezaniKebrya2021} &  \\
                & C-ADAM & \citepappendix{tutunov2020compositional} &  & SGDHess & \citepappendix{Tran2021} &  \\
                & CADA & \citepappendix{Chen2021} &  & SGDM & \citepappendix{liu2020new} &  \\
                & Cool Momentum & \citepappendix{oleks2020coolmomentum} &  & SGDR & \citepappendix{Loshchilov2017} &  \\
                & CProp & \citepappendix{Preechakul2019} &  & SHAdagrad & \citepappendix{huang2020adaptive} &  \\
                & Curveball & \citepappendix{Henriques2018} &  & Shampoo & \citepappendix{Anil2020,0001KS18} &  \\
                & Dadam & \citepappendix{nazari2019dadam} &  & SignAdam++ & \citepappendix{WangLTSLSJ19} &  \\
                & DeepMemory & \citepappendix{DeepMemory} &  & SignSGD & \citepappendix{BernsteinWAA18} &  \\
                & DGNOpt & \citepappendix{Liu2021} &  & SKQN/S4QN & \citepappendix{yang2020structured} &  \\
                & DiffGrad & \citepappendix{Dubey2020} &  & SM3 & \citepappendix{Anil2019} &  \\
                & EAdam & \citepappendix{Yuan2020} &  & SMG & \citepappendix{Tran2020} &  \\
                & EKFAC & \citepappendix{George2018} &  & SNGM & \citepappendix{zhao2020stochastic} &  \\
                & Eve & \citepappendix{Hayashi2018} &  & SoftAdam & \citepappendix{fetterman2019} &  \\
                & Expectigrad & \citepappendix{Daley2020} &  & SRSGD & \citepappendix{wang2020scheduled} &  \\
                & FastAdaBelief & \citepappendix{Zhou2021} &  & Step-Tuned SGD & \citepappendix{Castera2021} &  \\
                & FRSGD & \citepappendix{Wang2020} &  & SWATS & \citepappendix{Keskar2017} &  \\
                & G-AdaGrad & \citepappendix{Chakrabarti2021} &  & SWNTS & \citepappendix{chen2019} &  \\
                & GADAM & \citepappendix{zhang2018gadam} &  & TAdam & \citepappendix{ilboudo2020} &  \\
                & Gadam & \citepappendix{Granziol2020} &  & TEKFAC & \citepappendix{Gao2020} &  \\
                & GOALS & \citepappendix{Chae2021} &  & VAdam & \citepappendix{KhanNTLGS18} &  \\
                & GOLS-I & \citepappendix{kafka2019gradientonly} &  & VR-SGD & \citepappendix{ShangZLCTZTJ20} &  \\
                & Grad-Avg & \citepappendix{Purkayastha2020} &  & vSGD-b/vSGD-g/vSGD-l & \citepappendix{SchaulZL13} &  \\
                & GRAPES & \citepappendix{Dellaferrera2021} &  & vSGD-fd & \citepappendix{Schaull3} &  \\
                & Gravilon & \citepappendix{kelterborn2020gravilon} &  & WNGrad & \citepappendix{wu18} &  \\
                & Gravity & \citepappendix{Bahrami2021} &  & YellowFin & \citepappendix{ZhangMR17} &  \\
                & HAdam & \citepappendix{jiang2019higherorder} &  & Yogi & \citepappendix{ZaheerRSKK18} &  \\
                \bottomrule
        \end{tabularx}
\end{tiny}
\end{table}

\clearpage

\begin{table}[H]
	\caption{Overview of commonly used parameter schedules. Note, while we list the schedules parameters, it isn't clearly defined what aspects of a schedule are (tunable) parameters and what is a-priori fixed. In this column, $\alpha_0$ denotes the initial learning rate, $\alpha_{\text{lo}}$ and $\alpha_{\text{up}}$ the lower and upper bound,  $\Delta t$ indicates an epoch count at which to switch decay styles, $k$ denotes a decaying factor.}
	\label{tab:LRS}
	\centering
	\begin{scriptsize}
	\begin{tabularx}{.9\textwidth}{lXrYX}
		\toprule
		\multicolumn{2}{l}{\textbf{Name}} &  \textbf{Ref.} & \textbf{Illustration} & \textbf{Parameters} \\
		\midrule
		Constant & &  &\begin{tikzpicture}[baseline=(current bounding box.center),
  declare function={
    func(\x)= 0.2
   ;
  }
]
\begin{axis}[
hide axis,
axis background/.style={fill=TUgray!10},
width=\figurewidth,height=\figureheight,
ymin=-0.1, ymax=1.1,
xmin=0, xmax=100,
domain=0:100,samples=101,
]

\addplot [TUred,very thick] {func(x)};
\end{axis}
\end{tikzpicture}  & $\alpha_{0}$  \\
		\addlinespace
		Step Decay & constant factor & & \begin{tikzpicture}[baseline=(current bounding box.center),
  declare function={
    func(\x)= (\x < 25) * (1.0)   +
    			and(\x >= 25, \x < 50) * (0.5)     +
    			and(\x >= 50, \x < 75) * (0.25)     +
    			(\x >= 75) * (0.125)
   ;
  }
]
\begin{axis}[
  hide axis,
  axis background/.style={fill=TUgray!10},
  width=\figurewidth,height=\figureheight,
  ymin=-0.1, ymax=1.1,
  xmin=0, xmax=100,
  domain=0:100,samples=101,
]

\addplot [TUred,very thick, domain=0:24] {func(x)};
\addplot [TUred,very thick, domain=25:49] {func(x)};
\addplot [TUred,very thick, domain=50:74] {func(x)};
\addplot [TUred,very thick, domain=75:100] {func(x)};
\end{axis}
\end{tikzpicture}  & $\alpha_{0}$, $\Delta t_1, \dots$, $k$  \\
		& multi-step & & \begin{tikzpicture}[baseline=(current bounding box.center),
  declare function={
    func(\x)= (\x < 50) * (1.0)   +
    			and(\x >= 50, \x < 80) * (0.2)     +
    			(\x >= 80) * (0.1)
   ;
  }
]
\begin{axis}[
  hide axis,
  axis background/.style={fill=TUgray!10},
  width=\figurewidth,height=\figureheight,
  ymin=-0.1, ymax=1.1,
  xmin=0, xmax=100,
  domain=0:100,samples=101,
]

\addplot [TUred,very thick, domain=0:49] {func(x)};
\addplot [TUred,very thick, domain=50:79] {func(x)};
\addplot [TUred,very thick, domain=80:100] {func(x)};
\end{axis}
\end{tikzpicture}  & $\alpha_{0}, \Delta t_1, \dots$, $k_1, \dots$  \\
		\addlinespace

		Smooth Decay & linear decay & \eg \citepappendix{Goodfellow2016} & \begin{tikzpicture}[baseline=(current bounding box.center),
  declare function={
    func(\x)=(\x<60) * (1.0 * (1 - \x/60) + \x/60*0.2) +
    		(\x>=60) * 0.2
   ;
  }
]
\begin{axis}[
  hide axis,
  axis background/.style={fill=TUgray!10},
  width=\figurewidth,height=\figureheight,
  ymin=-0.1, ymax=1.1,
  xmin=0, xmax=100,
  domain=0:100,samples=101,
]

\addplot [TUred,very thick] {func(x)};
\end{axis}
\end{tikzpicture}  & $\alpha_{0}$, ($\Delta t$, $\alpha_{\text{lo}}$)  \\
		& polynomial decay & & \begin{tikzpicture}[baseline=(current bounding box.center),
  declare function={
	func(\x) = (1 - \x / 100) ^ (2)
   ;
  }
]
\begin{axis}[
  hide axis,
  axis background/.style={fill=TUgray!10},
  width=\figurewidth,height=\figureheight,
  ymin=-0.1, ymax=1.1,
  xmin=0, xmax=100,
  domain=0:100,samples=101,
]

\addplot [TUred,very thick] {func(x)};
\end{axis}
\end{tikzpicture} & $\alpha_{0}$, $k$, ($\alpha_{\text{lo}}$)  \\
		& exponential decay & & \begin{tikzpicture}[baseline=(current bounding box.center),
  declare function={
    func(\x)= 1.0 * 0.01^(\x/100)
   ;
  }
]
\begin{axis}[
  hide axis,
  axis background/.style={fill=TUgray!10},
  width=\figurewidth,height=\figureheight,
  ymin=-0.1, ymax=1.1,
  xmin=0, xmax=100,
  domain=0:100,samples=101,
]

\addplot [TUred,very thick] {func(x)};
\end{axis}
\end{tikzpicture}  & $\alpha_{0}$, $k$, ($\alpha_{\text{lo}}$)  \\
		& inverse time decay & \eg \citepappendix{Bottou2012} & \begin{tikzpicture}[baseline=(current bounding box.center),
  declare function={
    func(\x)= 1.0/(1 + 50.0*(\x/100.0))
   ;
  }
]
\begin{axis}[
  hide axis,
  axis background/.style={fill=TUgray!10},
  width=\figurewidth,height=\figureheight,
  ymin=-0.1, ymax=1.1,
  xmin=0, xmax=100,
  domain=0:100,samples=101,
]

\addplot [TUred,very thick] {func(x)};
\end{axis}
\end{tikzpicture} & $\alpha_{0}$, $k$, ($\alpha_{\text{lo}}$)   \\
		& cosine decay & \citepappendix{Loshchilov2017} & \begin{tikzpicture}[baseline=(current bounding box.center),
  declare function={
    func(\x)= 0.5*(1 + cos(deg(\x/100 * pi)))
   ;
  }
]
\begin{axis}[
  hide axis,
  axis background/.style={fill=TUgray!10},
  width=\figurewidth,height=\figureheight,
  ymin=-0.1, ymax=1.1,
  xmin=0, xmax=100,
  domain=0:100,samples=100,
]

\addplot [TUred,very thick] {func(x)};
\end{axis}
\end{tikzpicture}  & $\alpha_{0}$, ($\alpha_{\text{lo}}$) \\
		& linear cosine decay & \citepappendix{Bello2017} & \begin{tikzpicture}[baseline=(current bounding box.center),
  declare function={
  	linear(\x)= 1 - \x/100;
  	cosdecay(\x)= 0.5*(1 + cos(deg(\x/100 * pi)))  ;
    func(\x)= linear(\x) * cosdecay(\x) ;
  }
]
\begin{axis}[
  hide axis,
  axis background/.style={fill=TUgray!10},
  width=\figurewidth,height=\figureheight,
  ymin=-0.1, ymax=1.1,
  xmin=0, xmax=100,
  domain=0:100,samples=100,
]

\addplot [TUred,very thick] {func(x)};
\end{axis}
\end{tikzpicture}  & $\alpha_{0}$, ($\alpha_{\text{lo}}$)  \\ 
		\addlinespace

		Cyclical & triangular & \citepappendix{Smith2017} & \begin{tikzpicture}[baseline=(current bounding box.center),
declare function={
	cycle(\x) = floor(1 + \x/(2*12.5));
	xfunc(\x) = abs(\x/12.5 - 2 *(cycle(\x)) + 1);
	func(\x)= 0.0 + (1.0 - 0.0) * max(0, 1-xfunc(\x))
	;
}
]
\begin{axis}[
hide axis,
axis background/.style={fill=TUgray!10},
width=\figurewidth,height=\figureheight,
ymin=-0.1, ymax=1.1,
xmin=0, xmax=100,
domain=0:100,samples=101,
]

/\addplot [TUred,very thick] {func(x)};
\end{axis}
\end{tikzpicture} & $\alpha_{\text{lo}}, \alpha_{\text{up}}, \Delta t$ \\
		& triangular\newline + decay & \citepappendix{Smith2017} & \begin{tikzpicture}[baseline=(current bounding box.center),
declare function={
	cycle(\x) = floor(1 + \x/(2*12.5));
	xfunc(\x) = abs(\x/12.5 - 2 *(cycle(\x)) + 1);
	func(\x)= 0.0 + (1.0 - 0.0) * max(0, 1-xfunc(\x))/(2^(cycle(\x)-1))
	;
}
]
\begin{axis}[
hide axis,
axis background/.style={fill=TUgray!10},
width=\figurewidth,height=\figureheight,
ymin=-0.1, ymax=1.1,
xmin=0, xmax=100,
domain=0:100,samples=101,
]

/\addplot [TUred,very thick] {func(x)};
\end{axis}
\end{tikzpicture} & $\alpha_{\text{lo}}, \alpha_{\text{up}}, \Delta t, k$    \\
		& triangular\newline + exponential decay & \citepappendix{Smith2017} & \begin{tikzpicture}[baseline=(current bounding box.center),
declare function={
	cycle(\x) = floor(1 + \x/(2*12.5));
	xfunc(\x) = abs(\x/12.5 - 2 *(cycle(\x)) + 1);
	func(\x)= 0.0 + (1.0 - 0.0) * max(0, 1-xfunc(\x))*0.99^(\x)
	;
}
]
\begin{axis}[
hide axis,
axis background/.style={fill=TUgray!10},
width=\figurewidth,height=\figureheight,
ymin=-0.1, ymax=1.1,
xmin=0, xmax=100,
domain=0:100,samples=101,
]

/\addplot [TUred,very thick] {func(x)};
\end{axis}
\end{tikzpicture} & $\alpha_{\text{lo}}, \alpha_{\text{up}}, \Delta t$   \\
		& cosine \newline + warm restarts & \citepappendix{Loshchilov2017} & \begin{tikzpicture}[baseline=(current bounding box.center),
declare function={
	func(\x)= (\x < 25) * (0.5*(1 + cos(deg(\x/25 * pi)))) +
		  and (\x >= 25, \x <50) * (0.5*(1 + cos(deg((\x-25)/25 * pi)))) + 
		  and (\x >= 50, \x <75) * (0.5*(1 + cos(deg((\x-50)/25 * pi)))) + 
		  	  (\x >= 75) * (0.5*(1 + cos(deg((\x-75)/25 * pi))))
	;
}
]
\begin{axis}[
hide axis,
axis background/.style={fill=TUgray!10},
width=\figurewidth,height=\figureheight,
ymin=-0.1, ymax=1.1,
xmin=0, xmax=100,
domain=0:100,samples=101,
]

\addplot [TUred,very thick, domain=0:24] {func(x)};
\addplot [TUred,very thick, domain=25:49] {func(x)};
\addplot [TUred,very thick, domain=50:74] {func(x)};
\addplot [TUred,very thick, domain=75:100] {func(x)};
\end{axis}
\end{tikzpicture}

 & $\alpha_{\text{up}}, \Delta t$, ($\alpha_{\text{lo}}$)  \\
		& cosine \newline + warm restarts \newline + decay & \citepappendix{Loshchilov2017} & \begin{tikzpicture}[baseline=(current bounding box.center),
declare function={
	func(\x)= (\x < 14) * (0.5*(1 + cos(deg(\x/14 * pi)))) +
		  and (\x >= 14, \x <42) * 0.5*(0.5*(1 + cos(deg((\x-14)/28 * pi)))) + 
			  (\x >= 42) * 0.25*(0.5*(1 + cos(deg((\x-42)/58 * pi))))
	;
}
]
\begin{axis}[
hide axis,
axis background/.style={fill=TUgray!10},
width=\figurewidth,height=\figureheight,
ymin=-0.1, ymax=1.1,
xmin=0, xmax=100,
domain=0:100,samples=101,
]

\addplot [TUred,very thick, domain=0:13] {func(x)};
\addplot [TUred,very thick, domain=14:41] {func(x)};
\addplot [TUred,very thick, domain=42:100] {func(x)};
\end{axis}
\end{tikzpicture} & $\alpha_{\text{up}}, \Delta t$, $k$, ($\alpha_{\text{lo}}$)   \\
		\addlinespace

		Warmup & constant warmup & \eg \citepappendix{He2016} & \begin{tikzpicture}[baseline=(current bounding box.center),
  declare function={
    func(\x)= (\x < 20) * (0.2)   +
    			(\x >= 20) * (0.5)
   ;
  }
]
\begin{axis}[
  hide axis,
  axis background/.style={fill=TUgray!10},
  width=\figurewidth,height=\figureheight,
  ymin=-0.1, ymax=1.1,
  xmin=0, xmax=100,
  domain=0:100,samples=101,
]

\addplot [TUred,very thick, domain=0:19] {func(x)};
\addplot [TUred,very thick, domain=20:100] {func(x)};
\end{axis}
\end{tikzpicture}  & $\alpha_{\text{lo}}, \alpha_{0}, \Delta t$    \\
		& gradual warmup  & \citepappendix{goyal2017accurate} & \begin{tikzpicture}[baseline=(current bounding box.center),
  declare function={
    func(\x)= (\x < 20) * (\x * 0.04)   +
    			(\x >= 20) * (0.8)
   ;
  }
]
\begin{axis}[
  hide axis,
  axis background/.style={fill=TUgray!10},
  width=\figurewidth,height=\figureheight,
  ymin=-0.1, ymax=1.1,
  xmin=0, xmax=100,
  domain=0:100,samples=101,
]

\addplot [TUred,very thick] {func(x)};
\end{axis}
\end{tikzpicture}  & $\alpha_{0}, \Delta t, (\alpha_{\text{lo}})$   \\
		& gradual warmup 	\newline + multi-step decay & \citepappendix{goyal2017accurate}  & \begin{tikzpicture}[baseline=(current bounding box.center),
declare function={
	func(\x)= (\x < 20) * (\x * 0.04)   +
	and(\x >= 20, \x < 50) * (0.8)     +
	and(\x >= 50, \x < 75) * (0.4)     +
	(\x >= 75) * (0.2)
	;
}
]
\begin{axis}[
  hide axis,
  axis background/.style={fill=TUgray!10},
  width=\figurewidth,height=\figureheight,
  ymin=-0.1, ymax=1.1,
  xmin=0, xmax=100,
  domain=0:100,samples=101,
]

\addplot [TUred,very thick, domain=0:49] {func(x)};
\addplot [TUred,very thick, domain=50:74] {func(x)};
\addplot [TUred,very thick, domain=75:100] {func(x)};
\end{axis}
\end{tikzpicture}  & $\alpha_{0}, \Delta t, \Delta t_{\text{steps}}, k_1, \dots , (\alpha_{\text{lo}})$    \\
		& gradual warmup 	\newline + step number decay & \citepappendix{Vaswani2017} & \begin{tikzpicture}[baseline=(current bounding box.center),
declare function={
	func(\x) = 4 * min(\x^(-0.5),\x*20^(-1.5))
	;
}
]
\begin{axis}[
  hide axis,
  axis background/.style={fill=TUgray!10},
  width=\figurewidth,height=\figureheight,
  ymin=-0.1, ymax=1.1,
  xmin=0, xmax=100,
  domain=0:100,samples=101,
]

\addplot [TUred,very thick] {func(x)};
\end{axis}
\end{tikzpicture}  & $\alpha_{0}, \Delta t, (\alpha_{\text{lo}})$   \\
		& slanted triangular & \citepappendix{Howard2018} & \begin{tikzpicture}[baseline=(current bounding box.center),
declare function={
	p(\x) = (\x < 20) * (\x/20) +
			(\x>= 20) * (1 - (\x - 20)/(20*(100/20 - 1)));
	func(\x) = 0.8 * (1+p(\x)*(32-1))/32
	;
}
]
\begin{axis}[
  hide axis,
  axis background/.style={fill=TUgray!10},
  width=\figurewidth,height=\figureheight,
  ymin=-0.1, ymax=1.1,
  xmin=0, xmax=100,
  domain=0:100,samples=101,
]

\addplot [TUred,very thick] {func(x)};
\end{axis}
\end{tikzpicture}  & $\alpha_{0}, \Delta t, (\alpha_{\text{lo}})$    \\
		& long trapezoid & \citepappendix{Xing2018} & \begin{tikzpicture}[baseline=(current bounding box.center),
declare function={
	p(\x) = (\x < 20) * (\x/20) +
		  and (\x>= 20, \x < 80) * (1) +
		  (\x >= 80) * (1 - (\x - 80)/(80*(100/80 - 1)));
	func(\x) = 0.8 * (1+p(\x)*(32-1))/32;
}
]
\begin{axis}[
  hide axis,
  axis background/.style={fill=TUgray!10},
  width=\figurewidth,height=\figureheight,
  ymin=-0.1, ymax=1.1,
  xmin=0, xmax=100,
  domain=0:100,samples=101,
]

\addplot [TUred,very thick] {func(x)};
\end{axis}
\end{tikzpicture}  & $\alpha_{\text{0}}, \Delta t_{\text{up}}, \Delta t_{\text{down}}, (\alpha_{\text{lo}})$  \\
		\addlinespace
		Super-Convergence & 1cycle & \citepappendix{Smith2017super} & \begin{tikzpicture}[baseline=(current bounding box.center),
  declare function={
    func(\x)= (\x < 40) * (\x * 0.02 + 0.1)   +
    			and(\x >= 40, \x < 80) * (-0.02 * \x + 1.7)   +
    			(\x >= 80) * (\x * -0.005 + 0.5)
   ;
  }
]
\begin{axis}[
  hide axis,
  axis background/.style={fill=TUgray!10},
  width=\figurewidth,height=\figureheight,
  ymin=-0.1, ymax=1.1,
  xmin=0, xmax=100,
  domain=0:100,samples=101,
]

\addplot [TUred,very thick] {func(x)};
\end{axis}
\end{tikzpicture}  & $\alpha_{\text{up}}, \Delta t, \Delta t_{\text{cutoff}}, (\alpha_{\text{lo}})$    \\
		\bottomrule
	\end{tabularx}
\end{scriptsize}
\end{table}
\clearpage

\section{List of optimizers selected}
\begin{table}[H]
	\caption{Selected optimizers for our benchmarking process with their respective color, hyperparameters, default values, tuning distributions and scheduled hyperparameters. Here, $\mathcal{LU}(\cdot, \cdot)$ denotes the log-uniform distribution while $\mathcal{U}\{ \cdot, \cdot\}$ denotes the discrete uniform distribution.} 
	\label{tab:selected_optimizers}
	\centering
	\begin{scriptsize}
	\begin{tabularx}{.95\textwidth}{lrYYYY}
		\toprule
		\textbf{Optimizer}      & \textbf{Ref.}                 & \textbf{Parameters} & \textbf{Default} & \textbf{Tuning Distribution} & \textbf{Scheduled} \\ \midrule
		 \colordot{amsbound} \textbf{\amsbound}      &    \citepappendix{Luo2019}              & $\alpha$            &    $10^{-3}$              &      $\mathcal{LU}(10^{-4},1)$                        & \ding{51}                   \\
		&& $\alpha_l$          &       $0.1$           &    $\mathcal{LU}(10^{-3}, 0.5)$                          &                    \\
		&& $\beta_1$           &      $0.9$            &      $\mathcal{LU}(0.5, 0.999)$                        &                    \\
		&& $\beta_2$           &      $0.999$           &    $\mathcal{LU}(0.8, 0.999)$                          &                    \\
		&& $\gamma$            &     $10^{-3}$           &      $\mathcal{LU}(10^{-4}, 10^{-1})$                         &                    \\
		&& $\varepsilon$          &        $10^{-8}$        &        \ding{55}       &                    \\
		\rule{0pt}{3.5ex} \colordot{amsgrad} \textbf{\amsgrad} &   \citepappendix{Reddi2018}                      & $\alpha$            &    $10^{-2}$               &     $\mathcal{LU}(10^{-4},1)$                            & \ding{51}                   \\
		&& $\beta_1$           &    $0.9$                &        $\mathcal{LU}(0.5, 0.999)$                        &                    \\
		&& $\beta_2$           &    $0.999$              &         $\mathcal{LU}(0.8, 0.999)$                     &                    \\
		&& $\varepsilon$          &    $10^{-8}$              &        \ding{55}                      &                    \\
		\rule{0pt}{3.5ex} \colordot{adabelief} \textbf{\adabelief}      &    \citepappendix{Zhuang2020}              & $\alpha$            &    $10^{-3}$              &      $\mathcal{LU}(10^{-4},1)$                        & \ding{51}                   \\
		&& $\beta_1$           &      $0.9$            &      $\mathcal{LU}(0.5, 0.999)$                        &                    \\
		&& $\beta_2$           &      $0.999$           &    $\mathcal{LU}(0.8, 0.999)$                          &                    \\
		&& $\varepsilon$          &        $10^{-14}$        &        \ding{55}       &                    \\
		\rule{0pt}{3.5ex} \colordot{adabound} \textbf{\adabound} &      \citepappendix{Luo2019}                  & $\alpha$            &  $10^{-3}$         &      $\mathcal{LU}(10^{-4},1)$                           & \ding{51}                   \\
		&& $\alpha_l$          &    $0.1$              &       $\mathcal{LU}(10^{-3}, 0.5)$                        &                    \\
		&& $\beta_1$           &    $0.9$              &       $\mathcal{LU}(0.5, 0.999)$                         &                    \\
		&& $\beta_2$           &    $0.999$              &    $\mathcal{LU}(0.8, 0.999)$                          &                    \\
		&& $\gamma$            &    $10^{-3}$              &     $\mathcal{LU}(10^{-4}, 10^{-1})$                         &                    \\
		\rule{0pt}{3.5ex} \colordot{adadelta} \textbf{\adadelta} &    \citepappendix{Zeiler2012}                    & $\alpha$            &   $10^{-3}$                &    $\mathcal{LU}(10^{-4},1)$                             & \ding{51}                   \\
		&& $\varepsilon$          &    $10^{-8}$              &        \ding{55}                      &                    \\
		&& $1-\rho$             &   $0.95$               &  
		$\mathcal{LU}(10^{-4}, 1)$                            
		&                    \\
		\rule{0pt}{3.5ex} \colordot{adagrad} \textbf{\adagrad} &    \citepappendix{Duchi2011}                    & $\alpha$            &    $10^{-2}$               &     $\mathcal{LU}(10^{-4},1)$                            & \ding{51}                   \\
		&& $\varepsilon$          &   $10^{-7}$               &        \ding{55}                      &                    \\
		\rule{0pt}{3.5ex} \colordot{adam} \textbf{\adam}  & \citepappendix{Kingma2015} &  $\alpha$            & $10^{-3}$        & $\mathcal{LU}(10^{-4}, 1)$   & \ding{51}                 \\
		&& $\beta_1$           & $0.9$            & $\mathcal{LU}(0.5, 0.999)$     &                    \\
		&& $\beta_2$           & $0.999$          & $\mathcal{LU}(0.8, 0.999)$  &                    \\
		&& $\varepsilon$          & $10^{-8}$        & \ding{55}                           &                    \\
		\rule{0pt}{3.5ex} \colordot{lookaheadmomentum} \textbf{\textsc{Lookahead}}         &   \citepappendix{Zhang2019a}           & $\alpha$            &  $0.5$                 &     $\mathcal{LU}(10^{-4},1)$                            &                    \\
		\colordot{white} \textbf{\textsc{Momentum}}        &         & $\alpha_f$            &     $10^{-2}$              &      $\mathcal{LU}(10^{-4},1)$                        &    \ding{51}                \\       
		\colordot{white} abbr.~\lamom&& $k$                 &     5             &         $\mathcal{U}\{1,20\}$                   &                    \\
		&& $1-\rho$             &    $0.99$              &      
		$\mathcal{LU}(10^{-4}, 1)$                        &                    
		\\
		\rule{0pt}{3.5ex} \colordot{lookaheadradam} \textbf{\textsc{Lookahead}}    &     \citepappendix{Zhang2019a}              & $\alpha$            &  $0.5$                &     $\mathcal{LU}(10^{-4},1)$                            &  \\
		\colordot{white} \textbf{\textsc{RAdam}}                  &        & $\alpha_f$            &    $10^{-3}$               &         $\mathcal{LU}(1e-4,1)$                      &         \ding{51}                             \\
		\colordot{white} abbr.~\laradam&& $\beta_1$           &    $0.9$              &    $\mathcal{LU}(0.5, 0.999)$                            &                    \\
		&& $\beta_2$           &      $0.999$            &      $\mathcal{LU}(0.8, 0.999)$                         &                    \\
		&& $\varepsilon$          &     $10^{-7}$             &            \ding{55}                  &                    \\
		& & $k$                 &      5            &        $\mathcal{U}\{1,20\}$                      &                    \\
		\rule{0pt}{3.5ex} \colordot{momentum}  \textbf{\momentum}   &      \citepappendix{Polyak1964}                & $\alpha$            &      $10^{-2}$             &     $\mathcal{LU}(10^{-4},1)$                            & \ding{51}                   \\
		&& $1-\rho$             &   $0.99$               &    
		$\mathcal{LU}(10^{-4}, 1)$                          
		&                    \\
		\rule{0pt}{3.5ex} \colordot{nag} \textbf{\nag}     &   \citepappendix{Nesterov1983}                       & $\alpha$            &      $10^{-2}$             &   $\mathcal{LU}(10^{-4},1)$                              & \ding{51}                   \\
		&& $1-\rho$             &     $0.99$             &        
		$\mathcal{LU}(10^{-4}, 1)$                      &                    \\
		\rule{0pt}{3.5ex} \colordot{nadam} \textbf{\nadam}    &   \citepappendix{Dozat2016IncorporatingNM}                    & $\alpha$            &    $10^{-3}$               &      $\mathcal{LU}(10^{-4},1)$                           & \ding{51}                   \\
		&& $\beta_1$           &  $0.9$                &   $\mathcal{LU}(0.5, 0.999)$                             &                    \\
		&& $\beta_2$           &   $0.999$               &    $\mathcal{LU}(0.8, 0.999)$                          &                    \\
		&& $\varepsilon$          &     $10^{-7}$             &        \ding{55}                      &                    \\
		\rule{0pt}{3.5ex} \colordot{radam} \textbf{\radam}    &      \citepappendix{Liu2019}                  & $\alpha$            &   $10^{-3}$                &    $\mathcal{LU}(10^{-4},1)$                             & \ding{51}                   \\
		&& $\beta_1$           &     $0.9$             &    $\mathcal{LU}(0.5, 0.999)$                            &                    \\
		&& $\beta_2$           &      $0.999$            &    $\mathcal{LU}(0.8, 0.999)$                          &                    \\
		&& $\varepsilon$          &     $10^{-7}$             &        \ding{55}                      &                    \\
		\rule{0pt}{3.5ex} \colordot{rmsprop} \textbf{\rmsprop}  &     \citepappendix{Tieleman2012}                  & $\alpha$            &     $10^{-3}$              &     $\mathcal{LU}(10^{-4},1)$                            & \ding{51}                   \\
		&& $\varepsilon$          &   $10^{-10}$               &          \ding{55}                    &                    \\		
		&& $1-\rho$             &     $0.9$             &     
		$\mathcal{LU}(10^{-4}, 1)$                         &                    
		\\
		\rule{0pt}{3.5ex} \colordot{sgd} \textbf{\sgd} &        \citepappendix{Robbins1951}                    & $\alpha$            &     $10^{-2}$              &      $\mathcal{LU}(10^{-4},1)$                           & \ding{51}                   \\ \bottomrule
	\end{tabularx}
\end{scriptsize}
\end{table}

\clearpage

\section{Robustness to random seeds}
\label{sec:FailingSeeds}
Data subsampling, random weight initialization, dropout and other aspects of deep learning introduce stochasticity to the training process.
As such, judging the performance of an optimizer on a single run may be misleading due to random fluctuations.
In our benchmark we use $10$ different seeds of the final setting for each budget in order to judge the stability of the optimizer and the results.
However, to keep the magnitude of this benchmark feasible, we only use a single seed while tuning, analogously to how a single user would progress.
This means that our tuning process can sometimes choose hyperparameter settings which might not even converge for seeds other than the one used for tuning.

\begin{figure}[ht]
	\centering
	\includegraphics[width=.8\textwidth]{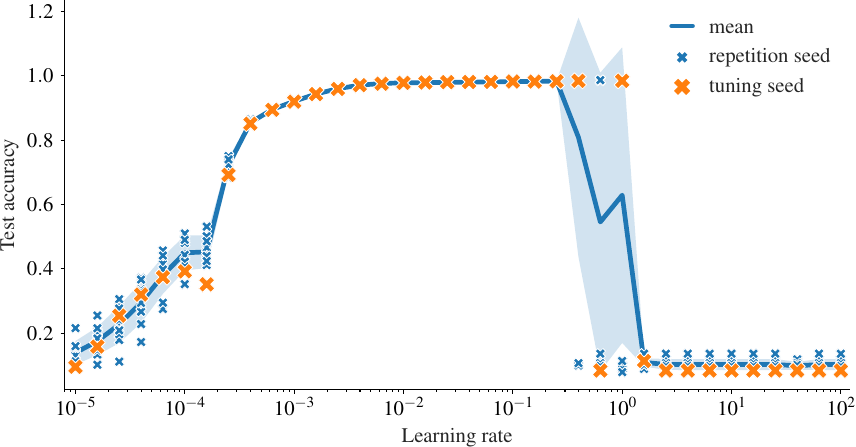}
	\caption{Performance of \sgd on a simple multilayer perceptron. For each learning rate, markers in orange (\textcolor{sns_ora}{\ding{54}}) show the initial seed which would be used for tuning, blue markers (\textcolor{sns_blue}{\ding{54}}) illustrate nine additional seeds with otherwise unchanged settings. The mean over all seeds is plotted as a blue line (\textcolor{sns_blue}{\textbf{---}}), showing one standard deviation as a shaded area (\textcolor{sns_blue_shaded}{\ding{122}}).}
	\label{fig:FailingSeeds}
\end{figure}

\Cref{fig:FailingSeeds} illustrates this behavior on an example problem where we used $10$ seeds throughout a tuning process using grid search.
The figure shows that in the beginning performance increases when increasing the learning rate, followed by an area were it sometimes works but other times diverges.
Picking hyperparameters from this ``danger zone'' can lead to unstable results.
In this case, where we only consider the learning rate, it is clear that decreasing the learning rate a bit to get away from this ``danger zone'' would lead to a more stable, but equally well-performing algorithm.
In more complicated cases, however, we are unable to use a simple heuristic such as this.
This might be the case, for example, when tuning multiple hyperparameters or when the effect of the hyperparameter on the performance is less straight forward.
Thus, this is a problem not created by improperly using the tuning method, but by an unstable optimization method.

In our benchmark, we observe a total of $18$, $24$, and $17$ divergent seeds for the small, medium, and large budget respectively.
This amounts to roughly $0.5\%$ of the runs in each budget.
Most of them occur when using \sgd ($10$, $15$, and $7$ cases for the small, medium and large budget respectively), \adagrad ($5$, $3$, and $5$ cases for the small, medium and large budget respectively) or \adadelta ($3$, $5$, and $3$ cases for the small, medium and large budget respectively), which might indicate that modern adaptive methods are less prone to this kind of behavior.
None of these cases occur when using a constant schedule, and most of them occur when using the \textit{trapezoidal} schedule ($11$, $11$, and $9$ cases for the small, medium and large budget respectively).
However, as our data on diverging seeds is very limited, it is not conclusive enough to draw solid conclusions.
\clearpage

\section{Re-Tuning experiments}
\label{sec:retuning}
\begin{figure}[htbp!]
	\centering
	\includegraphics[width=\textwidth]{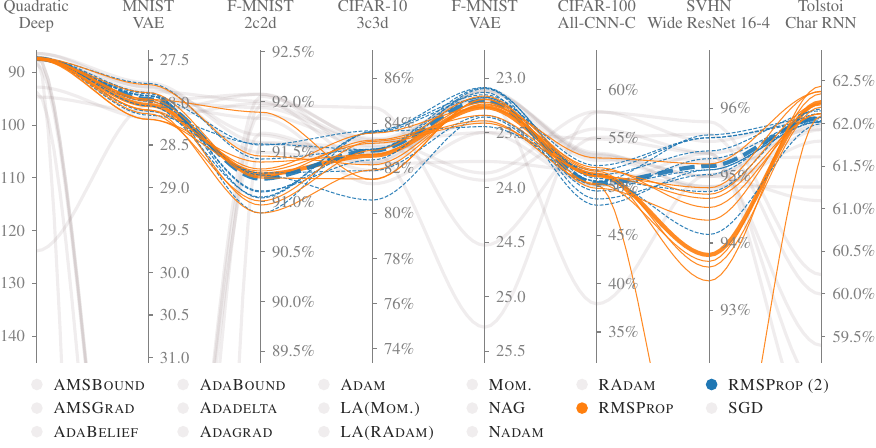}
	\caption{Mean test set performance of all $10$ seeds of \rmsprop (\textcolor{sns_ora}{\textbf{---}}) on all eight optimization problems using the \emph{small budget} for tuning and \emph{no learning rate schedule}. The mean is shown with a thicker line. We repeated the full tuning process on all eight problems using different random seeds, which is shown in dashed lines blue (\textcolor{sns_blue}{\textbf{- -}}). The mean performance of all other optimizers is shown in transparent gray lines.}
	\label{fig:appendix_pc_retuning_rmsprop}
\end{figure}

\begin{figure}[htbp!]
	\centering
	\includegraphics[width=\textwidth]{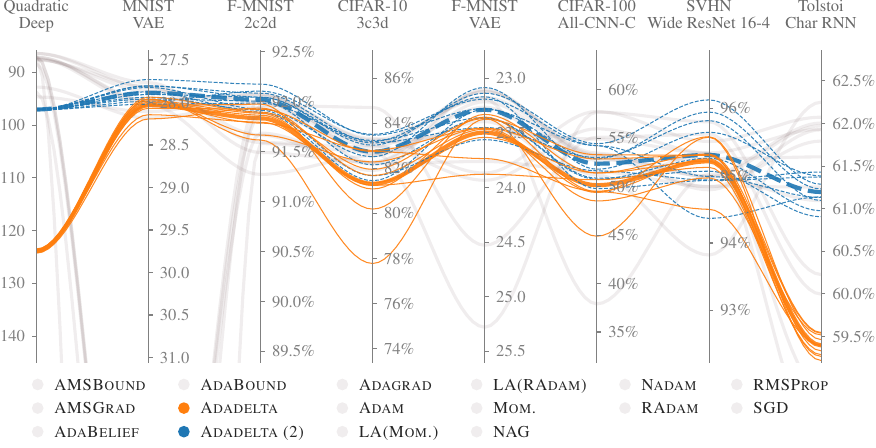}
	\caption{Mean test set performance of all $10$ seeds of \adadelta (\textcolor{sns_ora}{\textbf{---}}) on all eight optimization problems using the \emph{small budget} for tuning and \emph{no learning rate schedule}. The mean is shown with a thicker line. We repeated the full tuning process on all eight problems using different random seeds, which is shown in dashed lines blue (\textcolor{sns_blue}{\textbf{- -}}). The mean performance of all other optimizers is shown in transparent gray lines.}
	\label{fig:appendix_pc_retuning_adadelta}
\end{figure}

In order to test the stability of our benchmark and especially the tuning method, we selected two optimizers in our benchmark and re-tuned them on all problems a second time.
We used completely independent random seeds for both tuning and the $10$ repetitions with the final setting.
\Cref{fig:appendix_pc_retuning_rmsprop} and \Cref{fig:appendix_pc_retuning_adadelta} show the distribution of all $10$ random seeds for both the original tuning as well as the re-tuning runs for \rmsprop and \adadelta.
It is evident, that re-tuning results in a shift of this distribution, since small (stochastic) changes during tuning can result in a different chosen hyperparameter setting.

These differences also highlight how crucial it is to look at multiple problems.
Individually, small changes, such as re-doing the tuning with different seeds can lead to optimization methods changing rankings.
However, they tend to average out when looking at an unbiased list of multiple problems.
These results also further supports the statement made in \Cref{sec:Results} that there is no optimization method clearly domination the competition, as small performance margins might vanish when re-tuning.

\clearpage

\section{List of schedules selected}
\label{sec:schedules}
\begin{figure}[ht]
	\centering
	\includegraphics[width=.84\textwidth]{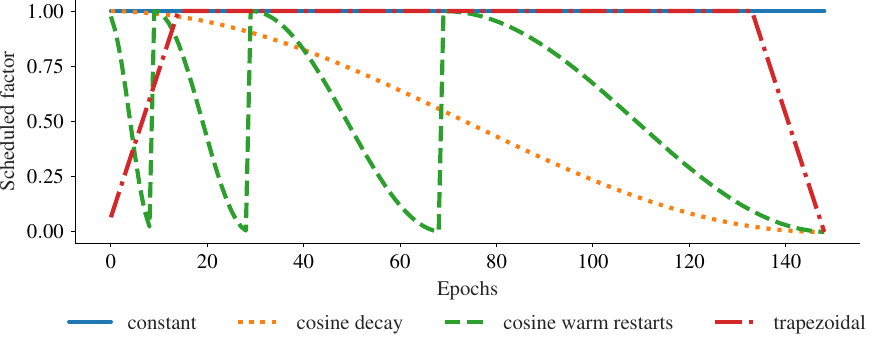}
	\caption{Illustration of the selected learning rate schedules for a training duration of $150$ epochs.}
	\label{fig:schedules}
\end{figure}

The schedules selected for our benchmark are illustrated in \Cref{fig:schedules}. All learning rate schedules are multiplied by the initial learning rate found via tuning or picked as the default choice.

We use a \emph{cosine decay} \citep{Loshchilov2017} that starts at $1$ and decays in the form of a half period of a cosine to $0$.
As an example of a cyclical learning rate schedule, we test a \emph{cosine with warm restarts} schedule with a cycle length $\Delta t =10$ which increases by a factor of $2$ after each cycle without any discount factor.
Depending on the number of epochs we train our model, it is possible that training stops shortly after one of those warm restarts.
Since performance typically declines shortly after increasing the learning rate, we don't report the final performance for this schedule, but instead the performance achieved after the last complete period (just before the next restart).
This approach is suggested by the original work of \citet{Loshchilov2017}.
However, we still use the final performance while tuning.

A representation of a schedule including warm-up is the \emph{trapezoidal} schedule from \citet{Xing2018}.
For our benchmark we set a warm-up and cool-down period of $\nicefrac{1}{10}$ the training time.
\clearpage

\section{ArXiv Mentions}
\label{sec:arxiv_mentions}
\begin{figure}[H]
	\centering
	\includegraphics[width=.59\textwidth]{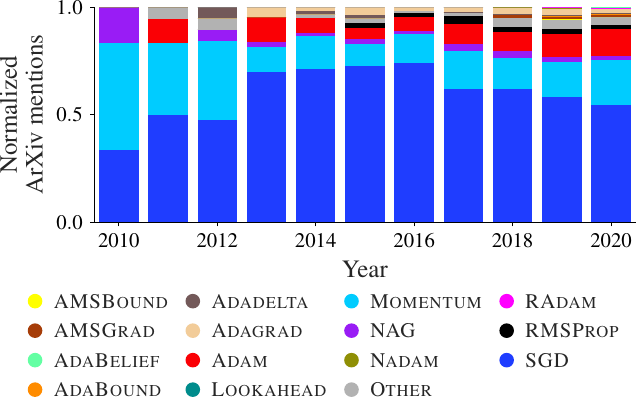}
	\caption{Percentage of times ArXiv titles and abstracts mention specific optimizer per year. This is a normalized version of \Cref{fig:arxiv}. The data for this figure is shown in \Cref{tab:app_arxiv_numbers}.}
	\label{fig:arxiv_normalized}
\end{figure}

\begin{table}[H]
		\caption{Mentions of each optimizer in titles and abstracts of papers on ArXiv per year. All non-selected optimizers from \Cref{tab:Optimizers} in the appendix are grouped into \textit{Other}.}
		\label{tab:app_arxiv_numbers}
		\centering
			\begin{tabularx}{.88\textwidth}{lrrrrrrrrrrr}
			\toprule
			\textbf{Optimizer} &  \textbf{2010} &  \textbf{2011} &  \textbf{2012} &  \textbf{2013} &  \textbf{2014} &  \textbf{2015} &  \textbf{2016} &  \textbf{2017} &  \textbf{2018} &  \textbf{2019} &  \textbf{2020} \\
			\midrule
			\colordot{amsbound} \amsbound              &     0 &     0 &     0 &     0 &     0 &     0 &     0 &     0 &     0 &     1 &     0 \\
			\colordot{amsgrad} \amsgrad                &     0 &     0 &     0 &     0 &     0 &     0 &     0 &     0 &     7 &     9 &    11 \\
			\colordot{adabelief} \adabelief            &     0 &     0 &     0 &     0 &     0 &     0 &     0 &     0 &     0 &     0 &     3 \\
			\colordot{adabound} \adabound              &     0 &     0 &     0 &     0 &     0 &     0 &     0 &     0 &     0 &     4 &     4 \\
			\colordot{adadelta} \adadelta              &     0 &     0 &     1 &     0 &     1 &     2 &     0 &     1 &     2 &     3 &     3 \\
			\colordot{adagrad} \adagrad                &     0 &     0 &     0 &     2 &     1 &     5 &     3 &     8 &    16 &    22 &    24 \\
			\colordot{adam} \adam                      &     0 &     2 &     0 &     5 &     4 &     7 &    11 &    31 &    47 &    83 &   119 \\
			\colordot{lookaheadmomentum} \lookaheadopt &     0 &     0 &     0 &     0 &     0 &     0 &     0 &     0 &     0 &     2 &     1 \\
			\colordot{momentum} \momentum              &     3 &     6 &     7 &     5 &     9 &    14 &    23 &    57 &    76 &   124 &   205 \\
			\colordot{nag} \nag                        &     1 &     0 &     1 &     1 &     1 &     3 &     3 &    11 &    17 &    18 &    19 \\
			\colordot{nadam} \nadam                    &     0 &     0 &     0 &     0 &     0 &     0 &     0 &     0 &     1 &     2 &     0 \\
			\colordot{otheropt} \otheropt              &     0 &     1 &     1 &     0 &     1 &     3 &     2 &     4 &    22 &    34 &    36 \\
			\colordot{radam} \radam                    &     0 &     0 &     0 &     0 &     0 &     0 &     0 &     0 &     0 &     2 &     1 \\
			\colordot{rmsprop} \rmsprop                &     0 &     0 &     0 &     0 &     0 &     3 &     3 &    13 &    13 &    18 &    18 \\
			\colordot{sgd} \sgd                        &     2 &     9 &     9 &    30 &    42 &    98 &   129 &   205 &   326 &   451 &   532 \\
			\bottomrule
		\end{tabularx}
\end{table}
\clearpage

\section{Improvement after tuning}
\label{sec:heatmaps_appendix}
When looking at \Cref{fig:Heatmap}, one might realize that few diagonal entries contain negative values. Since diagonal entries reflect the intra-optimizer performance change when tuning on the respective task, this might feel quite counterintuitive at first. \emph{In theory}, this can occur if the respective tuning distributions is chosen poorly, the tuning randomness simply got ``unlucky'', or  we observe significantly worse results for our additional seeds (see \Cref{fig:FailingSeeds}).

\begin{figure}[ht]
	\centering
	\includegraphics[width=\textwidth]{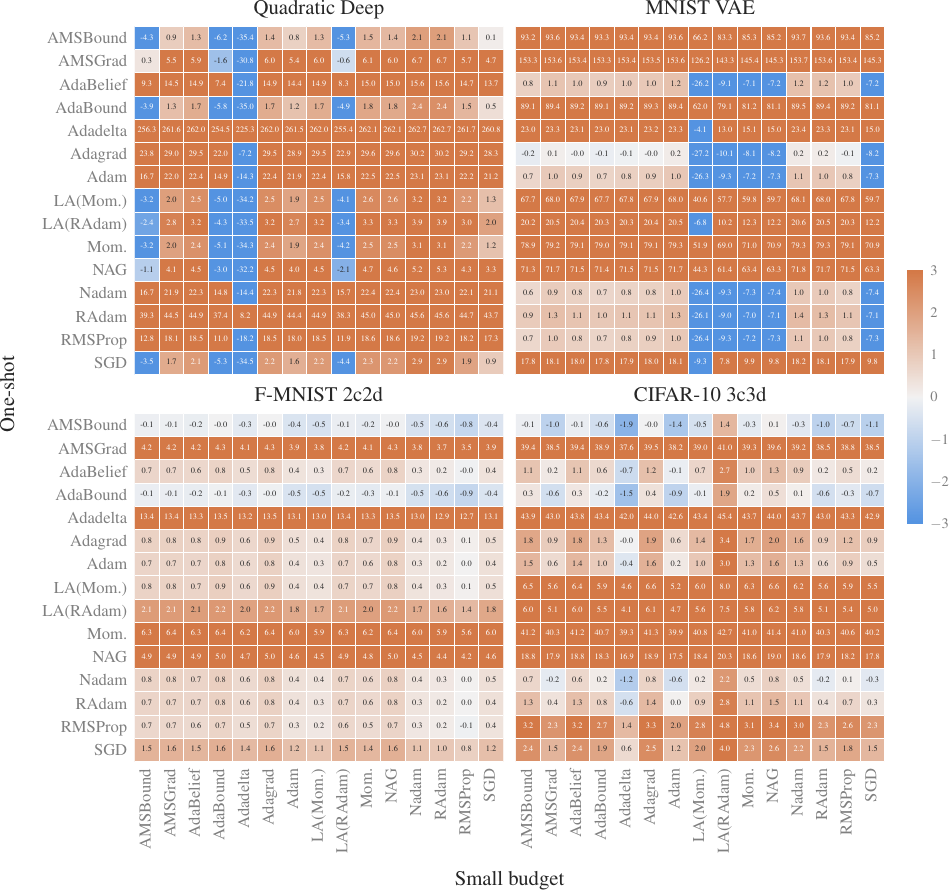}
	\caption{The absolute test set performance improvement after switching from any untuned optimizer ($y$-axis, \textit{one-shot}) to any tuned optimizer ($x$-axis, \textit{small budget}) as an average over $10$ random seeds for the \emph{constant} schedule. This is a detailed version of \Cref{fig:Heatmap} in the main text showing the first four problems.}
	\label{fig:Heatmap__o__s__0}
\end{figure}

\begin{figure}[ht]
	\centering
	\includegraphics[width=\textwidth]{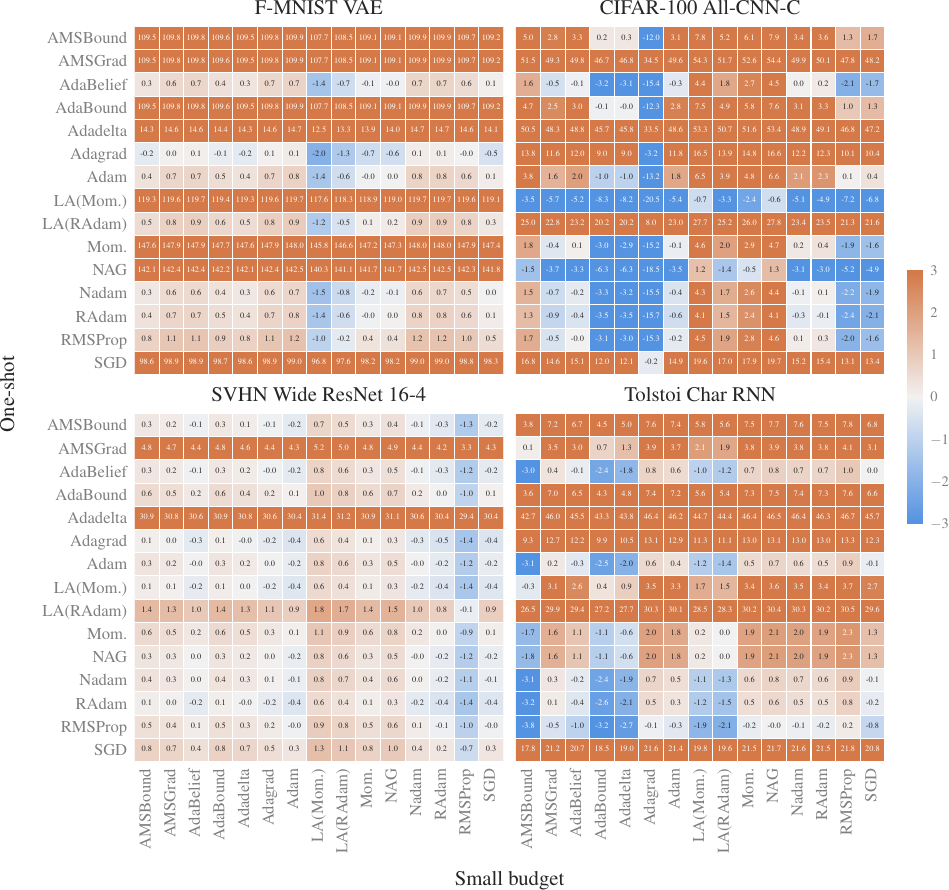}
	\caption{The absolute test set performance improvement after switching from any untuned optimizer ($y$-axis, \textit{one-shot}) to any tuned optimizer ($x$-axis, \textit{small budget}) as an average over $10$ random seeds for the \emph{constant} schedule. This is a detailed version of \Cref{fig:Heatmap} in the main text showing the last four problems.}
	\label{fig:Heatmap__o__s__1}
\end{figure}

\begin{figure}[ht]
	\centering
	\includegraphics[width=\textwidth]{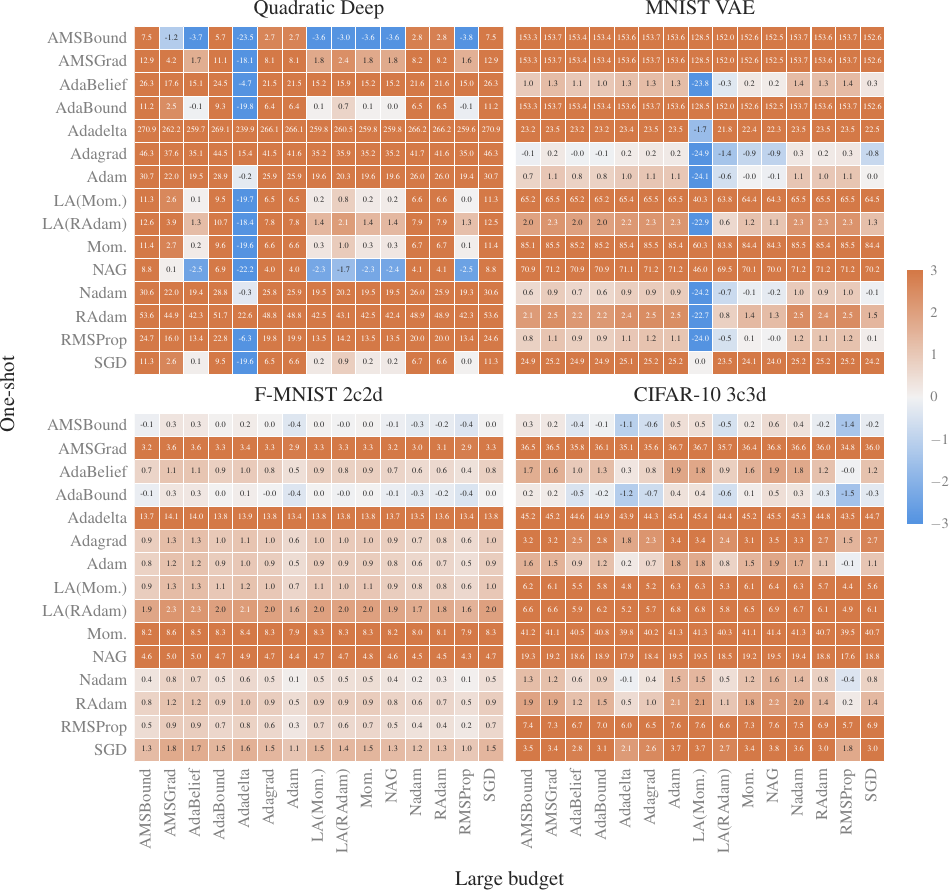}
	\caption{The absolute test set performance improvement after switching from any untuned optimizer ($y$-axis, \textit{one-shot}) to any tuned optimizer ($x$-axis, \textit{large budget}) for the \emph{constant} schedule. This is structurally the same plot as \Cref{fig:Heatmap__o__s__0} but comparing to the \emph{large budget} and only considering the seed that has been used for tuning.}
	\label{fig:Heatmap__o__l__0__TUNING_SEED}
\end{figure}

\begin{figure}[ht]
	\centering
	\includegraphics[width=\textwidth]{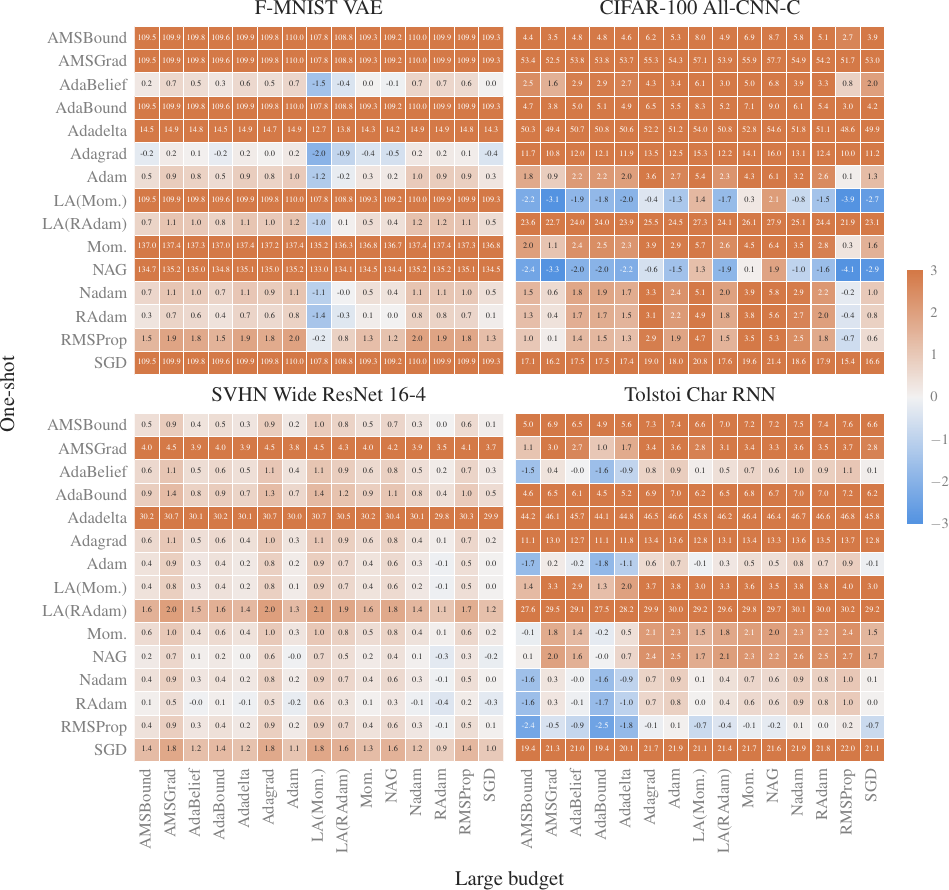}
	\caption{The absolute test set performance improvement after switching from any untuned optimizer ($y$-axis, \textit{one-shot}) to any tuned optimizer ($x$-axis, \textit{large budget}) for the \emph{constant} schedule. This is structurally the same plot as \Cref{fig:Heatmap__o__s__1} but comparing to the \emph{large budget} and only considering the seed that has been used for tuning.}
	\label{fig:Heatmap__o__l__1__TUNING_SEED}
\end{figure}

If we compare \Cref{fig:Heatmap__o__s__0,fig:Heatmap__o__s__1} to \Cref{fig:Heatmap__o__l__0__TUNING_SEED,fig:Heatmap__o__l__1__TUNING_SEED} we can see most negative diagonal entries vanish or at least diminish in magnitude.
For the latter two figures we allow for more tuning runs and only consider the seed that has been used for this tuning process.
The fact that the effect of negative diagonal entries reduces is an indication that they mostly result from the two latter reasons mentioned.

\clearpage

\section{Optimizer performance across test problems}
\label{sec:additional_plots}
\begin{figure}[htbp!]
	\centering
	\includegraphics[width=\textwidth]{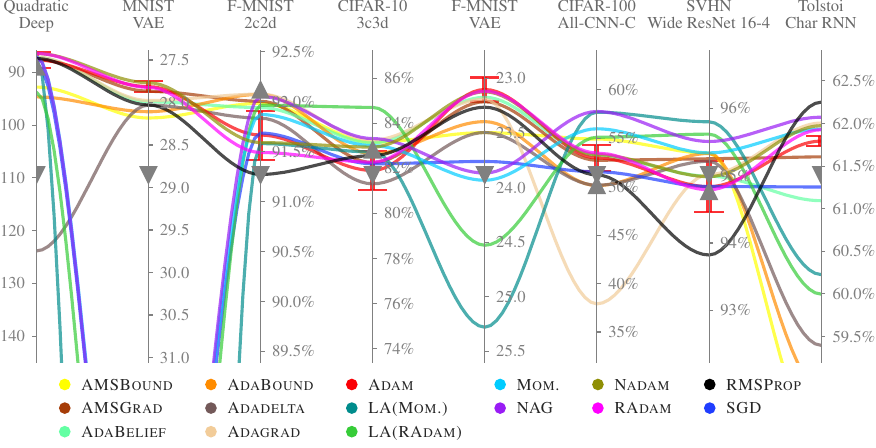}
	\caption{Mean test set performance over $10$ random seeds of all tested optimizers on all eight optimization problems using the \emph{small budget} for tuning and \emph{no learning rate schedule}. One standard deviation for the tuned \adam optimizer is shown with a red error bar (\textcolor{adam}{\textbf{I}}). The performance of the untuned versions of \adam (\textcolor{TUgray}{\ding{116}}) and \adabound (\textcolor{TUgray}{\ding{115}}) are marked for reference. Note, the upper bound of each axis represents the best performance achieved in the benchmark, while the lower bound is chosen in relation to the performance of \adam with default parameters. Tabular version available in the Appendix as \Cref{tab:app_tabular_small_none}.}
	\label{fig:appendix_pc_sb}
\end{figure}

\begin{figure}[htbp!]
	\centering
	\includegraphics[width=\textwidth]{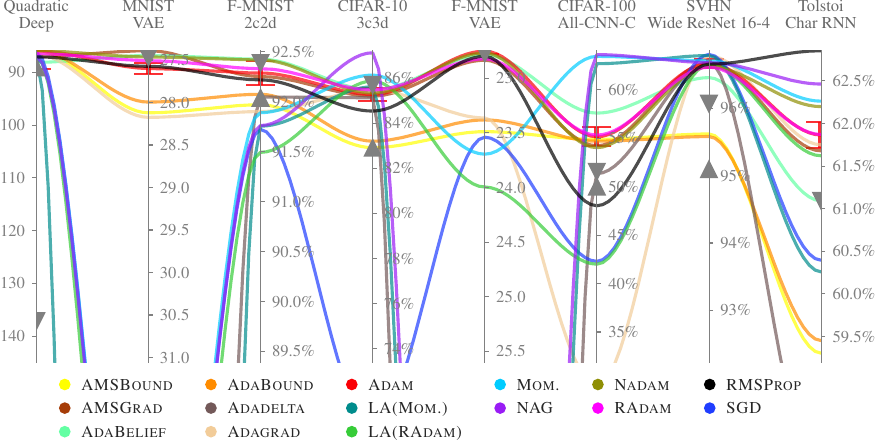}
	\caption{Mean test set performance over $10$ random seeds of all tested optimizers on all eight optimization problems using the \emph{medium budget} for tuning and the \emph{cosine learning rate schedule}. One standard deviation for the tuned \adam optimizer is shown with a red error bar (\textcolor{adam}{\textbf{I}}). The performance of the untuned versions of \adam (\textcolor{TUgray}{\ding{116}}) and \adabound (\textcolor{TUgray}{\ding{115}}) are marked for reference (this time with the \emph{cosine} learning rate schedule). Note, the upper bound of each axis represents the best performance achieved in the benchmark, while the lower bound is chosen in relation to the performance of \adam with default parameters (and no schedule). Tabular version available in the Appendix as \Cref{tab:app_tabular_medium_cosine}.}
	\label{fig:appendix_pc_cos}
\end{figure}

\begin{figure}[htbp!]
	\centering
	\includegraphics[width=\textwidth]{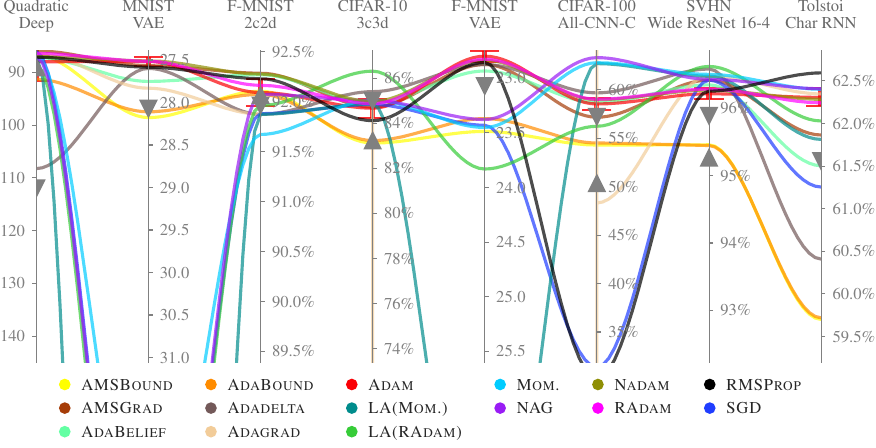}
	\caption{Mean test set performance over $10$ random seeds of all tested optimizers on all eight optimization problems using the \emph{large budget} for tuning and the \emph{trapezoidal learning rate schedule}. One standard deviation for the tuned \adam optimizer is shown with a red error bar (\textcolor{adam}{\textbf{I}}). The performance of the untuned versions of \adam (\textcolor{TUgray}{\ding{116}}) and \adabound (\textcolor{TUgray}{\ding{115}}) are marked for reference (this time with the \emph{trapezoidal} learning rate schedule). Note, the upper bound of each axis represents the best performance achieved in the benchmark, while the lower bound is chosen in relation to the performance of \adam with default parameters (and no schedule). Tabular version available in the Appendix as \Cref{tab:app_tabular_large_ltr}.}
	\label{fig:appendix_pc_ltr}
\end{figure}

\begin{figure}[htbp!]
	\centering
	\includegraphics[width=\textwidth]{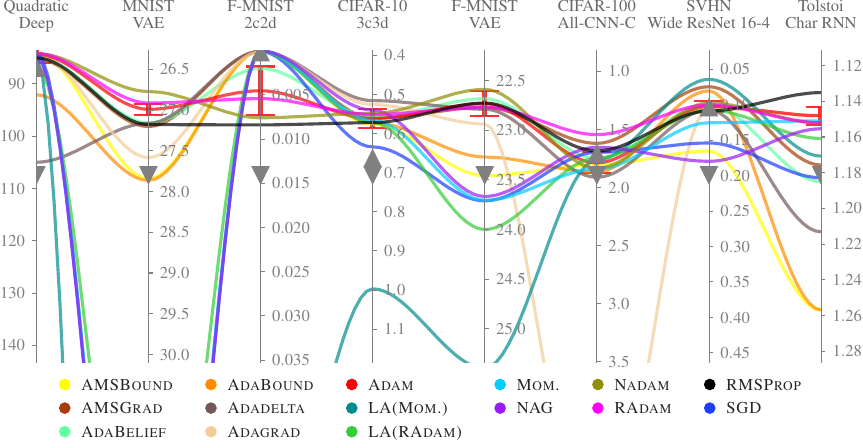}
	\caption{Mean \emph{training} loss performance over $10$ random seeds of all tested optimizers on all eight optimization problems using the \emph{large budget} for tuning and \emph{no learning rate schedule}. One standard deviation for the tuned \adam optimizer is shown with a red error bar (\textcolor{adam}{\textbf{I}}). The performance of the untuned versions of \adam (\textcolor{TUgray}{\ding{116}}) and \adabound (\textcolor{TUgray}{\ding{115}}) are marked for reference. Note, the upper bound of each axis represents the best performance achieved in the benchmark, while the lower bound is chosen in relation to the performance of \adam with default parameters (and no schedule). This figure is very similar to \Cref{fig:ParallelCoordinates}, but showing the \emph{training loss} performance instead of the \emph{test accuracy} (or \emph{test loss} if no accuracy is available). Tabular version available in the Appendix as \Cref{tab:app_tabular_large_none_TRAIN}.}
	\label{fig:appendix_pc_train_loss}
\end{figure}

Similarly to \Cref{fig:ParallelCoordinates}, we show the corresponding plots for the \emph{small budget} with \emph{no learning rate schedule} in \Cref{fig:appendix_pc_sb} and the \emph{medium budget} with the \emph{cosine} and \emph{trapezoidal learning rate schedule} in \Cref{fig:appendix_pc_cos,fig:appendix_pc_ltr}. Additionally, in \Cref{fig:appendix_pc_train_loss} we show the same setting as \Cref{fig:ParallelCoordinates} but showing the training loss instead of the test loss/accuracy.

The high-level trends mentioned in \Cref{sec:Results} also hold for the smaller tuning budget in \Cref{fig:appendix_pc_sb}. Namely, taking the winning optimizer for several untuned algorithms (here marked for \adam and \adabound) will result in a decent performance in most problems with much less effort.
Adding a tuned version \adam (or variants thereof) to this selection would result in a very competitive performance.
The absolute top-performance however, is achieved by changing optimizers across different problems.

Note, although the \emph{medium budget} is a true superset of the \emph{small budget} it is not given that it will always perform better.
Our tuning procedure guarantees that the \emph{validation} performance on the seed that has been used for tuning is as least as good on the medium budget than on the small budget.
But due to averaging over multiple seeds and reporting \emph{test} performance instead of \emph{validation} performance, this hierarchy is no longer guaranteed.
We discuss the possible effects of averaging over multiple seeds further in \Cref{sec:FailingSeeds}.

The same high-level trends also emerge when considering the \emph{cosine} or \emph{trapezoidal learning rate schedule} in \Cref{fig:appendix_pc_cos,fig:appendix_pc_ltr}.
We can also see that the top performance in general increase when adding a schedule (cf. \Cref{fig:ParallelCoordinates} and \Cref{fig:appendix_pc_ltr}).

Comparing \Cref{fig:ParallelCoordinates} and \Cref{fig:appendix_pc_train_loss} we can assess the generalization performance of the optimization method not only to an unseen test set, but also to a different performance metric (accuracy instead of loss).
Again, the overall picture of varying performance across different problems remains consistent when considering the training loss performance.
Similarily to the figures showing test set performance we cannot identify a clear winner, although \adam ands its variants, such as \radam perform near the top consistently.
Note that while \Cref{fig:appendix_pc_train_loss} shows the training loss, the optimizers have still be tuned to achieve the best validation performance (\ie accuracy if available, else the loss).
\clearpage

\section{Tabular version}
\begin{table}[H]
        \caption{Tabular version of \Cref{fig:ParallelCoordinates}. Mean test set performance and standard deviation over $10$ random seeds of all tested optimizers on all eight optimization problems using the \emph{large budget} for tuning and \emph{no learning rate schedule}. For comprehensability, mean and standard deviation are rounded.}
        \label{tab:app_tabular_large_none}
        \centering
        \begin{scriptsize}
                \begin{tabularx}{\textwidth}{lXXXXXXXX}
                        \toprule
                        \textbf{Optimizer}                    & \textbf{Quadratic Deep} & \textbf{MNIST VAE} & \textbf{F-MNIST 2c2d} & \textbf{CIFAR-10 3c3d} & \textbf{F-MNIST VAE} & \textbf{CIFAR-100} & \textbf{SVHN}   & \textbf{Tolstoi} \\
                        \midrule
                        \colordot{amsbound} \amsbound       &   86.35 $\pm$ 3.47 &   28.14 $\pm$ 0.15 &  92.15 $\pm$ 0.13 &  82.99 $\pm$ 0.78 &  23.55 $\pm$ 0.18 &     54.64 $\pm$ 1.33 &       95.31 $\pm$ 0.31 &  59.70 $\pm$ 0.16 \\
                        \colordot{amsgrad} \amsgrad         &   87.64 $\pm$ 1.00 &   27.85 $\pm$ 0.07 &  \textbf{92.26} $\pm$ 0.16 &  83.42 $\pm$ 0.65 &  23.11 $\pm$ 0.10 &     52.34 $\pm$ 1.03 &       95.58 $\pm$ 0.31 &  61.52 $\pm$ 0.13 \\
                        \colordot{adabelief} \adabelief     &   87.17 $\pm$ 0.03 &   28.01 $\pm$ 0.06 &  92.06 $\pm$ 0.24 &  82.85 $\pm$ 0.59 &  23.22 $\pm$ 0.08 &     53.76 $\pm$ 1.35 &       95.09 $\pm$ 0.30 &  61.26 $\pm$ 0.17 \\
                        \colordot{adabound} \adabound       &   94.66 $\pm$ 6.25 &   28.14 $\pm$ 0.13 &  92.03 $\pm$ 0.13 &  83.39 $\pm$ 0.53 &  23.38 $\pm$ 0.09 &     54.77 $\pm$ 0.94 &       95.40 $\pm$ 0.29 &  59.73 $\pm$ 0.20 \\
                        \colordot{adadelta} \adadelta       &  106.95 $\pm$ 0.14 &   27.87 $\pm$ 0.10 &  92.07 $\pm$ 0.11 &  83.34 $\pm$ 0.74 &  23.18 $\pm$ 0.13 &     53.18 $\pm$ 2.48 &       95.30 $\pm$ 0.60 &  60.54 $\pm$ 0.15 \\
                        \colordot{adagrad} \adagrad         &   86.70 $\pm$ 1.99 &   28.04 $\pm$ 0.29 &  92.05 $\pm$ 0.17 &  83.08 $\pm$ 0.41 &  23.16 $\pm$ 0.04 &    43.63 $\pm$ 21.35 &       95.34 $\pm$ 0.49 &  62.01 $\pm$ 0.10 \\
                        \colordot{adam} \adam               &   86.58 $\pm$ 1.95 &   27.77 $\pm$ 0.03 &  91.69 $\pm$ 0.16 &  82.95 $\pm$ 0.70 &  23.06 $\pm$ 0.10 &     54.84 $\pm$ 0.65 &       94.84 $\pm$ 0.30 &  61.97 $\pm$ 0.12 \\
                        \colordot{lookaheadmomentum} \lamom &   87.17 $\pm$ 0.07 &   52.86 $\pm$ 0.84 &  91.74 $\pm$ 0.19 &  74.01 $\pm$ 3.70 &  25.37 $\pm$ 0.35 &     57.32 $\pm$ 0.80 &       \textbf{95.82} $\pm$ 0.11 &  61.44 $\pm$ 0.17 \\
                        \colordot{lookaheadradam} \laradam  &   89.03 $\pm$ 0.87 &   34.26 $\pm$ 9.37 &  92.05 $\pm$ 0.16 &  83.00 $\pm$ 0.64 &  24.04 $\pm$ 0.25 &     54.92 $\pm$ 0.97 &       95.67 $\pm$ 0.11 &  61.73 $\pm$ 0.10 \\
                        \colordot{momentum} \momentum       &   87.04 $\pm$ 0.02 &  36.00 $\pm$ 11.09 &  91.87 $\pm$ 0.12 &  83.16 $\pm$ 0.56 &  23.86 $\pm$ 0.15 &     56.21 $\pm$ 0.67 &       95.37 $\pm$ 0.27 &  61.97 $\pm$ 0.12 \\
                        \colordot{nag} \nag                 &   87.08 $\pm$ 0.02 &  36.16 $\pm$ 10.99 &  91.87 $\pm$ 0.12 &  83.30 $\pm$ 0.88 &  23.85 $\pm$ 0.22 &     \textbf{57.85} $\pm$ 0.77 &       95.28 $\pm$ 0.23 &  61.74 $\pm$ 0.12 \\
                        \colordot{nadam} \nadam             &   86.45 $\pm$ 1.94 &   \textbf{27.73} $\pm$ 0.09 &  91.75 $\pm$ 0.42 &  \textbf{83.58} $\pm$ 0.45 &  \textbf{23.00} $\pm$ 0.07 &     53.44 $\pm$ 1.27 &       95.00 $\pm$ 0.25 &  62.01 $\pm$ 0.11 \\
                        \colordot{radam} \radam             &   86.43 $\pm$ 1.93 &   27.81 $\pm$ 0.06 &  91.63 $\pm$ 0.24 &  82.85 $\pm$ 0.52 &  23.10 $\pm$ 0.11 &     53.98 $\pm$ 1.00 &       94.83 $\pm$ 0.38 &  61.98 $\pm$ 0.13 \\
                        \colordot{rmsprop} \rmsprop         &   87.38 $\pm$ 0.12 &   27.86 $\pm$ 0.08 &  91.79 $\pm$ 0.36 &  82.16 $\pm$ 0.65 &  23.11 $\pm$ 0.08 &     52.16 $\pm$ 0.99 &       95.25 $\pm$ 0.34 &  \textbf{62.24} $\pm$ 0.07 \\
                        \colordot{sgd} \sgd                 &   \textbf{86.29} $\pm$ 3.44 &  36.17 $\pm$ 10.97 &  91.80 $\pm$ 0.23 &  82.64 $\pm$ 0.91 &  23.83 $\pm$ 0.22 &     50.58 $\pm$ 1.49 &       95.11 $\pm$ 0.31 &  61.29 $\pm$ 0.14 \\
                        \bottomrule
                \end{tabularx}
        \end{scriptsize}
\end{table}

\begin{table}[H]
	\caption{Tabular version of \Cref{fig:appendix_pc_sb}. Mean test set performance and standard deviation over $10$ random seeds of all tested optimizers on all eight optimization problems using the \emph{small budget} for tuning and \emph{no learning rate schedule}. For comprehensability, mean and standard deviation are rounded.}
	\label{tab:app_tabular_small_none}
	\centering
	\begin{scriptsize}
		\begin{tabularx}{\textwidth}{lXXXXXXXX}
			\toprule
			\textbf{Optimizer}                    & \textbf{Quadratic Deep} & \textbf{MNIST VAE} & \textbf{F-MNIST 2c2d} & \textbf{CIFAR-10 3c3d} & \textbf{F-MNIST VAE} & \textbf{CIFAR-100} & \textbf{SVHN}   & \textbf{Tolstoi} \\
			\midrule
			\colordot{amsbound} \amsbound       &   92.80 $\pm$ 5.99 &   28.18 $\pm$ 0.14 &  91.99 $\pm$ 0.10 &  83.15 $\pm$ 0.65 &  23.50 $\pm$ 0.11 &     54.91 $\pm$ 0.54 &       95.33 $\pm$ 0.17 &  58.25 $\pm$ 0.19 \\
			\colordot{amsgrad} \amsgrad         &   87.58 $\pm$ 0.71 &   27.87 $\pm$ 0.08 &  92.01 $\pm$ 0.09 &  82.25 $\pm$ 0.54 &  23.21 $\pm$ 0.06 &     52.71 $\pm$ 0.97 &       95.25 $\pm$ 0.21 &  61.61 $\pm$ 0.14 \\
			\colordot{adabelief} \adabelief     &   87.18 $\pm$ 0.03 &   27.99 $\pm$ 0.06 &  91.94 $\pm$ 0.33 &  83.13 $\pm$ 0.60 &  23.17 $\pm$ 0.07 &     53.17 $\pm$ 1.15 &       94.99 $\pm$ 0.31 &  61.09 $\pm$ 0.09 \\
			\colordot{adabound} \adabound       &   94.66 $\pm$ 6.25 &   28.11 $\pm$ 0.09 &  \textbf{92.08} $\pm$ 0.20 &  82.64 $\pm$ 1.03 &  23.40 $\pm$ 0.06 &    50.10 $\pm$ 16.39 &       95.33 $\pm$ 0.16 &  58.88 $\pm$ 0.16 \\
			\colordot{adadelta} \adadelta       &  123.86 $\pm$ 0.24 &   28.03 $\pm$ 0.08 &  91.84 $\pm$ 0.11 &  81.31 $\pm$ 1.40 &  23.50 $\pm$ 0.17 &     50.14 $\pm$ 2.29 &       95.21 $\pm$ 0.29 &  59.40 $\pm$ 0.11 \\
			\colordot{adagrad} \adagrad         &   87.14 $\pm$ 1.02 &   27.98 $\pm$ 0.16 &  \textbf{92.08} $\pm$ 0.23 &  83.25 $\pm$ 0.51 &  23.19 $\pm$ 0.08 &    37.90 $\pm$ 24.22 &       95.02 $\pm$ 0.21 &  62.01 $\pm$ 0.11 \\
			\colordot{adam} \adam               &   87.68 $\pm$ 1.44 &   27.81 $\pm$ 0.06 &  91.67 $\pm$ 0.25 &  81.90 $\pm$ 0.86 &  \textbf{23.10} $\pm$ 0.11 &     52.96 $\pm$ 1.34 &       94.84 $\pm$ 0.38 &  61.79 $\pm$ 0.06 \\
			\colordot{lookaheadmomentum} \lamom &   87.16 $\pm$ 0.06 &   55.20 $\pm$ 0.86 &  91.58 $\pm$ 0.15 &  82.72 $\pm$ 1.24 &  25.28 $\pm$ 0.23 &     57.68 $\pm$ 0.60 &       \textbf{95.80} $\pm$ 0.10 &  60.23 $\pm$ 0.26 \\
			\colordot{lookaheadradam} \laradam  &   93.75 $\pm$ 3.15 &   38.11 $\pm$ 9.73 &  91.97 $\pm$ 0.22 &  \textbf{84.70} $\pm$ 0.30 &  24.53 $\pm$ 0.15 &     55.09 $\pm$ 0.98 &       95.62 $\pm$ 0.19 &  60.00 $\pm$ 0.11 \\
			\colordot{momentum} \momentum       &   87.03 $\pm$ 0.02 &  36.08 $\pm$ 11.04 &  91.87 $\pm$ 0.16 &  83.00 $\pm$ 0.71 &  23.93 $\pm$ 0.30 &     55.96 $\pm$ 0.92 &       95.34 $\pm$ 0.23 &  61.93 $\pm$ 0.10 \\
			\colordot{nag} \nag                 &   87.08 $\pm$ 0.02 &  36.18 $\pm$ 10.97 &  92.05 $\pm$ 0.13 &  83.32 $\pm$ 0.57 &  23.87 $\pm$ 0.33 &     \textbf{57.75} $\pm$ 0.71 &       95.51 $\pm$ 0.21 &  62.07 $\pm$ 0.10 \\
			\colordot{nadam} \nadam             &   86.45 $\pm$ 1.94 &   \textbf{27.77} $\pm$ 0.06 &  91.59 $\pm$ 0.25 &  82.94 $\pm$ 0.61 &  23.12 $\pm$ 0.06 &     53.30 $\pm$ 0.90 &       94.99 $\pm$ 0.18 &  61.97 $\pm$ 0.08 \\
			\colordot{radam} \radam             &   \textbf{86.43} $\pm$ 1.93 &   27.82 $\pm$ 0.06 &  91.49 $\pm$ 0.40 &  82.27 $\pm$ 0.53 &  23.12 $\pm$ 0.07 &     53.47 $\pm$ 0.86 &       94.79 $\pm$ 0.38 &  61.93 $\pm$ 0.14 \\
			\colordot{rmsprop} \rmsprop         &   87.40 $\pm$ 0.14 &   28.03 $\pm$ 0.13 &  91.27 $\pm$ 0.28 &  82.56 $\pm$ 0.71 &  23.26 $\pm$ 0.08 &     51.20 $\pm$ 0.89 &       93.82 $\pm$ 1.64 &  \textbf{62.25} $\pm$ 0.12 \\
			\colordot{sgd} \sgd                 &   88.37 $\pm$ 3.55 &  36.18 $\pm$ 10.96 &  91.69 $\pm$ 0.15 &  82.20 $\pm$ 1.32 &  23.76 $\pm$ 0.25 &     51.53 $\pm$ 1.37 &       94.84 $\pm$ 0.56 &  61.25 $\pm$ 0.12 \\
			\bottomrule
		\end{tabularx}
	\end{scriptsize}
\end{table}

\begin{table}[H]
	\caption{Tabular version of \Cref{fig:appendix_pc_cos}. Mean test set performance and standard deviation over $10$ random seeds of all tested optimizers on all eight optimization problems using the \emph{medium budget} for tuning and the \emph{cosine learning rate schedule}. For comprehensability, mean and standard deviation are rounded.}
	\label{tab:app_tabular_medium_cosine}
	\centering
	\begin{scriptsize}
		\begin{tabularx}{\textwidth}{lXXXXXXXX}
			\toprule
			\textbf{Optimizer}                    & \textbf{Quadratic Deep} & \textbf{MNIST VAE} & \textbf{F-MNIST 2c2d} & \textbf{CIFAR-10 3c3d} & \textbf{F-MNIST VAE} & \textbf{CIFAR-100} & \textbf{SVHN}   & \textbf{Tolstoi} \\
			\midrule
			\colordot{amsbound} \amsbound       &   85.94 $\pm$ 3.41 &   28.12 $\pm$ 0.19 &  91.97 $\pm$ 0.15 &   82.91 $\pm$ 0.83 &   23.49 $\pm$ 0.07 &     54.87 $\pm$ 0.70 &       95.62 $\pm$ 0.15 &  59.31 $\pm$ 0.36 \\
			\colordot{amsgrad} \amsgrad         &   87.00 $\pm$ 0.55 &   \textbf{27.39} $\pm$ 0.04 &  92.25 $\pm$ 0.22 &   85.20 $\pm$ 0.34 &   22.83 $\pm$ 0.06 &     54.21 $\pm$ 1.99 &       96.68 $\pm$ 0.07 &  61.68 $\pm$ 0.17 \\
			\colordot{adabelief} \adabelief     &   88.12 $\pm$ 0.04 &   27.45 $\pm$ 0.05 &  \textbf{92.43} $\pm$ 0.14 &   85.47 $\pm$ 0.26 &   22.78 $\pm$ 0.04 &     57.58 $\pm$ 0.57 &       96.46 $\pm$ 0.08 &  61.09 $\pm$ 0.17 \\
			\colordot{adabound} \adabound       &   \textbf{85.92} $\pm$ 3.41 &   28.00 $\pm$ 0.09 &  92.08 $\pm$ 0.17 &   83.20 $\pm$ 0.62 &   23.38 $\pm$ 0.08 &     54.68 $\pm$ 0.81 &       95.58 $\pm$ 0.10 &  59.45 $\pm$ 0.36 \\
			\colordot{adadelta} \adadelta       &  164.58 $\pm$ 0.35 &  58.46 $\pm$ 61.52 &  92.05 $\pm$ 0.08 &   85.12 $\pm$ 0.28 &  60.55 $\pm$ 49.27 &     51.34 $\pm$ 0.64 &       96.68 $\pm$ 0.05 &  57.77 $\pm$ 0.19 \\
			\colordot{adagrad} \adagrad         &   86.61 $\pm$ 1.94 &   28.17 $\pm$ 0.27 &  91.90 $\pm$ 0.23 &   85.48 $\pm$ 0.35 &   23.36 $\pm$ 0.05 &    29.40 $\pm$ 28.41 &       96.78 $\pm$ 0.07 &  61.75 $\pm$ 0.07 \\
			\colordot{adam} \adam               &   \textbf{85.92} $\pm$ 3.41 &   27.60 $\pm$ 0.06 &  92.29 $\pm$ 0.12 &   85.27 $\pm$ 0.29 &   \textbf{22.75} $\pm$ 0.03 &     55.14 $\pm$ 0.97 &       96.67 $\pm$ 0.06 &  61.86 $\pm$ 0.16 \\
			\colordot{lookaheadmomentum} \lamom &   87.06 $\pm$ 0.02 &  76.78 $\pm$ 24.04 &  91.76 $\pm$ 0.20 &   85.61 $\pm$ 0.24 &  46.09 $\pm$ 21.85 &     62.67 $\pm$ 0.81 &       96.78 $\pm$ 0.08 &  60.26 $\pm$ 0.23 \\
			\colordot{lookaheadradam} \laradam  &   87.08 $\pm$ 0.42 &  37.41 $\pm$ 10.15 &  91.49 $\pm$ 0.24 &   85.87 $\pm$ 0.18 &   24.00 $\pm$ 0.12 &    42.00 $\pm$ 27.55 &       96.65 $\pm$ 0.09 &  61.62 $\pm$ 0.16 \\
			\colordot{momentum} \momentum       &   87.06 $\pm$ 0.02 &  36.33 $\pm$ 10.85 &  91.89 $\pm$ 0.12 &   86.13 $\pm$ 0.19 &   23.70 $\pm$ 0.18 &     63.43 $\pm$ 0.56 &       96.71 $\pm$ 0.05 &  62.26 $\pm$ 0.13 \\
			\colordot{nag} \nag                 &   87.06 $\pm$ 0.02 &  36.53 $\pm$ 10.71 &  91.76 $\pm$ 0.13 &   \textbf{87.12} $\pm$ 0.19 &  41.41 $\pm$ 21.65 &     \textbf{63.61} $\pm$ 0.46 &       96.68 $\pm$ 0.08 &  62.46 $\pm$ 0.10 \\
			\colordot{nadam} \nadam             &   85.93 $\pm$ 3.41 &   27.46 $\pm$ 0.10 &  92.42 $\pm$ 0.12 &   85.34 $\pm$ 0.34 &   22.77 $\pm$ 0.07 &     54.02 $\pm$ 0.71 &       96.62 $\pm$ 0.07 &  62.20 $\pm$ 0.12 \\
			\colordot{radam} \radam             &   86.49 $\pm$ 1.94 &   27.51 $\pm$ 0.05 &  92.33 $\pm$ 0.10 &   85.47 $\pm$ 0.36 &   22.82 $\pm$ 0.08 &     55.31 $\pm$ 0.86 &       96.61 $\pm$ 0.07 &  61.87 $\pm$ 0.19 \\
			\colordot{rmsprop} \rmsprop         &   87.09 $\pm$ 0.01 &   27.57 $\pm$ 0.05 &  92.22 $\pm$ 0.18 &   84.54 $\pm$ 0.25 &   22.80 $\pm$ 0.04 &    48.02 $\pm$ 15.69 &       96.65 $\pm$ 0.06 &  \textbf{62.85} $\pm$ 0.06 \\
			\colordot{sgd} \sgd                 &   86.30 $\pm$ 3.41 &  36.47 $\pm$ 10.76 &  91.72 $\pm$ 0.21 &  70.50 $\pm$ 30.76 &   23.54 $\pm$ 0.13 &    42.29 $\pm$ 27.05 &       \textbf{96.80} $\pm$ 0.08 &  60.40 $\pm$ 0.11 \\
			\bottomrule
		\end{tabularx}
	\end{scriptsize}
\end{table}

\begin{table}[H]
	\caption{Tabular version of \Cref{fig:appendix_pc_ltr}. Mean test set performance and standard deviation over $10$ random seeds of all tested optimizers on all eight optimization problems using the \emph{large budget} for tuning and \emph{trapezoidal learning rate schedule}. For comprehensability, mean and standard deviation are rounded.}
	\label{tab:app_tabular_large_ltr}
	\centering
	\begin{scriptsize}
		\begin{tabularx}{\textwidth}{lXXXXXXXX}
			\toprule
			\textbf{Optimizer}                    & \textbf{Quadratic Deep} & \textbf{MNIST VAE} & \textbf{F-MNIST 2c2d} & \textbf{CIFAR-10 3c3d} & \textbf{F-MNIST VAE} & \textbf{CIFAR-100} & \textbf{SVHN}   & \textbf{Tolstoi} \\
			\midrule
			\colordot{amsbound} \amsbound       &   86.78 $\pm$ 2.04 &   28.18 $\pm$ 0.19 &  92.11 $\pm$ 0.16 &  83.11 $\pm$ 0.84 &                                   23.49 $\pm$ 0.11 &     54.28 $\pm$ 1.23 &       95.46 $\pm$ 0.21 &  59.70 $\pm$ 0.14 \\
			\colordot{amsgrad} \amsgrad         &   \textbf{85.94} $\pm$ 3.42 &   27.57 $\pm$ 0.06 &  \textbf{92.29} $\pm$ 0.12 &  84.71 $\pm$ 0.31 &                                   22.87 $\pm$ 0.06 &     57.15 $\pm$ 0.89 &       96.42 $\pm$ 0.06 &  61.86 $\pm$ 0.14 \\
			\colordot{adabelief} \adabelief     &   87.19 $\pm$ 0.02 &   27.75 $\pm$ 0.05 &  92.27 $\pm$ 0.10 &  84.90 $\pm$ 0.32 &                                   22.93 $\pm$ 0.07 &     58.66 $\pm$ 0.50 &       96.35 $\pm$ 0.07 &  61.50 $\pm$ 0.15 \\
			\colordot{adabound} \adabound       &   91.34 $\pm$ 5.60 &   28.11 $\pm$ 0.09 &  92.08 $\pm$ 0.14 &  83.23 $\pm$ 0.58 &                                   23.37 $\pm$ 0.05 &     54.50 $\pm$ 1.23 &       95.45 $\pm$ 0.18 &  59.72 $\pm$ 0.17 \\
			\colordot{adadelta} \adadelta       &  108.26 $\pm$ 0.14 &   27.60 $\pm$ 0.08 &  91.87 $\pm$ 0.20 &  85.40 $\pm$ 0.17 &                                   22.87 $\pm$ 0.08 &     59.67 $\pm$ 0.38 &       96.58 $\pm$ 0.07 &  60.41 $\pm$ 0.11 \\
			\colordot{adagrad} \adagrad         &   86.51 $\pm$ 1.95 &   27.83 $\pm$ 0.15 &  91.88 $\pm$ 0.12 &  84.84 $\pm$ 0.23 &  7e23 $\pm$ 2e24 &    48.31 $\pm$ 23.66 &       96.48 $\pm$ 0.10 &  62.35 $\pm$ 0.16 \\
			\colordot{adam} \adam               &   88.01 $\pm$ 3.63 &   27.52 $\pm$ 0.06 &  92.09 $\pm$ 0.14 &  84.66 $\pm$ 0.42 &                                   \textbf{22.80} $\pm$ 0.05 &     58.52 $\pm$ 0.61 &       96.22 $\pm$ 0.08 &  62.31 $\pm$ 0.10 \\
			\colordot{lookaheadmomentum} \lamom &   87.12 $\pm$ 0.02 &   52.89 $\pm$ 0.00 &  91.87 $\pm$ 0.17 &  84.85 $\pm$ 0.60 &                                  28.24 $\pm$ 13.23 &     62.69 $\pm$ 0.42 &       96.48 $\pm$ 0.10 &  61.81 $\pm$ 0.17 \\
			\colordot{lookaheadradam} \laradam  &   88.67 $\pm$ 1.24 &  36.14 $\pm$ 10.99 &  91.96 $\pm$ 0.14 &  \textbf{86.31} $\pm$ 0.25 &                                   23.83 $\pm$ 0.14 &    56.22 $\pm$ 18.42 &       \textbf{96.62} $\pm$ 0.08 &  62.03 $\pm$ 0.14 \\
			\colordot{momentum} \momentum       &   87.06 $\pm$ 0.02 &   33.77 $\pm$ 9.62 &  91.67 $\pm$ 0.22 &  85.02 $\pm$ 0.30 &                                   23.45 $\pm$ 0.22 &     62.78 $\pm$ 0.34 &       96.50 $\pm$ 0.08 &  62.40 $\pm$ 0.08 \\
			\colordot{nag} \nag                 &   87.06 $\pm$ 0.02 &  35.80 $\pm$ 11.20 &  92.08 $\pm$ 0.16 &  85.00 $\pm$ 0.44 &                                   23.38 $\pm$ 0.16 &     \textbf{63.30} $\pm$ 0.31 &       96.43 $\pm$ 0.11 &  62.41 $\pm$ 0.09 \\
			\colordot{nadam} \nadam             &   87.03 $\pm$ 3.66 &   \textbf{27.51} $\pm$ 0.08 &  92.28 $\pm$ 0.11 &  84.96 $\pm$ 0.37 &                                   22.83 $\pm$ 0.08 &     58.96 $\pm$ 0.77 &       96.27 $\pm$ 0.10 &  62.28 $\pm$ 0.11 \\
			\colordot{radam} \radam             &   86.43 $\pm$ 1.93 &   \textbf{27.51} $\pm$ 0.05 &  92.17 $\pm$ 0.17 &  84.86 $\pm$ 0.32 &                                   22.83 $\pm$ 0.07 &     59.01 $\pm$ 0.73 &       96.29 $\pm$ 0.09 &  62.24 $\pm$ 0.13 \\
			\colordot{rmsprop} \rmsprop         &   87.14 $\pm$ 0.03 &   27.58 $\pm$ 0.07 &  92.23 $\pm$ 0.13 &  84.11 $\pm$ 0.16 &                                   22.85 $\pm$ 0.05 &    30.15 $\pm$ 29.15 &       96.25 $\pm$ 0.09 &  \textbf{62.59} $\pm$ 0.11 \\
			\colordot{sgd} \sgd                 &   86.05 $\pm$ 3.40 &  35.71 $\pm$ 11.26 &  91.88 $\pm$ 0.17 &  84.83 $\pm$ 0.27 &                                   23.43 $\pm$ 0.19 &    31.36 $\pm$ 30.38 &       96.42 $\pm$ 0.07 &  61.25 $\pm$ 0.11 \\
			\bottomrule
		\end{tabularx}
	\end{scriptsize}
\end{table}

\begin{table}[H]
	\caption{Tabular version of \Cref{fig:appendix_pc_train_loss}. Mean \emph{training} set performance and standard deviation over $10$ random seeds of all tested optimizers on all eight optimization problems using the \emph{large budget} for tuning and \emph{no learning rate schedule}. For comprehensability, mean and standard deviation are rounded.}
	\label{tab:app_tabular_large_none_TRAIN}
	\centering
	\begin{scriptsize}
		\begin{tabularx}{\textwidth}{lXXXXXXXX}
			\toprule
			\textbf{Optimizer}                    & \textbf{Quadratic Deep} & \textbf{MNIST VAE} & \textbf{F-MNIST 2c2d} & \textbf{CIFAR-10 3c3d} & \textbf{F-MNIST VAE} & \textbf{CIFAR-100} & \textbf{SVHN}   & \textbf{Tolstoi} \\
			\midrule
			\colordot{amsbound} \amsbound       &   84.10 $\pm$ 3.34 &   27.84 $\pm$ 0.15 &  \textbf{0.00} $\pm$ 0.00 &  0.58 $\pm$ 0.02 &  23.46 $\pm$ 0.22 &      1.80 $\pm$ 0.09 &        0.17 $\pm$ 0.00 &   1.26 $\pm$ 0.01 \\
			\colordot{amsgrad} \amsgrad         &   85.24 $\pm$ 1.21 &   27.20 $\pm$ 0.09 &  \textbf{0.00} $\pm$ 0.00 &  0.56 $\pm$ 0.01 &  22.77 $\pm$ 0.10 &      1.62 $\pm$ 0.11 &        0.07 $\pm$ 0.00 &   1.18 $\pm$ 0.01 \\
			\colordot{adabelief} \adabelief     &   84.87 $\pm$ 0.30 &   27.16 $\pm$ 0.07 &  \textbf{0.00} $\pm$ 0.00 &  0.56 $\pm$ 0.02 &  22.68 $\pm$ 0.07 &      1.75 $\pm$ 0.11 &        0.10 $\pm$ 0.00 &   1.19 $\pm$ 0.01 \\
			\colordot{adabound} \adabound       &   92.10 $\pm$ 5.64 &   27.86 $\pm$ 0.17 &  \textbf{0.00} $\pm$ 0.00 &  0.57 $\pm$ 0.02 &  23.27 $\pm$ 0.14 &      1.87 $\pm$ 0.09 &        0.08 $\pm$ 0.01 &   1.26 $\pm$ 0.01 \\
			\colordot{adadelta} \adadelta       &  104.99 $\pm$ 0.30 &   27.16 $\pm$ 0.12 &  \textbf{0.00} $\pm$ 0.00 &  \textbf{0.52} $\pm$ 0.01 &  22.79 $\pm$ 0.15 &      1.91 $\pm$ 0.17 &        0.11 $\pm$ 0.01 &   1.21 $\pm$ 0.01 \\
			\colordot{adagrad} \adagrad         &   84.40 $\pm$ 1.73 &   27.58 $\pm$ 0.34 &  \textbf{0.00} $\pm$ 0.00 &  0.53 $\pm$ 0.02 &  22.94 $\pm$ 0.06 &      5.25 $\pm$ 6.85 &        0.10 $\pm$ 0.01 &   1.15 $\pm$ 0.01 \\
			\colordot{adam} \adam               &   84.33 $\pm$ 1.76 &   26.99 $\pm$ 0.07 &  \textbf{0.00} $\pm$ 0.00 &  0.56 $\pm$ 0.03 &  22.73 $\pm$ 0.12 &      1.79 $\pm$ 0.09 &        0.10 $\pm$ 0.01 &   1.15 $\pm$ 0.00 \\
			\colordot{lookaheadmomentum} \lamom &   84.85 $\pm$ 0.30 &   52.85 $\pm$ 0.74 &  0.06 $\pm$ 0.02 &  1.00 $\pm$ 0.16 &  25.40 $\pm$ 0.39 &      1.76 $\pm$ 0.06 &        \textbf{0.06} $\pm$ 0.00 &   1.17 $\pm$ 0.01 \\
			\colordot{lookaheadradam} \laradam  &   86.68 $\pm$ 1.10 &   34.33 $\pm$ 9.29 &  \textbf{0.00} $\pm$ 0.00 &  0.57 $\pm$ 0.02 &  24.00 $\pm$ 0.26 &      1.75 $\pm$ 0.08 &        0.11 $\pm$ 0.00 &   1.16 $\pm$ 0.00 \\
			\colordot{momentum} \momentum       &   84.77 $\pm$ 0.30 &  35.98 $\pm$ 11.06 &  \textbf{0.00} $\pm$ 0.00 &  0.57 $\pm$ 0.02 &  23.71 $\pm$ 0.13 &      1.84 $\pm$ 0.07 &        0.13 $\pm$ 0.00 &   1.15 $\pm$ 0.01 \\
			\colordot{nag} \nag                 &   84.77 $\pm$ 0.30 &  36.15 $\pm$ 10.96 &  \textbf{0.00} $\pm$ 0.00 &  0.54 $\pm$ 0.02 &  23.67 $\pm$ 0.22 &      1.65 $\pm$ 0.03 &        0.18 $\pm$ 0.00 &   1.16 $\pm$ 0.01 \\
			\colordot{nadam} \nadam             &   84.19 $\pm$ 1.74 &   \textbf{26.77} $\pm$ 0.12 &  0.01 $\pm$ 0.01 &  0.55 $\pm$ 0.02 &  \textbf{22.59} $\pm$ 0.08 &      1.84 $\pm$ 0.08 &        0.10 $\pm$ 0.00 &   1.15 $\pm$ 0.01 \\
			\colordot{radam} \radam             &   84.18 $\pm$ 1.74 &   26.91 $\pm$ 0.10 &  0.01 $\pm$ 0.00 &  0.55 $\pm$ 0.02 &  22.77 $\pm$ 0.08 &      \textbf{1.54} $\pm$ 0.10 &        0.10 $\pm$ 0.00 &   1.15 $\pm$ 0.01 \\
			\colordot{rmsprop} \rmsprop         &   85.02 $\pm$ 0.32 &   27.18 $\pm$ 0.13 &  0.01 $\pm$ 0.01 &  0.57 $\pm$ 0.02 &  22.73 $\pm$ 0.08 &      1.69 $\pm$ 0.13 &        0.11 $\pm$ 0.01 &   \textbf{1.14} $\pm$ 0.00 \\
			\colordot{sgd} \sgd                 &   \textbf{83.95} $\pm$ 3.25 &  36.15 $\pm$ 10.95 &  \textbf{0.00} $\pm$ 0.00 &  0.63 $\pm$ 0.02 &  23.71 $\pm$ 0.23 &      1.70 $\pm$ 0.10 &        0.15 $\pm$ 0.01 &   1.18 $\pm$ 0.00 \\
			\bottomrule
		\end{tabularx}
	\end{scriptsize}
\end{table}
\end{appendices}

\clearpage
\bibliographystyleappendix{bibliography/icml2021}
\bibliographyappendix{bibliography/references}

\end{document}